\documentclass[journal,compsoc]{IEEEtran}
\usepackage{amsmath, amsfonts, amssymb, cite, epsfig, mathtools,adjustbox}
\usepackage{algorithmic,amsthm}
\usepackage{graphicx}
\usepackage{algorithm}
\usepackage{subcaption}
\usepackage{color}
\usepackage{bm}
\DeclareMathOperator*{\diff}{diff}

\def \1{\mathbf{\,1}}

\newtheorem{theorem}{Theorem}
\newtheorem{definition}{Definition}
\graphicspath{{figures/}}

\def \FigWidthSmall{0.43\textwidth}

\def \SimFigWidth{0.48\textwidth}

\def\1{\mathbf 1}

\begin{document}
\title{LoMar: A Local Defense Against Poisoning Attack on Federated Learning}
\author{Xingyu~Li,
	Zhe~Qu,
	Shangqing~Zhao,
	Bo~Tang,~\IEEEmembership{Member,~IEEE},
	Zhuo~Lu,~\IEEEmembership{Senior~Member,~IEEE},
	and~Yao~Liu,~\IEEEmembership{Senior~Member,~IEEE}
	\IEEEcompsocitemizethanks{\IEEEcompsocthanksitem Xingyu~Li and Bo~Tang are with the Department of Electrical and Computer Engineering at Mississippi State University, Mississippi State, MS, 39762. E-mail: xl292@msstate.edu, tang@ece.msstate.edu. 
		\IEEEcompsocthanksitem Zhe~Qu and Zhuo~Lu are with the Department of Electrical Engineering at University of South Florida, Tampa, FL, 33620. E-mail: zhequ@usf.edu, zhuolu@usf.edu, yliu@cse.usf.edu.
		\IEEEcompsocthanksitem Shangqing~Zhao is with the School of Computer Science at University of Oklahoma, Tulsa, OK 74135. E-mail: shangqing@ou.edu.
		\IEEEcompsocthanksitem Yao~Liu is with the Department of Computer Science and Engineering at University of South Florida, Tampa, FL, 33620. E-mail: yliu@cse.usf.edu.
		\IEEEcompsocthanksitem{Xingyu~Li and Zhe~Qu are co-first authors.}
	}
}

\IEEEcompsoctitleabstractindextext{
	\begin{abstract}
		Federated learning (FL) provides a high efficient decentralized machine learning framework, where the training data remains distributed at remote clients in a network. Though FL enables a privacy-preserving mobile edge computing framework using IoT devices, recent studies have shown that this approach is susceptible to poisoning attacks from the side of remote clients. To address the poisoning attacks on FL, we provide a \textit{two-phase} defense algorithm called $\textit{\underline{Lo}cal \underline{Ma}licious Facto\underline{r}}$ (LoMar). In phase I, LoMar scores model updates from each remote client by measuring the relative distribution over their neighbors using a kernel density estimation method. In phase II,  an optimal threshold is approximated to distinguish malicious and clean updates from a statistical perspective. Comprehensive experiments on four real-world datasets have been conducted, and the experimental results show that our defense strategy can effectively protect the FL system. {Specifically, the defense performance on Amazon dataset under a label-flipping attack indicates that, compared with FG+Krum, LoMar increases the target label testing accuracy from $96.0\%$ to $98.8\%$, and the overall averaged testing accuracy from $90.1\%$ to $97.0\%$.}
	\end{abstract}
	
	}

\maketitle
\IEEEpeerreviewmaketitle

\IEEEraisesectionheading{\section{Introduction}\label{Sec:introduction}}
\IEEEPARstart{F}{ederated} Learning (FL) \cite{konecny2017federated, mcmahan2017communication} has been demonstrated to be an efficient distributed machine learning framework to train a joint model from decentralized data. Recently, it has been paid more attention in the research field because of the highly developed IoT applications \cite{dabek2004vivaldi}. FL provides a privacy-preserved learning framework to address distributed optimization problems by allowing communications of learning information between remote users in the network, instead of sharing the private training datasets. Typically, a FL system consists of two parts: \textit{remote clients} and \textit{an aggregator}. Each remote client manages its private training data and performs a local learning process to obtain learning model updates, and the aggregator repeatedly updates a joint model from the received remote learning updates with an aggregation rule.



\begin{figure}[ht]
	\centering
	\includegraphics[width = \FigWidthSmall]{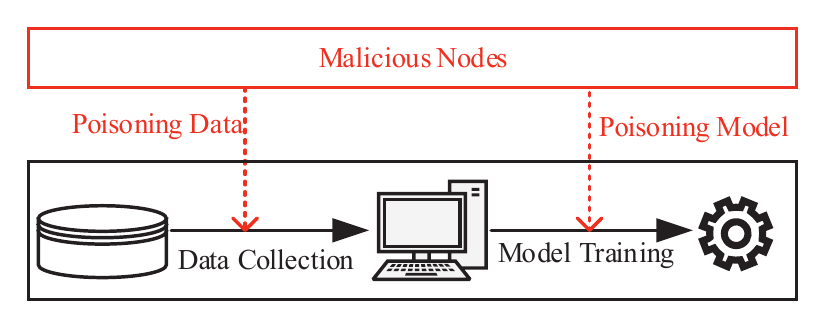}
	\caption{Data and model poisoning attacks of local training process on the remote clients in a FL system.}
	\label{Fig:systemArch}
\end{figure}

However, the distributed architecture of FL makes this learning system vulnerable to various attacks, as the remote clients could be easily compromised by attackers. Typically, the attacker can leverage the privacy property (i.e., the private remote training dataset) to intrude a number of clients and manipulate their local training process, which leads to a decreased performance of the joint model. Figure~\ref{Fig:systemArch} shows that the FL system is exposed to poisoning attacks at two stages: i) \textit{local data collection:} data poisoning attacks can inject malicious data or modify existing data during the local data collection process; ii) \textit{remote model training:} model poisoning attacks can directly inject poisoned parameters into the remotely trained model which is sent back to the aggregator. As such, poisoning attacks produce malicious updates to the aggregation process of FL. Note that, through a careful manipulation, both data and model poisoning attacks can make the FL joint learning model converged so that the attackers are hard to be detected \cite{bagdasaryan2018backdoor, bhagoji2019analyzing, melis2019exploiting, xiao2012adversarial, fung2018mitigating, fang2020local, bhagoji2018model}.

It is necessary to develop a defense method to protect the FL system against poisoning attacks. Typically, a FL defense method is considered to be successful if it can sanitize the poisoned remote updates to obtain a trusted joint model. One type of existing mechanisms \cite{blanchard2017machine, yin2018byzantine, el2018hidden, baracaldo2017mitigating} is to detect malicious updates based on Euclidean distances or angle differences between each pair of remote updates \cite{fung2018mitigating, barreno2010security}. The other type of defense methods are developed based on the Byzantine tolerance \cite{blanchard2017machine, xie2018generalized}.
However, recent studies have shown that poisoning attacks with constrained malicious data have the potential to bypass existing defense methods \cite{melis2019exploiting, tolpegin2020data}. We notice that most of the existing defense approaches only regard the malicious updates as the global anomaly to the FL system, and they do not analyze local feature patterns of the malicious remote updates. In this paper, we introduce a new defense method based on a local feature analysis strategy: the maliciousness of poisoned remote updates is evaluated according to their model parameter features. 

{We propose our \textit{two-phase} defense algorithm \textit{\underline{Lo}cal \underline{Ma}licious Facto\underline{r}} (LoMar), which is able to detect the anomalies in FL from a local view, instead of the existing global view. The main idea of the proposed LoMar is to evaluate the remote update maliciousness based on the statistical characteristic analysis of the model parameters, which is intuitively motivated by the fact that each remote update in the FL system can be considered as being generated from a specific distribution of the parameters. Specifically, once the aggregator receives remote updates from a client, instead of using the whole remote updates set, LoMar performs the feature analysis of this update with its nearest neighbors. To measure the degree of maliciousness, a non-parametric local kernel density estimation method is applied to measure the relative distribution of the remote update in its neighborhood.} {We evaluate our proposed LoMar defense algorithm via both theoretical analysis and comprehensive experiments. In summary, we highlight the contribution of this paper as follows:   }
{
	\begin{itemize}
		\item We propose a new \textit{two-phase} defense algorithm, called LoMar, in order to address poisoning attacks against FL.
		\item The proposed LoMar defense algorithm measures the malicious degree of remote updates based on its neighborhood by analyzing the statistical model parameter features via a non-parametric relative kernel density estimation method.
		\item Besides the provided theoretical analysis of LoMar, we conduct extensive performance evaluation of LoMar under two categories of poisoning attacks on FL. The results show that the proposed LoMar outperforms existing FL defense algorithms.
\end{itemize}}	

The rest of this paper is organized as follows:  Section~\ref{Sec:Formulation} formulates our defense problem against poisoning attacks on the FL. Section~\ref{Sec:KDEdefense} details the development of the proposed LoMar algorithm. Section~\ref{Sec:Evaluation} provides our experiment results and performance evaluations. Section~\ref{Sec:Relatedworks} further summarizes the related work, followed by a conclusion and future work discussion in Section~\ref{Sec:Conclusion}.

\section{Background and Problem Formulation}\label{Sec:Formulation}
\subsection{Federated Learning}\label{Sec:FL}
FL systems are developed to train a joint model $\mathbf{w}$ in order to address a distributed optimization problem (e.g., image classification) in a decentralized network. Typically, a FL system is composed by $N$ remote clients and one aggregator. Considering that the whole training data in the FL system is $\mathcal{D}$ and each remote client has its private training dataset (e.g., $\mathcal{D}_i$ for the $i$-th client), we have $\mathcal{D} = \{\mathcal{D}_1, \dots, \mathcal{D}_i, \dots, \mathcal{D}_{N} \}$, where the size of $i$-th private dataset is denoted as $l_i = |\mathcal{D}_i|$ and the total number of training samples is $l = |\mathcal{D}| = \sum_i^N l_i $. Different from the conventional learning methods, the training data in FL is only stored and processed on the remote clients. {Meanwhile, the aggregator maintains a joint model and repeatedly updates it with the received local learning updates from remote clients according to a certain aggregation rule. At the beginning of a FL process, the aggregator initializes a joint model $\mathbf{w}^{0}$ and distributes it to all remote clients for local training.} Specifically, at the $t$-th iteration, the FL system repeats the following three steps to obtain the joint model $\mathbf{w}^{t}$ from the current $\mathbf{w}^{t-1}$. 

\textbf{Step I.} The aggregator delivers the joint model $\mathbf{w}^{t-1}$ to all remote clients. 

\textbf{Step II.} Each remote client performs its local learning process with its local training data and the received joint model $\mathbf{w}^{t-1}$. During the local learning process, an updated local model $\mathbf{w}_i^{t}$ is produced by the stochastic gradient descent as follows:
\begin{equation}\label{Eq:localupdate}
\mathbf{w}_i^{t} = {\mathbf{w}}^{t-1} - \eta_i\nabla {L}_i (\mathbf{w}^{t-1}, \mathcal{D}_{i}),
\end{equation}
where $\eta_i$ is the learning rate of local model training and $ \nabla {L}_i (\mathbf{w}^{t-1}, \mathbf{D}_{l_i})$ denotes the gradient of local optimization loss. Once the local training is finished (e.g., with several local epochs), each client sends back the local training update $\bm{u}_i^{t} = \mathbf{w}_i^{t} - \mathbf{w}^{t-1}$.  

\textbf{Step III.} The aggregator aggregates all local model updates to obtain the updated joint model $\mathbf{w}^{t}$ using a specific aggregation rule, e.g., the typical weighted averaging rule (FedAvg), given by: 
\begin{equation}\label{Eq:update}
\mathbf{w}^{t} =\mathbf{w}^{t-1} +  \sum_{i=1}^{M} \alpha_{i}\bm{u}_i^{t},
\end{equation}
where $\alpha_i = \frac{l_i}{l}$. Usually, the aggregation rule in Eq.~\ref{Eq:update}
guarantees an optimized solution for the distributed optimization problem in the FL system. 

\subsection{Attacker} \label{Sec:attacker}
\begin{figure}[tb]
	\centering
	\includegraphics[width=\FigWidthSmall]{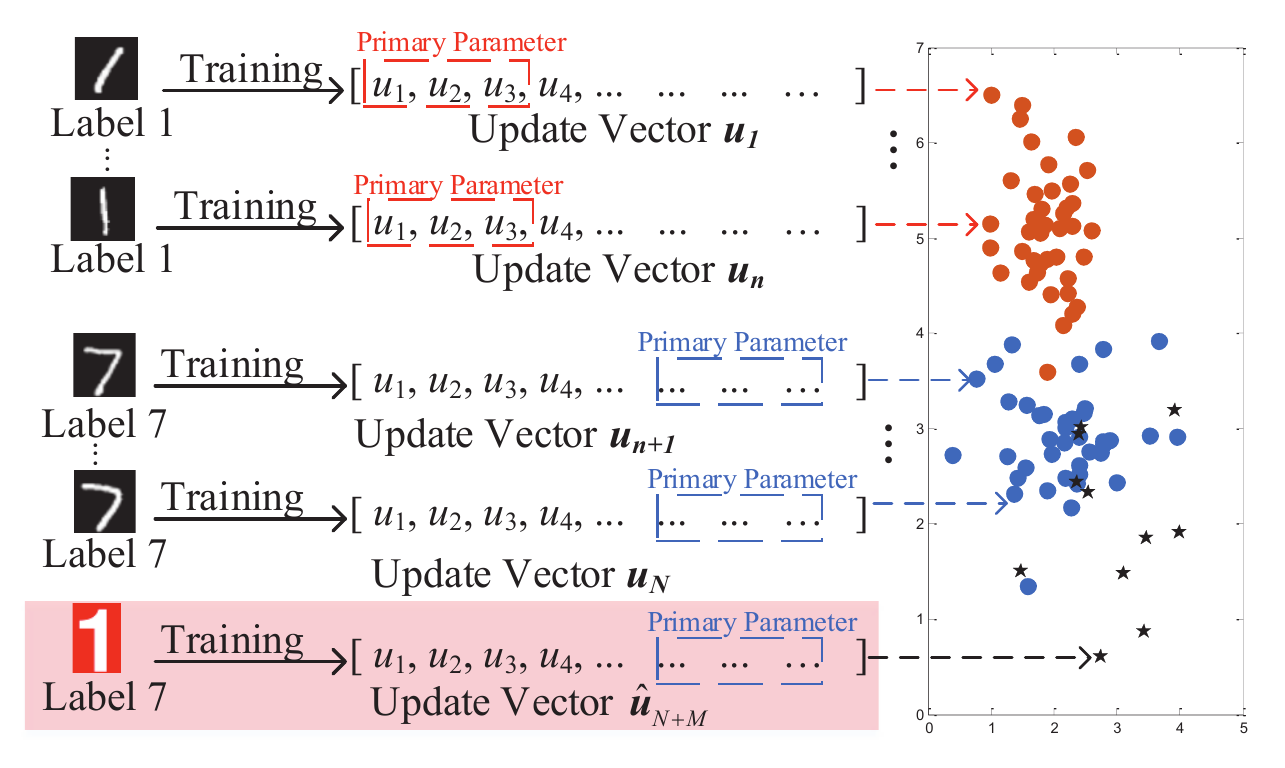}
	\caption{{Model updates under attack on MNIST.}}
	\label{fig:classifier_malicious}
\end{figure}
{
	In general, there are two types of poisoning attacks in the machine learning field: \textit{targeted poisoning attack} and \textit{non-targeted poisoning attack}. Though the ultimate goal for both types can be considered as decreasing the performance of FL learned joint model, the implementation details can vary. For example, the attacker in targeted poisoning attacks typically aims to take control of a specific ratio of the learning objective (e.g., the targeted labels in the image classification). Moreover, some of these attacks develop their attack mechanisms based on a trade-off strategy to make the ML model converge on the other labels while successfully controlling the target label. In this situation, the attacks are more difficult to be detected. And for the non-targeted poisoning attacks, they usually manipulate the ML model such that it would have a high error rate for testing data samples and be unusable for further learning tasks, which does not target a specific  classification label.} 

{
	As FL gets more popular in recent years, the security problems in the FL systems gain much interest from the ML community. Especially, compared to centralized learning models, implementing an attack on the FL system is much easier because of its loose structure and plenty of spaces between the remote clients and the aggregator. Therefore, multiple poisoning attacks have been developed to break the FL systems, most of which describe their attacking objective as an optimization problem and manipulate the joint model by sending poisoned remote updates. For simplicity, we denote the joint model from a clean FL system as $\mathbf{w}$ while the poisoned is represented as $\hat{\mathbf{w}}$ in the rest of this paper. Note that for better presentation, the attacker introduced in the following paper is considered as a targeted poisoning attack, which intends to manipulate one specific label.}

{
	Usually, the attackers can produce the poisoned local training updates by injecting new malicious clients into the system or manipulating original clean clients. In this paper, we consider the former scenario to present the notations. Note that the capacity of the attacker in this paper is regulated as follows: \textit{i)} the attacker injects $M$ malicious clients into the FL system, where the budget of $M$ is $M \leq 0.4N$; \textit{ii)} by default, the number of target label is considered as one. Hence, we denote the malicious training dataset, models, and updates are denoted by $\mathbf{D}_m$, $\mathbf{w}_m$, and $\bm{u}_m$, respectively.}

\subsection{Defender} \label{Sec:defender}
In order to address the poisoning attacks, we consider the ultimate goal of the defender is to successfully remove the malicious impact out of the FL system. In general, there are two existing types of mechanisms to develop a defender: {i) one is to detect the malicious based on the analysis of remote updates to the joint model (i.e., Auror\cite{shen2016auror}); ii) the other is to develop a new aggregating function to achieve the FL learning objective with the poisoned remote updates, which is also known as Byzantine tolerance (e.g., Krum \cite{yin2018byzantine})}. However, existing defense mechanisms are proved to have weak performance against poisoning attacks on distributed machine learning models, especially on FL. {For example, as shown in \cite{bhagoji2019analyzing}, the baseline label-flipping attack can bypass most of the existing defense approaches by adding a simple stealth metric.}

{After a full literature review of existing poisoning attack and defense mechanisms on FL, we summarize the reason why current defenses fail. According to the principle of parsimony, there is a common sense in the research field of ML attack that a successful attacker leverages an efficient yet effective impact on the learning model.} This indicates that compared to the clean benign updates, the poisoned remote updates share a unique difference for their attack objectives. Intuitively, the poisoning attacks on a FL system can be easily addressed if the defender finds a feasible measurement to distinguish the clean and poisoned updates. However, due to the complexity (the huge number of layers and high dimension of the input information) of the remote updates, developing a precise criterion from a global view and considering the malicious as a global anomaly can lead to the failure of developing a successful defense. 

For instance, we introduce an observation of remote updates under label flipping $1$ - $7$ attack on MNIST dataset with a logistic regression classifier in  Fig.~\ref{fig:classifier_malicious}, {where the defender needs to identify the malicious updates on label $1$. However, the defense would fail as the malicious updates may locate close to clean ones by only manipulating a small size of parameters, which makes them hard to be detected.}


In order to address this challenge, it is necessary to develop a new FL defense algorithm with a feasible measurement on the remote updates for better maliciousness detection. As such, we propose a new defender, called Local malicious factor (LoMar). Different from the existing defense methods, LoMar provides its malicious detection strategy based on the parameter features of the remote updates, which is considered to be a local criterion, instead of from a global view. Specifically, LoMar uses a scored-factor $F(i)$ to denote the degree of maliciousness for each remote client and develops a new aggregation rule on the aggregator side, which comes with a binary controller $\delta(i)$ to protect the joint model from poisoned updates. Formally, we describe the new aggregation rule of the FL system under the protection of LoMar as follows:
\begin{equation}\label{Eq:aggregationrule}
\tilde{\mathbf{w}}^t = \tilde{\mathbf{w}}^{t-1} + \sum_{i=1}^{N+M}\delta(i)\alpha_i \bm{u}_{i}^{t}.
\end{equation}
where $\tilde{\mathbf{w}}^t$ denotes the joint model in a potentially poisoned FL system with LoMar protection. 

\section{Design of LoMar Defense}\label{Sec:KDEdefense}
\subsection{Overview}

Our motivations of developing the \textit{two-phase} LoMar defense on the FL system come from two main challenges: i) finding the local density distribution of the remote update from the parameter perspective can be very difficult and expensive;  ii) it is also extremely hard to find an appropriate approach for the malicious detection of each remote update, i.e., the clean updates will be considered as malicious if the defense mechanism is too strict and the poisoning updates cannot be fully detected if its too loose. Moreover, it is unrealistic for the defender to obtain a clean joint model $\mathbf{w}$ for further analysis because the FL system can not be trusted from the start of the FL training process as it might have been already poisoned. In order to address these challenges, intuitively, although the true distribution of each remote update is not available in the FL, we can obtain a local density estimation for the $i$-th remote update as $\tilde{q}(\bm{u}_i)$, based on which all possible poison attacks can be then detected using the local outlier detection method. 

\begin{figure}[ht]
	\centering
	\includegraphics[width=\SimFigWidth]{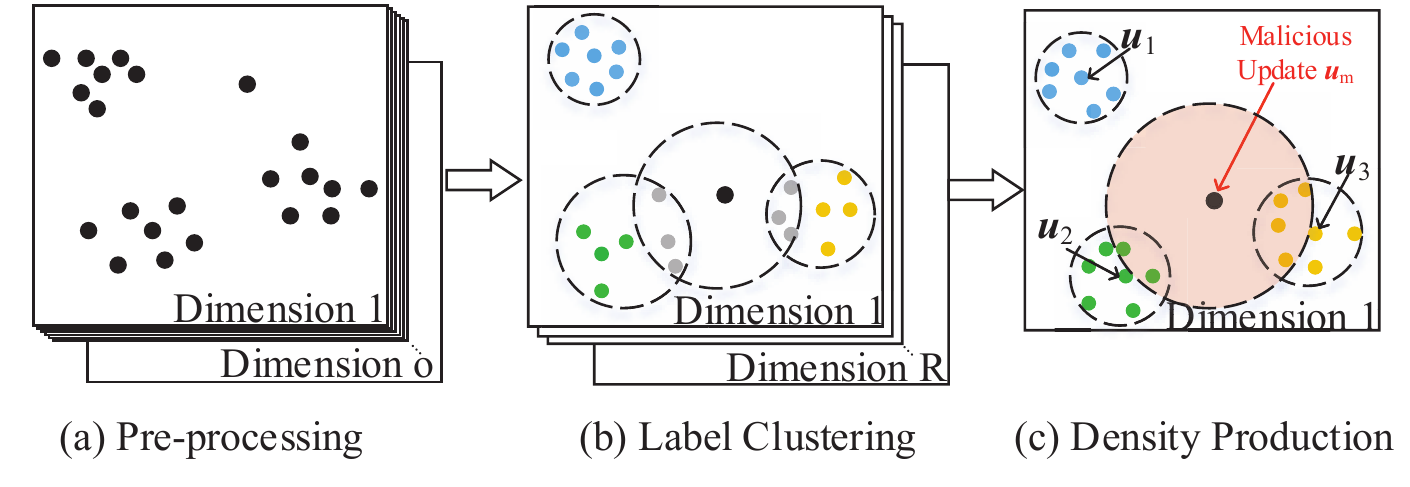}
	\caption{KDE estimation structure in LoMar.}
	\label{Fig:KDE}
\end{figure}

In phase I, for the $i$-th remote update at the $t$-th iteration, LoMar finds its $k$-nearest neighbors to be its neighborhood reference set $\mathcal{U}_i = \{\bm{u}_{i,1}, \dots, \bm{u}_{i,k}\}$, where $k \ll N+M$. {Achieving this in a FL system can be extremely difficult, as the probability distribution of remote clients is not known to the aggregator. As such, we use a state-of-art non-parametric estimation method, which is the kernel density estimation (KDE) \cite{tang2015kerneladasyn}, to estimate the local density distribution function for $\bm{u}_i$ according to each output label in the FL joint model.} {Specifically, in this paper we assume an extremely difficult scenario that the defender has no idea about the number of malicious in the FL system. Thus, we pursue an alternative solution that defines the value of $k \leq 0.4 N$ to be a loose bound, corresponding to the maximum number of potential malicious. Particularly, in the recent studies of data poisoning attacks against FL, the number of data poison is considered an important constrain. For example, \cite{huang2020metapoison} sets an upper bound of the malicious budget at only $10\%$. And work in \cite{fraboni2021free} shows that if the size of malicious is up to $50\%$, the learning speed of FL network can be very slow that making the attack easy to be found.} Then, we define the malicious factor $F(i)$ to describe the numerical malicious degree for the $i$-th client. We introduce the process of the KDE estimation in LoMar in Figure.~\ref{Fig:KDE}: i) the remote updates are divided by the dimension of the parameter features in the pre-processing process; ii) the parameters are clustered based on the output labels; iii) the kernel density estimation of each output label is performed and being a product to one numerical output. 

In phase II, we provide a boolean vector to determine the malicious status of an uncertain remote update as $1$ (clean) or $0$ (malicious). In this procedure, a threshold $\epsilon$ is developed from an inequality approach to manipulate the numerical factor $F(i)$ in phase I into the boolean output $\delta(i)$. Theoretical analysis is also performed regarding the selection of threshold in terms of false alarm rate. 

\subsection{Phase I: Malicious Client Factor}
In a FL system, the remote updates set at the $t$-th iteration is $\mathcal{U} = \{\bm{u}_1^t, \bm{u}_2^t, \dots, \bm{u}_{N+M}^t\}$. {Note that in FL, the $i$-th remote update $\bm{u}_i^t$ is always related to the iteration number, and thus in the rest of this section, we simplify the $\bm{u}_i^t$ to $\bm{u}_i$, where the superscript of $\bm{u}_i$ would be denoted to present the proposed LoMar algorithm.} The process of finding the malicious client factor $F(i)$ comes from two steps: finding the $k$-nearest neighborhoods and developing $F(i)$ with the neighborhood from the local KDE.  

\noindent\textbf{$k$-nearest neighborhood mapping.} For the remote update $\bm{u}_i$, we firstly develop a neighborhood reference set $\mathcal{U}_i = \{\bm{u}_{i,1}, \dots, \bm{u}_{i,j}, \dots, \bm{u}_{i,k}\}$, which collects the $k$-nearest neighborhoods of each update instead of the whole remote updates. {To evaluate their relative positions to the kernel center $\bm{u}_i$, we calculate the averaged $l_2$ norm distance between $\bm{u}_i$ to neighbors in $\mathcal{U}_i$, where the averaged distance is defined as $\bar{d}_i = \frac{1}{k} \sum_{j=1}^{k} \diff(\bm{u}_i, \bm{u}_{i,j})$. To further explain this definition, the function $\diff(\cdot, \cdot)$ leverages a squared Euclidean distance $||\cdot||^2$ between $\bm{u}_i$ and its neighbor.} {Additionally, we further investigate the difference of all updates in $\mathcal{U}_i$ to represent the mean value of $\bar{d}_{i}$ for each $\bm{u}_i \in \mathcal{U}_i$ as $\bar{D}_i$}
\begin{align}
\centering
\bar{D}_i = \frac{1}{k(k-1)} \sum_{i = 1}^{k} \sum_{j = 1}^{k} \diff(\bm{u}_i, \bm{u}_{i,j}), ~~ i \neq j.   
\end{align}
Intuitively, in low density areas, $\bar{d}_i$ is large and the ratio $\frac{\bar{d}_i}{\bar{D}_i}$ will spread out. In high density areas, the result will be reversed. Previous studies \cite{loftsgaarden1965nonparametric, breiman1977variable} have shown that the $k$-nearest neighborhood mapping is adaptive to estimate the local sample density from the distance perspective. 

\noindent\textbf{Density based malicious factor $F(i)$.} As we mentioned in Sec.~\ref{Sec:Formulation}, the objective of the FL system is to train a joint model $\mathbf{w}$ for a distributed learning problem. Note that a typical machine learning problem usually comes with multiple output labels, e.g., MNIST which aims to recognize digits 0-9. {Therefore, it is obvious that for each $\bm{u}_i$, the updated parameters from remote clients make different contribution to the output, which could be denoted as $\mathbf{y} = [y_1, \dots, y_r, \dots y_R ]$ , where $r \in [1, R]$ denotes the $r$-th output label. Considering the relationship between $\mathcal{U}_i$ and $\mathbf{y}$ to be a probabilistic graph model, then we could use the expected probability distribution of each output label, e.g.,  $\mathbb{P}(y_r) = \mathbb{P}(y_r | \mathcal{U}_i)$ for the $r$-th output, to approach the joint distribution $\mathbb{P}({y}) = \prod_{r=1}^R \mathbb{P}(y_r)$. Therefore, though our defender still has no access to the true statistic distribution of $\bm{u}_i$, we can approach an estimated probability distribution based on the analysis of each output label that satisfies $\sum_{r=1}^{R} \mathbb{P}(\bm{u}_i | y_r) = 1$. In this paper, we use a non-parametric kernel density estimation function to approach the estimated local distribution for the $i$-th remote update $\bm{u}_i$ with its $k$-nearest neighborhood $\mathcal{U}_i$. Specifically, the estimated distribution of $i$-th client on $r$-th output label is denoted as  $\tilde{q}(\bm{u}^{r}_i)$. Formally, the process of obtaining the $\tilde{q}(\bm{u}^{r}_i)$ can be formulated by:}
\begin{equation}
\centering
\tilde{q}(\bm{u}^{r}_i) = \frac{1}{k}\sum_{j=1}^{k}K\left(\frac{\bm{u}^{r}_i - \bm{u}^{r}_{i,j}}{h^{r}}\right), 
\label{Eq:model_clusterlocal}
\end{equation}
where $K(\cdot)$ is the kernel function and $\bm{u}^{r}_{i,j}$ denotes the $r$-th output from the $j$-th neighbor in $\mathcal{U}_i$. {And $h$ is the bandwidth of $K(\cdot)$, which is a user-defined constant value that is correlated to the learning classifier. Note that theoretically, the value of $h$ could be different for each output label. For better presentation, we consider it to be constant in this paper.} Particularly, we consider the $K(\cdot)$ is a Gaussian function as follows:
\begin{equation}\label{Eq:model_gaussian}
K\left(\frac{\bm{u}^{r}_i - \bm{u}^{r}_{i,j}}{h}\right) = \frac{1}{(\sqrt{2\pi)}h}\exp\left(-\frac{|\bm{u}^{r}_i - \bm{u}^{r}_{i,j}|}{2h}\right).
\end{equation}
Based on the estimated $\tilde{q}(\bm{u}^{r}_i)$, we define the maliciousness factor $F(i)^r$ for the $i$-th remote update for $r$-th output label in its neighborhoods as follows:
\begin{equation}
F(i)^r = \frac{\sum_{j=1}^k \tilde{q}(\bm{u}_{j}^r)}{k \tilde{q}(\bm{u}_i^r)},
\label{Eq:Fir}
\end{equation}
where $\bm{u}^{r}_{j}, j \in [1, k]$ is the local neighbors of the $i$-th remote update on the $r$-th label. Therefore, for each label, we could obtain a corresponding maliciousness degree factor $F(i)^r$ for the $i$-th remote update. We then estimate the numerical maliciousness degree factor $F(i)$ of $\bm{u}_i$ with the obtained $\bm{u}_i^r$ according to each output labels. {Note that for two different outputs $r$ and $r'$, we consider the probability distribution of $\bm{u}_i^r$ and $\bm{u}_i^{r'}$ in the same neighborhoods $\mathcal{U}_i$ could satisfy the following relationship that $\mathbb{P}(\bm{u}_{i}^{r}, \bm{u}_{i}^{r'} | y_r, y_{r'}) = \mathbb{P}(\bm{u}_{i}^{r} | y_r) \cdot \mathbb{P}(\bm{u}_{i}^{r'} | y_{r'})$. Then the joint probability of $\tilde{\mathbb{P}}(\bm{u}_{i})$ is:}
\begin{equation}
\centering
\tilde{\mathbb{P}}(\bm{u}_{i})  = \prod_{r=1}^{R} \mathbb{P}(\bm{u}_{i}^{r}).
\label{Eq:model_productProb}
\end{equation}
\begin{definition}\label{The:productkernel}
	From the relationship between each label $\bm{u}_i^{r}$ and the remote update $\bm{u}_i$, we give the definition $F(i)$ from the obtained $F(i)^r$ that
	\begin{equation}\label{Eq:model_productScore}
	\begin{split}
	F(i) & = \frac{\sum_{j = 1}^{k}\tilde{q}(\bm{u}_{j})}{k \tilde{q}(\bm{u}_i ) }\\
	& =  \frac{\sum_{j=1}^{k} \prod_{r=1}^R \tilde{q}(\bm{u}_{j}^{r})}{k \tilde{q} \prod_{r=1}^R(\bm{u}_i^{r}) } = \prod_{r=1}^R F(i)^r.    
	\end{split}
	\end{equation}
\end{definition}
{\noindent\textit{Explanation.} See Appendix~$B$ in detail. \hfill~$\Box$}

Note that the obtained $F(i)$ can be considered as the same as $\frac{\bar{d}_i}{\bar{D}_i}$ from the statistical perspective: the negative exponential operator of kernel function indicates that the value of $\bar{d}$ becomes smaller when the difference between $\bar{d}$ and $\bar{D}$ grows larger. {Specifically, we consider the higher value of $F(i)$ is, the relative position of $\bm{u}_i$ is closer to the kernel center of its local neighborhood. $F(i)$ can lead to a more accurate estimation when the value of $h$ is customized for each dimension in each cluster and we set this parameter as a constant value in our paper for simplicity, as finding the optimal hyperparameter is not the goal of this paper. The process to obtain the maliciousness degree factor vector $F(i)$ for the remote client could be summarized in Algorithm~\ref{aaa}. For convenience, we assume that the output labels are already known at the beginning of LoMar defense. In summary, we use LoMar to obtain the malicious factor in the following two steps: i) developing the $k$-nearest neighborhood map; ii) obtain $\tilde{q}(\bm{u}_i^{r})$ and calculate the malicious factor vector $F(i)$}.  

\begin{algorithm}[t!]
	\caption{Algorithm for malicious factor $F(i)$ in LoMar} \label{aaa}
	\begin{algorithmic}[1]
		\REQUIRE The $i$-th remote update $\bm{u}_i$ with its neighborhoods $\mathcal{U}_i$, learning output labels $\mathbf{y} = [y_1, \dots, y_r, \dots, y_R]$.\\
		\ENSURE  Factor for the degree of local maliciousness $F(i)$.\\  
		\FOR{$r = 1, 2, \dots, R$}
		\IF{The local neighborhood $\mathcal{U}_i$ is empty}
		\STATE Build the neighborhood $\mathcal{U}_i$ and obtain the reference set $\mathcal{U}_i^{r}$ for each output label. 
		\ENDIF
		\STATE Obtain $\tilde{q}(\bm{u}_i^{r})$ from $\mathcal{U}_i^{r}$ with Eq.~\eqref{Eq:model_clusterlocal}. 
		\STATE Calculate $F(i)^r$ for $r$-th label with Eq.~\eqref{Eq:Fir}. 
		\STATE $F(i) = \prod_{r=1}^R F(i)^{r}$ from Eq.~\eqref{Eq:model_productScore}.
		\ENDFOR
		\RETURN{$F(i)$}
	\end{algorithmic}
\end{algorithm}

\subsection{Phase II: Finding Decision Threshold}
After phase I of LoMar, we obtain the numerical malicious factor vector $F(i)$ for each remote client. In order to achieve the optimal defense goal in Eq.~\eqref{Eq:aggregationrule}, LoMar develops a binary controller with a feasible threshold $\epsilon$ to turn the numerical results $F(i)$ into boolean status $\delta(i)$ to finally identify and remove malicious clients. Formally, if $F(i) \geq \epsilon$, we consider the remote client to be clean, otherwise malicious if $F(i) < \epsilon$. 

However, there are several issues in the process of finding $\epsilon$ in the poisoned FL system: i) although the updates share the similar features according to  specific classifiers, the true distribution of $\bm{u}_i$ is still unavailable; ii) the threshold $\epsilon$ can hurt the learning objective of FL if it is too strict. In order to address these issues, we provide an asymptotic threshold of $\epsilon$ for our LoMar, which aims to determine most of the malicious remote updates with a high positive malicious detection rate to protect the clean updates in FL. 

Firstly, we discuss the lower bound of $\epsilon_m$ for malicious updates in a FL system under poisoning attacks. In particular, the following theorem tells the asymptotic false alarm rate for a given $\epsilon_m$:
\begin{theorem}
	Assuming that the lower bound of the probability distribution of malicious updates is $\epsilon_m$, then the possibility of false detecting a clean update $\bm{u}_i$ as malicious would be
	\begin{equation}
	\centering
	P(F(i) > \epsilon_m) \leq \exp\left(-\frac{4\pi(\epsilon_m-1)^2(k+1)^2\bar{h}}{k(2k+\epsilon_m+1)^2V^2}\right), 
	\label{Eq:inequality}
	\end{equation}
	where $k$ is the size of its $k$-nearest neighborhoods, $V$ denotes the neighborhood kernel density distribution volume of $\bm{u}_i$ and $\bar{h}$ indicates the average bandwidth of $\bm{u}_i$ (which is considered as a constant in this paper).
\end{theorem}
\noindent\textit{Proof.} See Appendix~$C$ in detail. \hfill~$\Box$ 

Then, we consider an optimal condition that there is a trusted FL system, which consists of only clean remote clients. In this scenario, there exists a statistical boundary for the expectation of $F(i)$ for the furthest clean update $\bm{u}_i$ to the kernel center. The following theorem shows how to determine the boundary of clean remote updates.
\begin{theorem}
	Consider that the remote update $\bm{u}_i$ is generated from a continuous distribution in a FL system. Then, when $N \to \infty$, there exist a boundary $\sigma$ that $F(i) = 1$ where the remote update $\bm{u}_i$ is considered to be clean (not abnormal) as $ \sigma = \frac{\mathbb{E}(\bar{D}_i )}{\mathbb{E}(\bar{d}_i )} = 1$.
\end{theorem}
\noindent\textit{Proof.} Considering the situation that the updates are uniformly distributed around $\bm{u}_i$, then the discrete space $\mathcal{U}_i$ could be viewed as a continuous arbitrary distribution. The general case when $\mathbb{E}(\bm{u}_i) = 0$ would be:
\begin{equation}
\centering
\begin{split}
\mathbb{E}(\bar{D}_i) &= \frac{1}{k(k-1)} \mathbb{E}\left(\sum_{j=1}^{k} K(\bm{u}_i, \bm{u}_{i,j}) \right) \\
&= \frac{1}{k} \sum_{j=1}^{k} \mathbb{E}(K(\bm{u}_i, \bm{u}_{i,j}) = \mathbb{E}(\bar{d}_i),
\end{split}
\label{Eq:kde_expectation}
\end{equation}
where $\mathbb{E}(\bar{D}_i)$ is the expectation of $\bar{D}_i$. Then we give a definition of the boundary $\sigma$ as $\sigma = \frac{\mathbb{E}(\bar{D}_i )}{\mathbb{E}(\bar{d}_i )}  = 1$. \hfill~$\Box$

This shows that when the value $F(i)$ reaches $1$, we consider it locates at the boundary of clean updates in a FL system without attacks. Thus, in order to protect the FL learning objective and remove as many the malicious remote updates in LoMar, we define the optimal threshold $\epsilon = \min \{1, \epsilon_m\}$.
We introduce the development of finding an optimal $\epsilon$ and obtaining a trusted joint model $\tilde{\mathbf{w}}^{t}$ at the $t$-th iteration in LoMar at Algorithm~\ref{Algorithm:2}.

\begin{algorithm}[t!]
	\caption{Algorithm of estimating $\epsilon$ in LoMar} \label{Algorithm:2}
	\begin{algorithmic}[1]
		\REQUIRE Local malicious factor $F(i)$ and weighted factor $\alpha_{i}$ for each client in the FL system, joint model $\tilde{\mathbf{w}}^{t-1}$.\\
		\ENSURE The updated joint model $\tilde{\mathbf{w}}^t$.\\
		\STATE Find the value of $\epsilon_m$ from Eq.~\eqref{Eq:inequality}.
		\STATE Determine $\epsilon = \min \{1, \epsilon_m\}$.
		\FOR{$i \in (1, N+M)$}
		\IF{$F(i) \geq \epsilon$}
		\STATE $\delta(i) = 1$.
		\ELSE
		\STATE $\delta(i) = 0$.
		\ENDIF
		\ENDFOR
		\STATE $\tilde{\mathbf{w}}^{t} = \tilde{\mathbf{w}}^{t-1} + \sum_{i=1}^{N+M}  \delta(i) \alpha_i  \bm{u}_i^t$ from Eq.~\eqref{Eq:aggregationrule}.
		\RETURN{$\tilde{\mathbf{w}}^{t}$}
	\end{algorithmic}
\end{algorithm}

{
	\subsection{Discussion}
	{In this work, we develop a defense algorithm LoMar to address the poisoning attacks on FL systems.  Note that the training data distribution can have a significant influence on both the implementation and defense of poisoning attacks against FL. Typically, as the degree of non-i.i.d training data distribution increases, the learned local updates on the remote clients can be more diverse, which leaves room for the implementation of the attackers and makes it more difficult to develop a defense. Thus, we investigate the performance of the proposed LoMar algorithm under different training data distribution and the results indicate that compared to the existing works, LoMar is more robust under this situation, where the experimental details are introduced in Sec.~\ref{SubSec:lambda}.  }
	
	{We also investigate the performance of the proposed LoMar from the view of FL efficiency. Note that as presented in \cite{mcmahan2017communication}, different from the centralized ML frameworks, the learning speed of a FL network is dominated by the communication cost between the aggregator and remote clients. As such, though the implementation of LoMar defense requires an extra computational cost on the aggregator side, the impact on FL efficiency could be limited. Moreover, the extra cost of LoMar could be mainly from the development of the $k$-nearest neighborhood, which is proved as $\mathcal{O}(N)$ in \cite{tang2017local}. Typically, in the settings of a FL network, the computational power on the aggregator is much powerful than remote clients that we believe the linear cost on the aggregator is also limited.}
	
	Note that several studies have a close proposal of this paper, which develop defense approaches to remove malicious updates against poisoning attacks on FL. For example, works in \cite{shen2016auror} provide Auror, which studies the statistical distribution of remote updates for malicious detection. Meanwhile, \cite{li2020learning} provides spectral anomaly detection (SAP) to detect the malicious updates for developing a robust FL system. We agree with Auror that though the defender has no access to the training data distribution because of the privacy features, we could approximate it via the observation and investigation to the parameters of the remote updates during the training of the joint model. 
	
	However, similar to the existing centralized ML defense approaches, Auror and SAP assume that the defender has access to a public trusted dataset, which shares the same distribution with the clean training data. On the contrary, our LoMar uses a non-parametric estimation method KDE, for studying the statistical features without knowing the clean training data distribution, which preserves the important privacy feature of FL. Additionally, the works in \cite{tolpegin2020data} investigate the label-flipping attack on FL and provide a PCA analysis-based defense strategy for malicious detection. However, this approach assumes that the malicious and clean updates could be easily divided into two clusters, which is still a global anomaly detection approach that could be easily cheated by model poisoning attacks with a stealth metric \cite{bhagoji2019analyzing}. Moreover, we provide the performance evaluation of LoMar against model poisoning attacks in  Appendix~$A$. In conclusion, compared to recent defense studies on detecting malicious remote clients in FL, our LoMar approach is able to address the poisoning attacker with stealth metric and develop the defense model under the privacy-preserved FL framework. 
}

\section{Evaluation} \label{Sec:Evaluation}
\subsection{Experiment Setup}
To evaluate LoMar, we conduct extensive experiments under the FL framework on multiple real-world datasets. The experiments are performed with Pytorch \cite{paszke2017automatic} and we implement the FL framework with the Python threading library by designing remote clients as lightweight threads. 

\subsubsection{Datasets}\label{SubSec:Data}
{We first introduce four real-world datasets used in this paper. Especially, to present the imbalanced number of data samples according to different output labels, we assume a reasonable scenario that the training data samples in each remote client could be divided into two parts: one subset of a major label with the most number of training samples and one subset of all output labels with an equal number of samples. Specifically, we use a hyperparameter $\lambda \in [0, 1)$ to represent the ratio of major label training samples in the remote client private dataset.}

\textbf{Real-world datasets.} We consider four popular datasets in the FL field: MNIST \cite{lecun1998gradient}, KDDCup99 (network intrusion patterns classifier) \cite{bay2000uci}, Amazon (Amazon product reviews) \cite{asuncion2007uci}, and VGGFace2 (facial recognition problem from Google image search) \cite{Cao18}. The general information of each dataset is introduced in Table~\ref{table:dataset}, which includes the size of the dataset and the number of the output labels.

\textbf{Data partitioning settings.} In the process of the training data partition, we introduce the hyperparameter $\lambda$ to control the imbalanced ratio of each remote client. For instance, if the $i$-th remote client has $l_i$ private training samples, then $(1-\lambda)l_i$ of the samples are randomly selected from the whole training dataset and the other $\lambda l_i$ samples are chosen from one specific label.  For MNIST, we partition the dataset into $1000$ remote clients and each remote client has $600$ training data samples. Because the KDD dataset has a low number of features and the data samples in each label are imbalanced: some labels only have $20$ data samples while others can have over $280,000$ data samples. We consider each clean client shares the same number of mixed labels and we set the number of the remote clients to $23$. For Amazon dataset, because the number of the features is extremely high and the number of data samples per class is very low, we set the number of the remote clients to $50$ and each client has $150$ data samples. And for the VGGFace2 dataset, for simplicity, we use the pre-processed training data before the FL training process (each training image is resized to $256 \times 256$). The number of remote clients is set to $200$ and the partitioning of the training data follows the same rule in MNIST. 

\begin{table}
	\centering
	\caption{Dataset information overview.}
	\begin{tabular}{|c|c|c|c|}
		\hline 
		\bf{Dataset}& \bf{Dataset Size} & \bf{Classes} & \bf{Features} \\
		\hline 
		\bf{MNIST}& $70,000$ & $10$ &  $784$ \\
		\hline
		\bf{KDDCup99}& $494,020$ & $23$  &  $41$  \\
		\hline
		\bf{VGGFace2 }& $7380$ & $10$  &  $150528$ \\
		\hline
		\bf{Amazon}& $1500$ & $50$ &  $10000$ \\
		\hline
	\end{tabular} 
	\label{table:dataset}
\end{table}

\subsubsection{Implementation Settings}
Note that our comprehensive experiments are developed to evaluate the performance of the defense algorithms instead of finding the best learning model with the highest FL learning performance. Based on this motivation, we implement our learning model and we define the default experimental setting of the FL framework as follows
\begin{itemize}
	\item The initialized joint model is set as $\mathbf{w}^0 = 0 $.
	\item The number of FL iteration time is set as $T = 200$.
	\item For each remote training process, the number of remote training is set as $E = 5$. 
	\item The batch size for each remote training SGD is $20$.
	\item For MNIST, KDD and Amazon datasets, which are with less information in each training data sample, we use a logistic regression model with one fully-connected layer. For VGGFace2, we implement a DNN classifier SqueezeNet with the torchvision package \cite{marcel2010torchvision}. The detail architecture of our SqueezeNet classifier is described in Table.~\ref{table:DNN_structure}. 
\end{itemize}

\begin{table}
	\centering
	\caption{SqueezeNet model setting.}
	\begin{tabular}{|c||c|c|c|}
		\hline
		{Layer Type}& {Size} & {Parameters} & {Value}\\
		\hline 
		{Conv + ReLU}& $3 \times 3 \times 64$ & {Layers} & $12$\\
		\hline
		{Max Pooling}& $3 \times 3$  & {Batch size} & $8$\\
		\hline
		{Conv + ReLU }& $7380$ & {Momentum} & $0.9$\\
		\hline
		{Max Pooling}& $3 \times 3$ & {Weight decay} & $0.0001$ \\
		\hline
		{Conv Kernel }& $10 \times 512$ & {Learning rate} & $0.001$\\
		\hline
		{Output}& $10$ & ~ & ~\\
		\hline
	\end{tabular}
	\label{table:DNN_structure}
\end{table}

\subsubsection{Compared Defense Algorithms}
We compare our LoMar defense algorithm with the following existing defense methods:

\textit{1.} \textbf{Krum} \cite{blanchard2017machine} is developed to protect distributed learning models by giving an alternative aggregation rule. At each iteration, this defense computes the Euclidean distance between each updates and removing the malicious updates from the aggregator which has a larger distance. The number of clients removed by the aggregator in our paper is set to $M$ as the number of selected clients is $N- 0.5 \times M -2$. 

\textit{2.} \textbf{FoolsGold(FG)} \cite{fung2018mitigating} is a defense method against sybil attacks \cite{douceur2002sybil}. It addresses the attacker by penalizing the learning rate of the malicious updates, which are detected by evaluating the angle difference between each update and the joint learning model. We claim that the memory usage in the FG paper violates the privacy preserving rule of the FL system and we only take the no-memory version of FG in this paper. We set the inverse sigmoid function in the FG method centers at $0.5$ and the confidence parameter to $1$.  

\textit{3.} \textbf{Median} defense method \cite{yin2018byzantine} uses an aggregator which sorts the updates value of this parameter among all $N+M$ clients and picks the median value as the contribution to the joint model at each iteration. Note that when $N+M$ is odd, the median value comes from one update, and when $N+M$ is even, it comes from the average of two updates. 

\begin{table*}
	\centering
	\caption{Testing accuracy under the label-flipping attack: $\lambda = 0, \tau = 0.1$. }
	\begin{tabular}{|c||c|c|c||c|c|c||c|c|c|}
		\hline
		~ &   \multicolumn{3}{c||}{\bf{MNIST}}  & \multicolumn{3}{c||}{\bf{KDD}} & \multicolumn{3}{c|}{\bf{Amazon}} \\
		\hline
		~ & Overall & Target & Other & Overall & Target & Other & Overall & Target & Other \\
		\hline
		LoMar & $\underline{\textbf{0.912}}$ & \textbf{0.977} & $\underline{\textbf{0.930}}$ & $\underline{\textbf{0.971}}$ & $\underline{\textbf{0.991}}$ & $\underline{\textbf{0.990}}$ & $\underline{\textbf{0.970}}$ & $\underline{\textbf{0.988}}$ & $\underline{\textbf{0.989}}$  \\
		\hline
		FG & 0.783 & 0.911 & 0.512 & $\underline{0.901}$ & $\underline{0.989}$ & $\underline{0.965}$ & \textbf{0.922} & \textbf{0.971} & \textbf{0.970}  \\
		\hline
		Krum  & \textbf{0.883} & 0.051 & $\underline{0.815}$ & 0.692 & 0.022 & \textbf{0.970} & 0.823 & 0.084 & 0.915  \\
		\hline
		FG + Krum & $\underline{0.812}$ & $\underline{\textbf{0.996}}$ & \textbf{0.880} & \textbf{0.957} & \textbf{0.985} & 0.918  & $\underline{0.901}$ & $\underline{0.960}$ & $\underline{0.917}$  \\
		\hline
		Median  & 0.655 & $\underline{0.925}$ &  0.610 & 0.555 & 0.971 &  0.708 & 0.572 & 0.959 &  0.500  \\
		\hline
		No Defense & {0.884} & 0.078 & 0.925 & 0.800 & 0.015 & 0.998 & 0.900 & 0.015 & 0.998  \\
		\hline
		No Attack & 0.931 & 0.982 & 0.946 & 0.980 & 0.996 & 0.998  & 0.990 & 0.994 & 0.996  \\
		\hline
	\end{tabular}
	\label{tab:multi-data}
\end{table*}

\subsubsection{Poisoning Attacks}
In this paper, we consider two types of poisoning attacks on the FL system: the data poisoning attack (which generates the poisoned remote updates by the malicious data) and the model poisoning attack (which creates the remote malicious update based on the FL aggregating rule and the attack objective). For the data poisoning attack, we implement the Label flipping attack \cite{biggio2012poisoning}. And for the model poisoning attack, we implement the stealthy model poisoning method in \cite{bhagoji2019analyzing}. Note that in this paper, the attack methods we select are the most representative and provide a general attacking formulation for different defense algorithms in each category. For instance, \cite{fang2020local} provides a typical local model poisoning attack against Byzantine defense mechanisms, however, it only focuses on the Byzantine related algorithms while not providing a successful attack on other defense strategies. We will introduce the experimental setting and the results of \textit{Model poisoning attack} in Appendix~$A$.

\textbf{Label-flipping attack parameter setting.}
The goal of the label flipping attack is to take control of the target label(class) in the FL system. Typically, the attacker address the target label by flip the output of target data samples from a source label. In our paper, we consider the malicious and the clean remote clients share the same number of training data samples. Specifically, we introduce a parameter $\tau \in [0.1,1)$ to control the malicious ratio in our experiments. Similar to the definition of $\lambda$, for the $i$-th remote client, the number of randomly selected training data samples is $(1-\tau)l_i$ and the number of label-flipped data samples is $\tau$. We then introduce our default attack implementation on the FL system corresponding to each dataset
\begin{itemize}
	\item \textbf{MNIST.} For the MNIST dataset, the attacker injects $100$ malicious clients into FL system. The source label is set to digit $7$ and the target label is set to digit $1$.
	\item \textbf{KDD.} For the KDD dataset, we pick two classes with more than $200,000$ data samples as the source label and target label, and create $3$ malicious clients whose poisoned samples are flipped from class $11$ to $9$.
	\item \textbf{Amazon.} For the Amazon dataset, we set the source label to $15$ and the target label to $10$. The number of malicious clients is set to $5$. 
	\item \textbf{VGGFace2.} For the VGGFace2 dataset, we pick up two target labels from two different source labels as label $3$ to label $2$ and label $10$ to label $9$. For each target label we create $10$ malicious clients and the total number of malicious client is $20$. 
\end{itemize}

\begin{figure*}[t!]
	\centering
	\begin{subfigure}{0.63\columnwidth}
		\centering
		\includegraphics[width = 1\columnwidth]{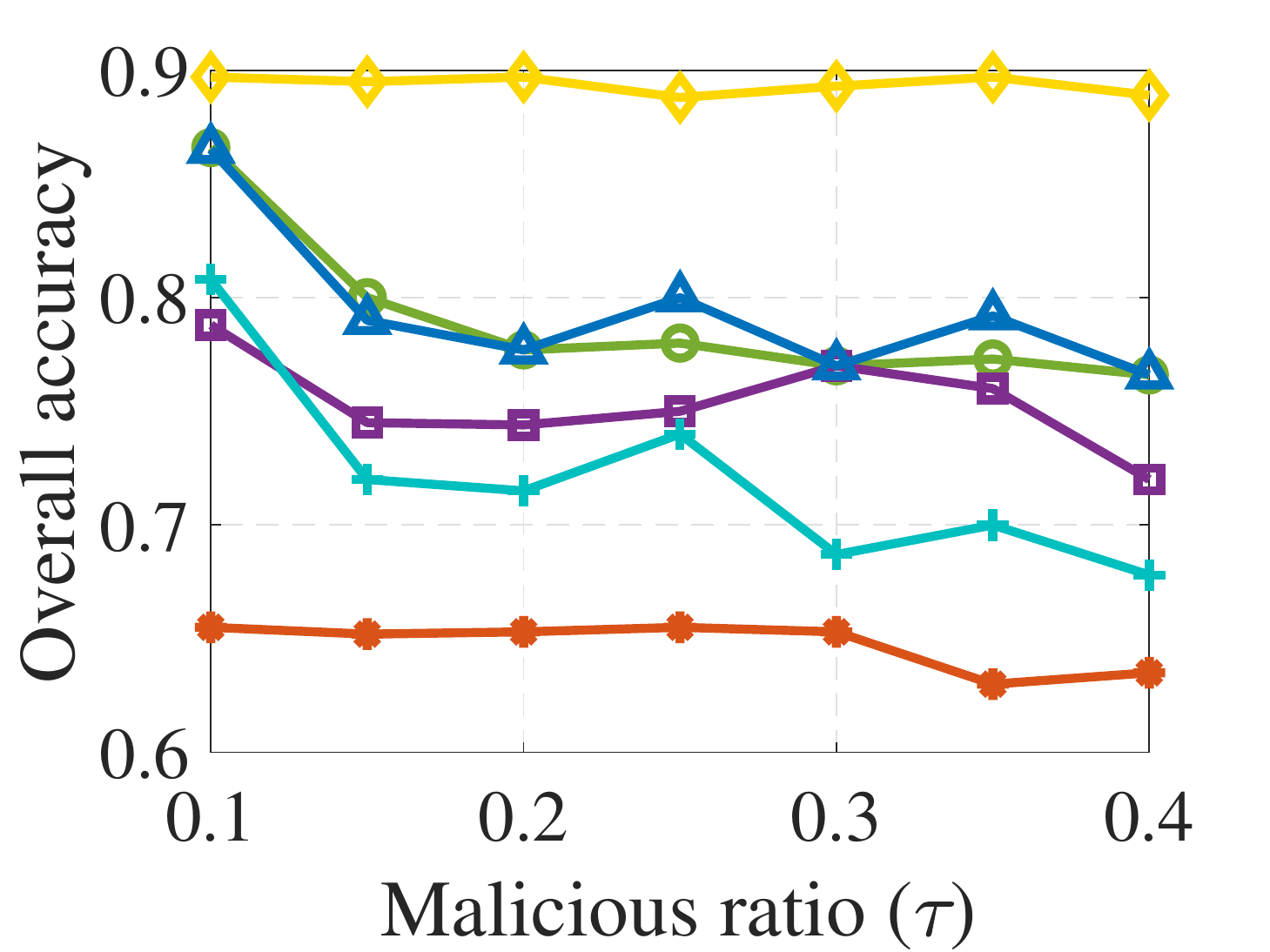}
	\end{subfigure}
	\hfill
	\begin{subfigure}{0.63\columnwidth}
		\includegraphics[width = 1\columnwidth]{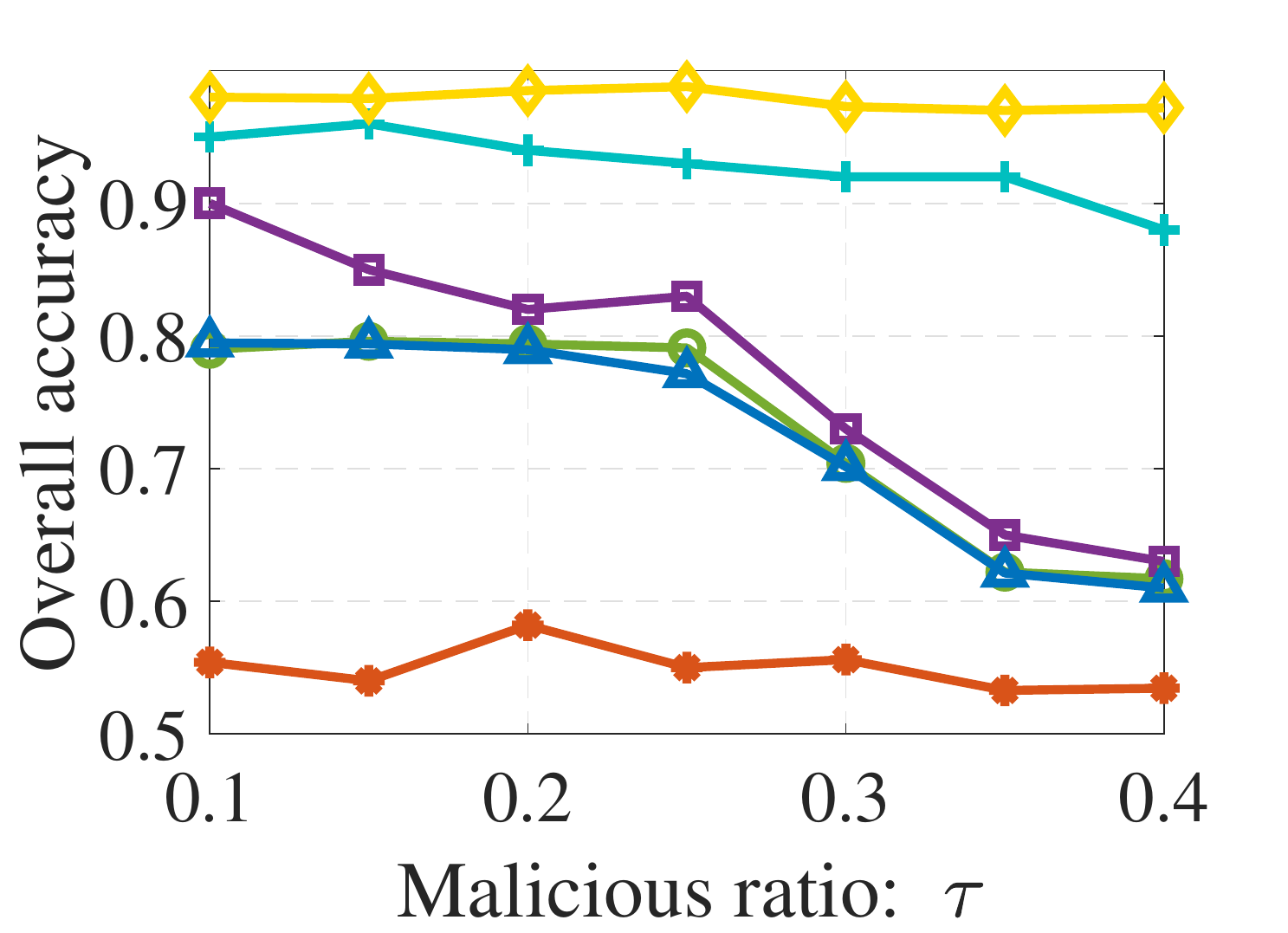}
	\end{subfigure}
	\hfill
	\begin{subfigure}{0.63\columnwidth}
		\includegraphics[width = 1\columnwidth]{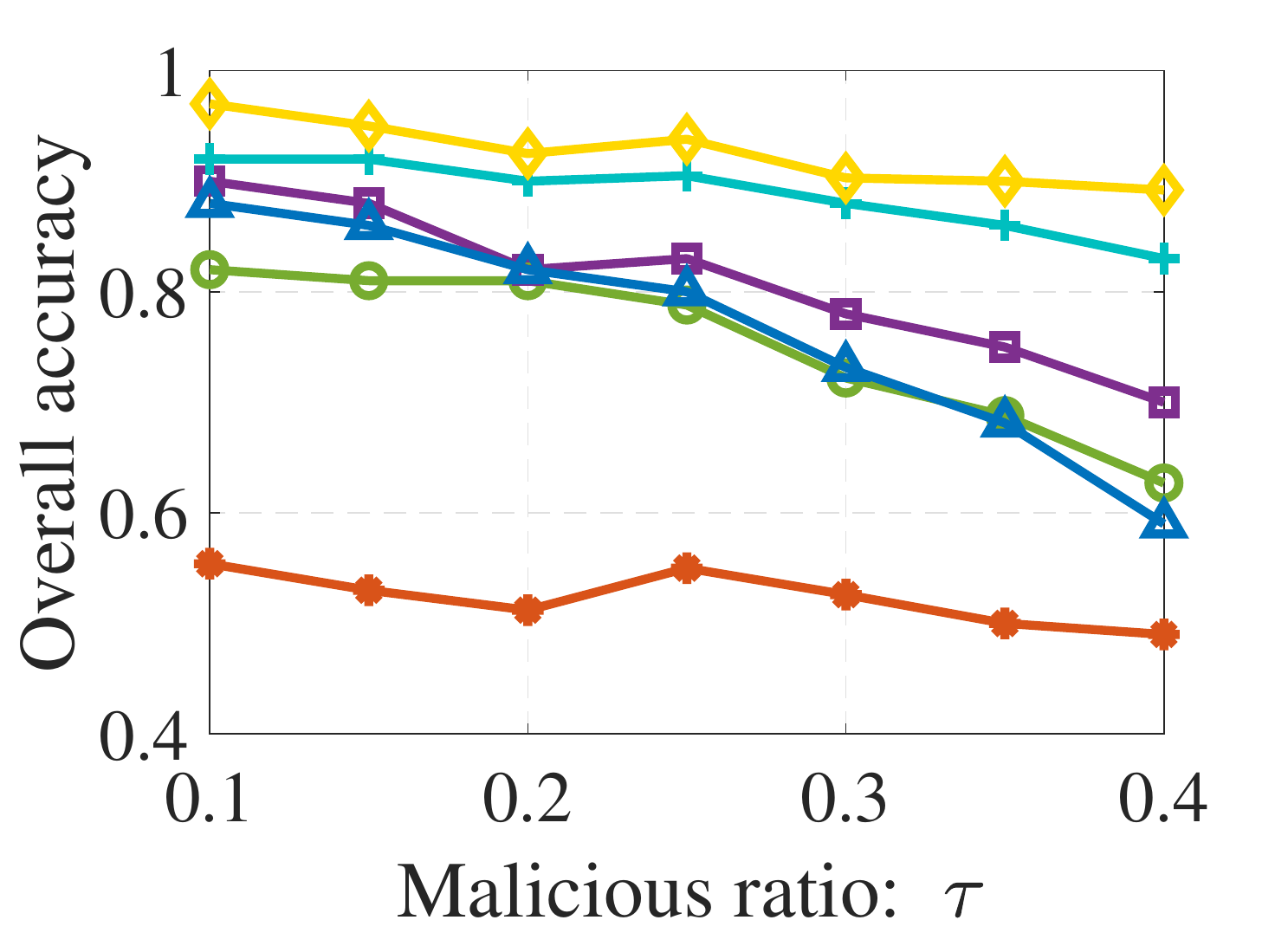}
	\end{subfigure}
	\vskip\baselineskip
	\begin{subfigure}{0.63\columnwidth}
		\centering
		\includegraphics[width = 1\columnwidth]{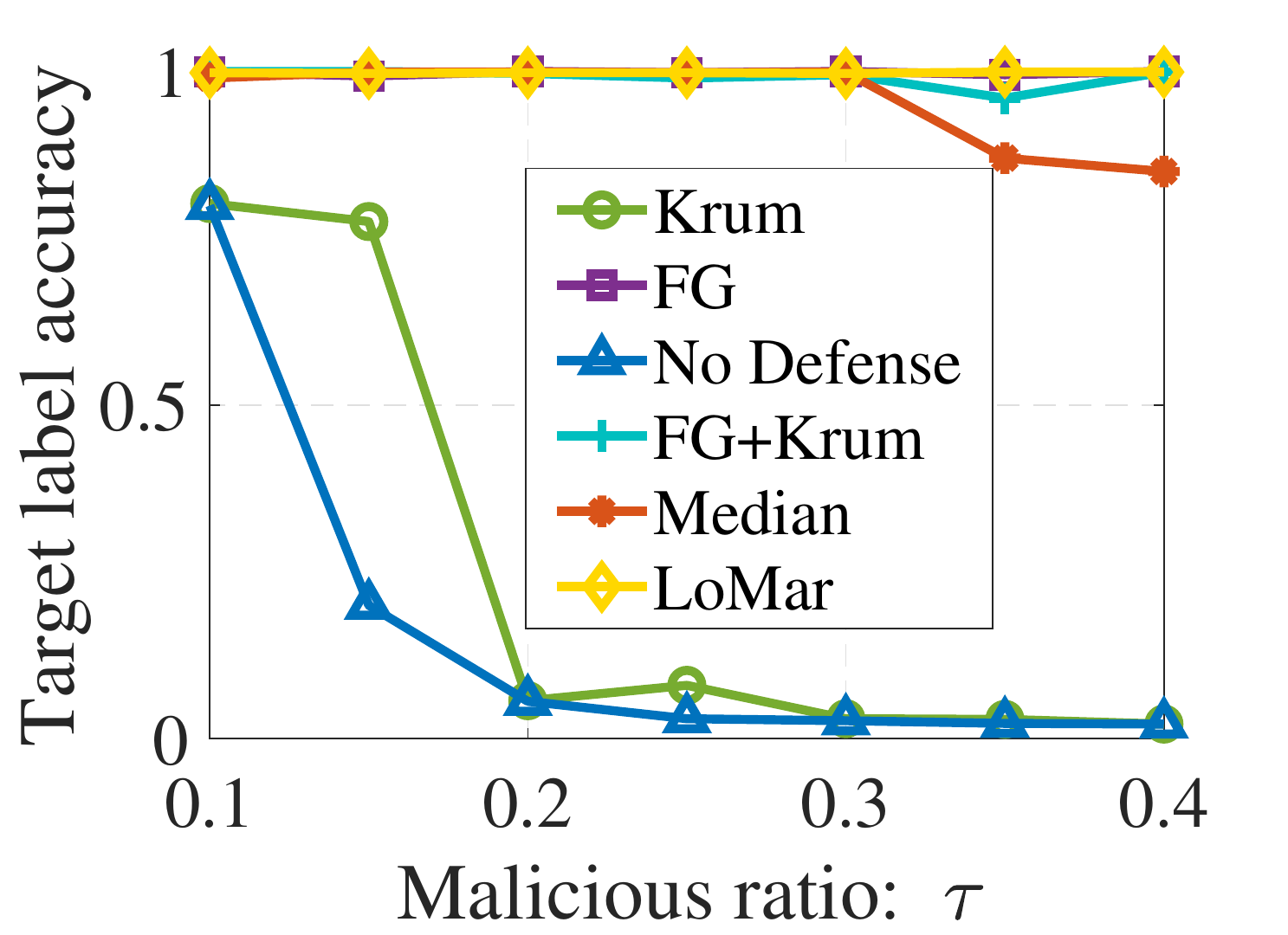}
		\caption{MNIST}
		\label{fig:target_mnist}
	\end{subfigure}
	\hfill
	\begin{subfigure}{0.63\columnwidth}
		\includegraphics[width = 1\columnwidth]{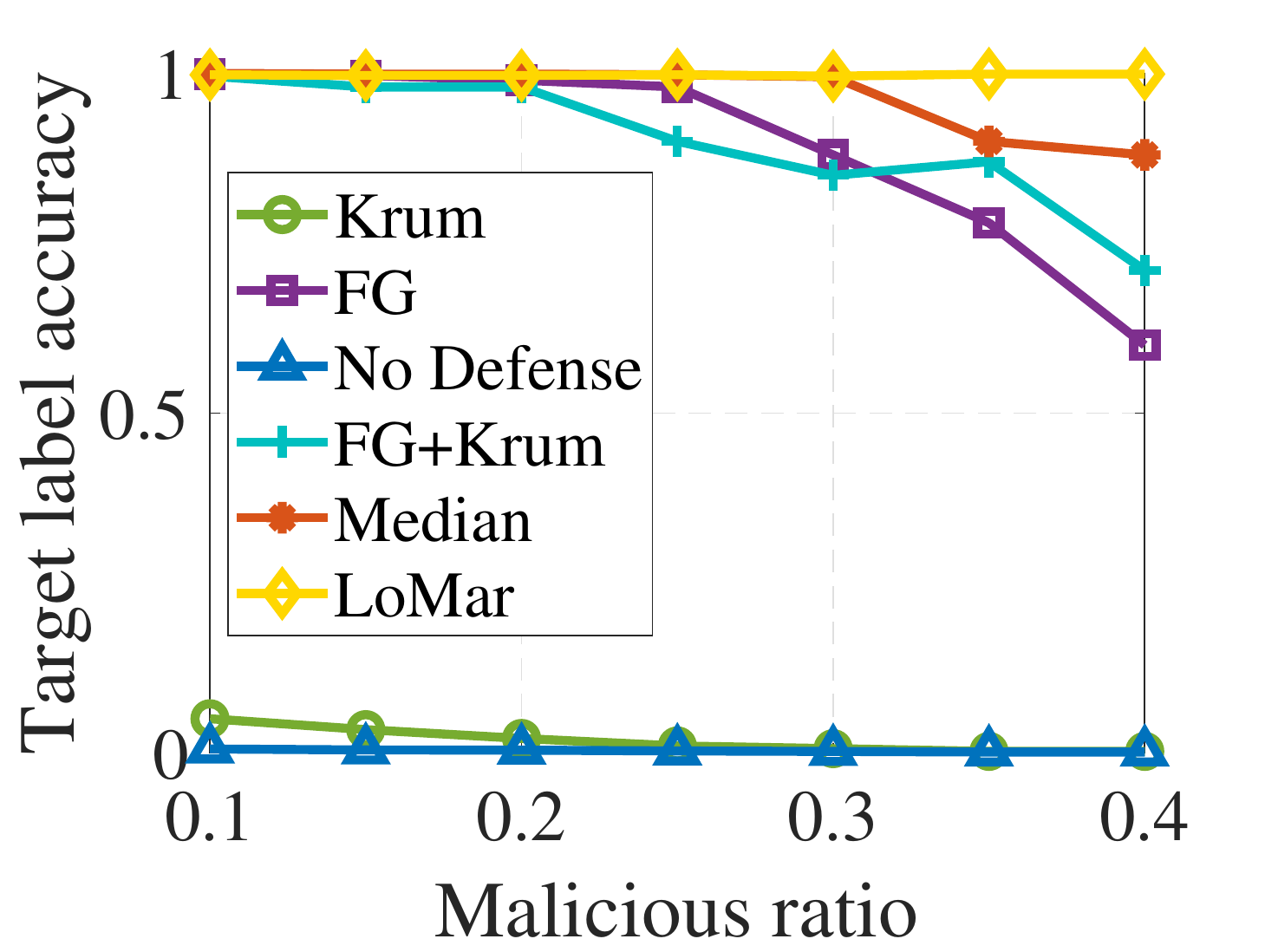}
		\caption{KDD}
		\label{fig:target_kdd}
	\end{subfigure}
	\hfill
	\begin{subfigure}{0.63\columnwidth}
		\includegraphics[width = 1\columnwidth]{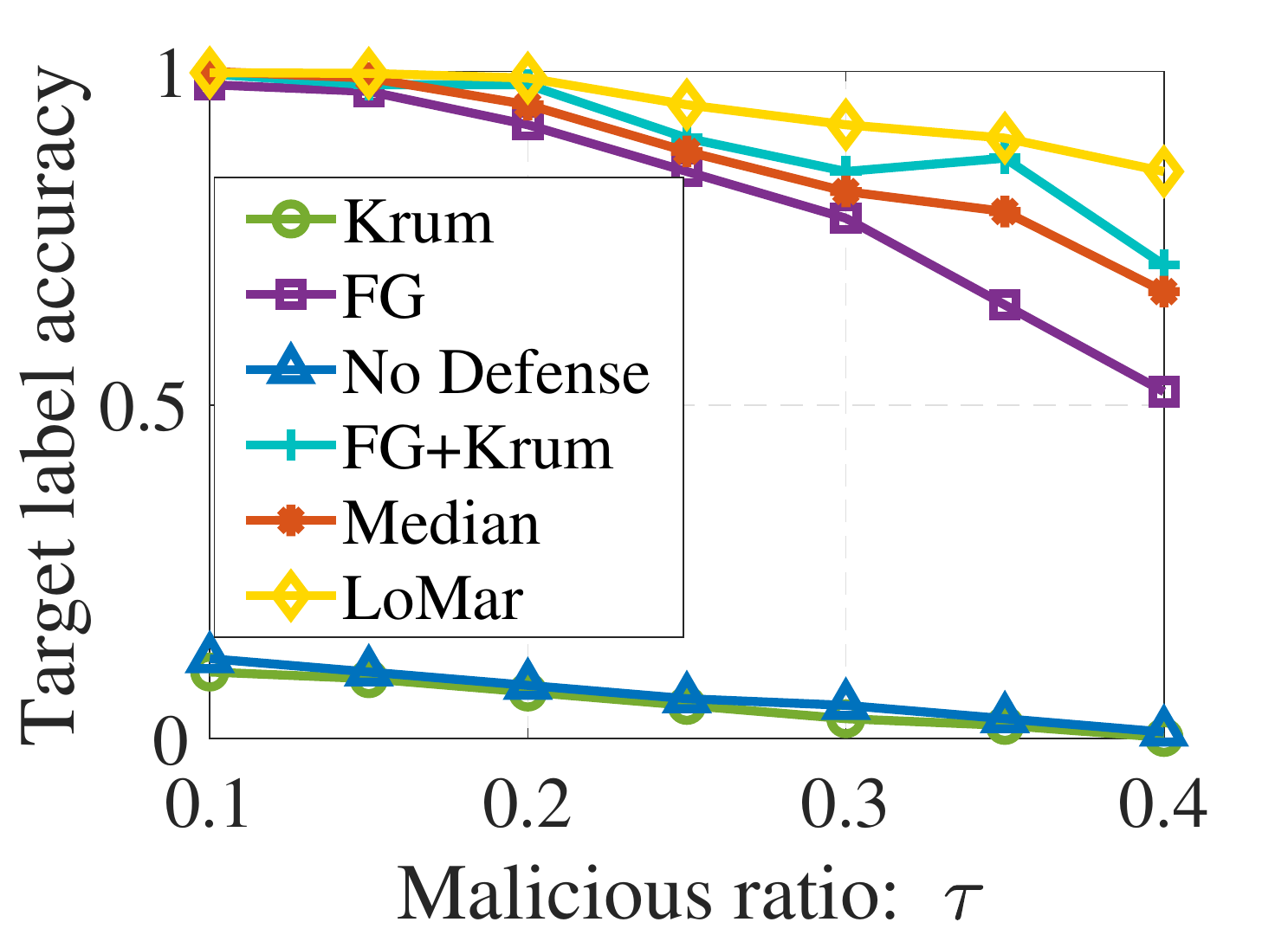}
		\caption{Amazon}
		\label{fig:target_amazon}
	\end{subfigure}
	\caption{Overall and target label testing accuracy for the compared algorithms under different malicious ratio $\tau = [0.1, 0.4]$ and $\lambda=0$ on MNIST, KDDCup99 and Amazon dataset.}
\end{figure*}

\begin{figure*}[ht]
	\centering
	\begin{subfigure}{0.63\columnwidth}
		\includegraphics[width = 1\columnwidth]{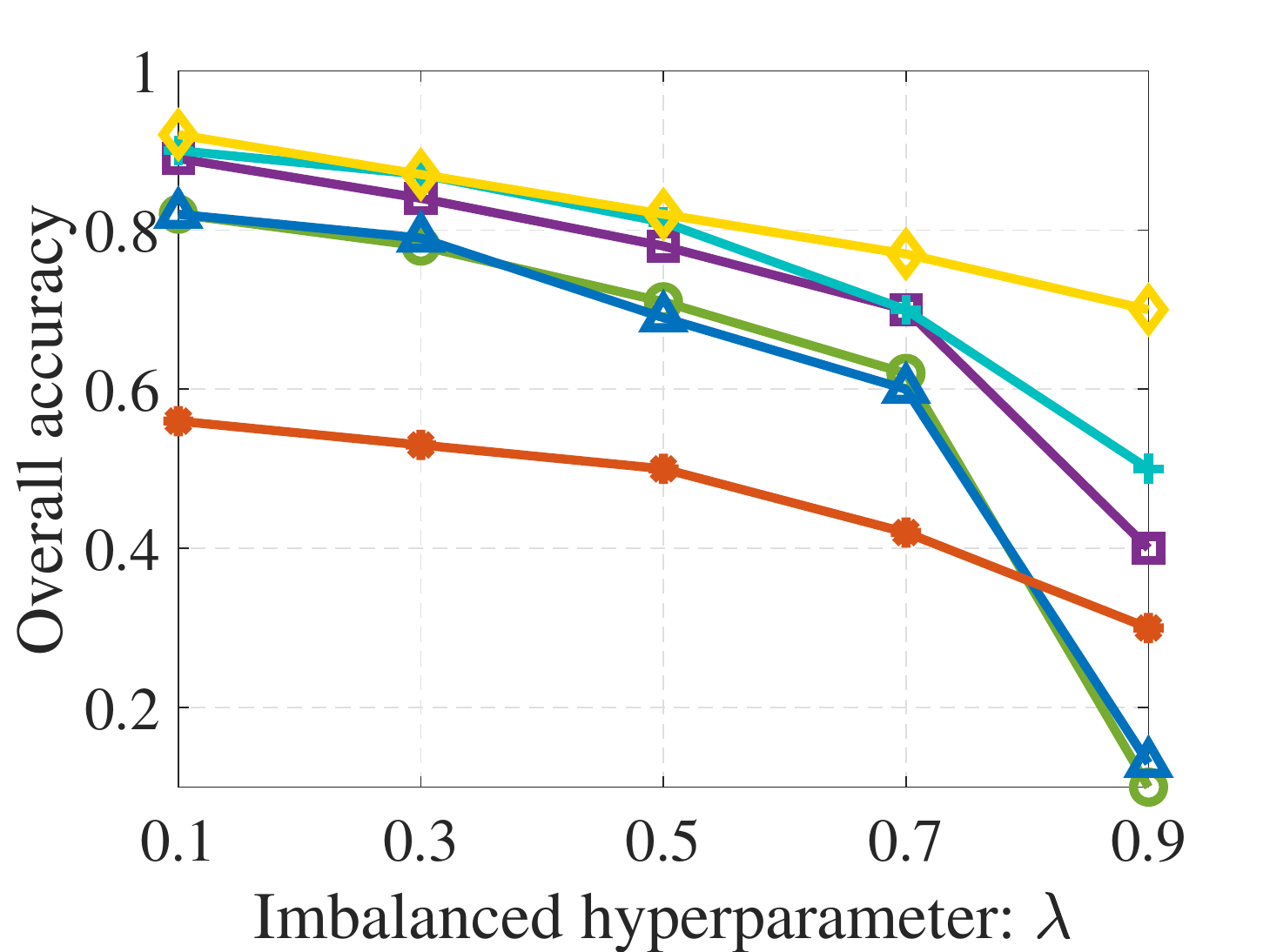}
	\end{subfigure}
	\hfill
	\begin{subfigure}{0.63\columnwidth}
		\includegraphics[width = 1\columnwidth]{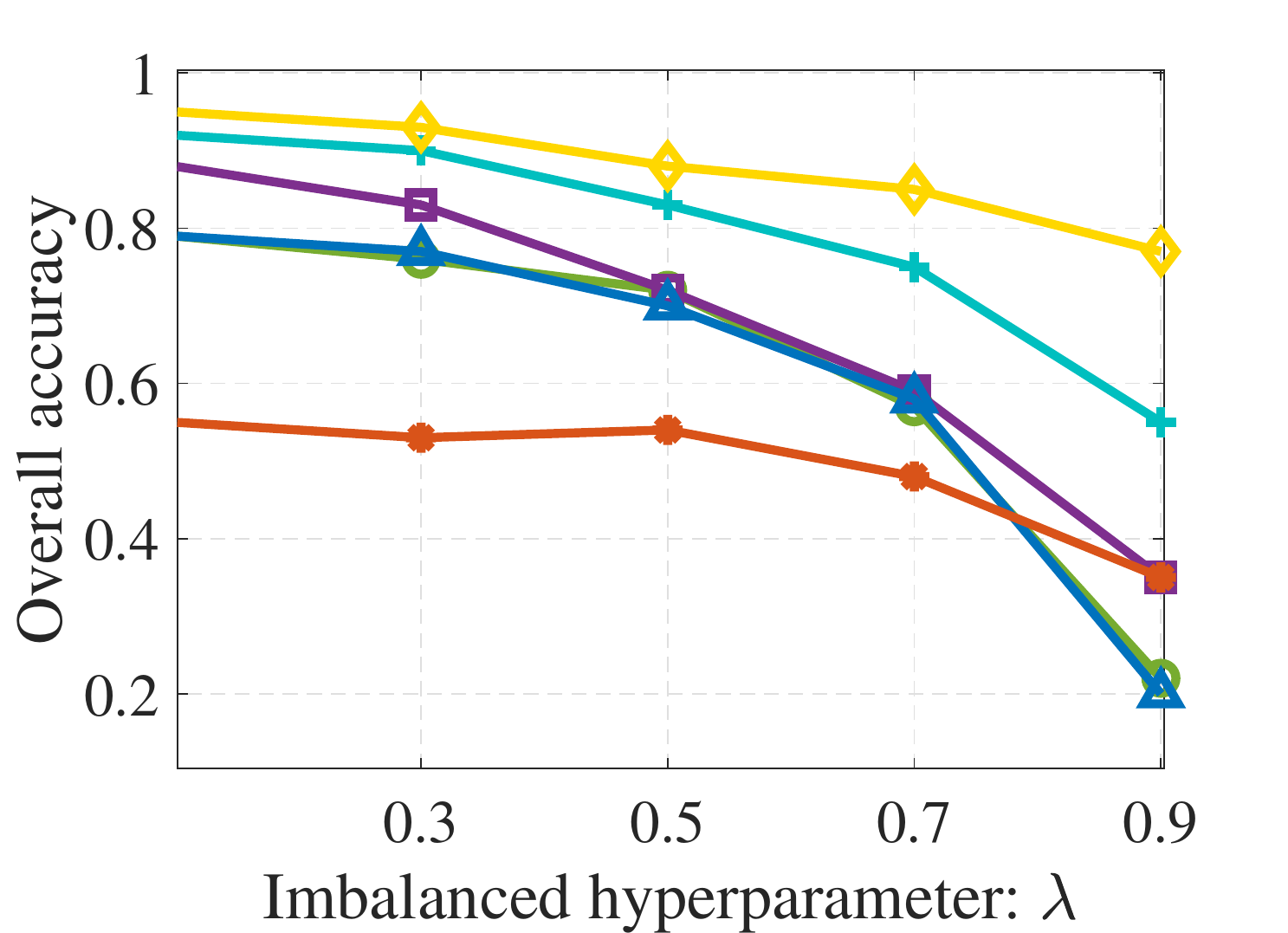}
	\end{subfigure}
	\hfill
	\begin{subfigure}{0.63\columnwidth}
		\includegraphics[width = 1\columnwidth]{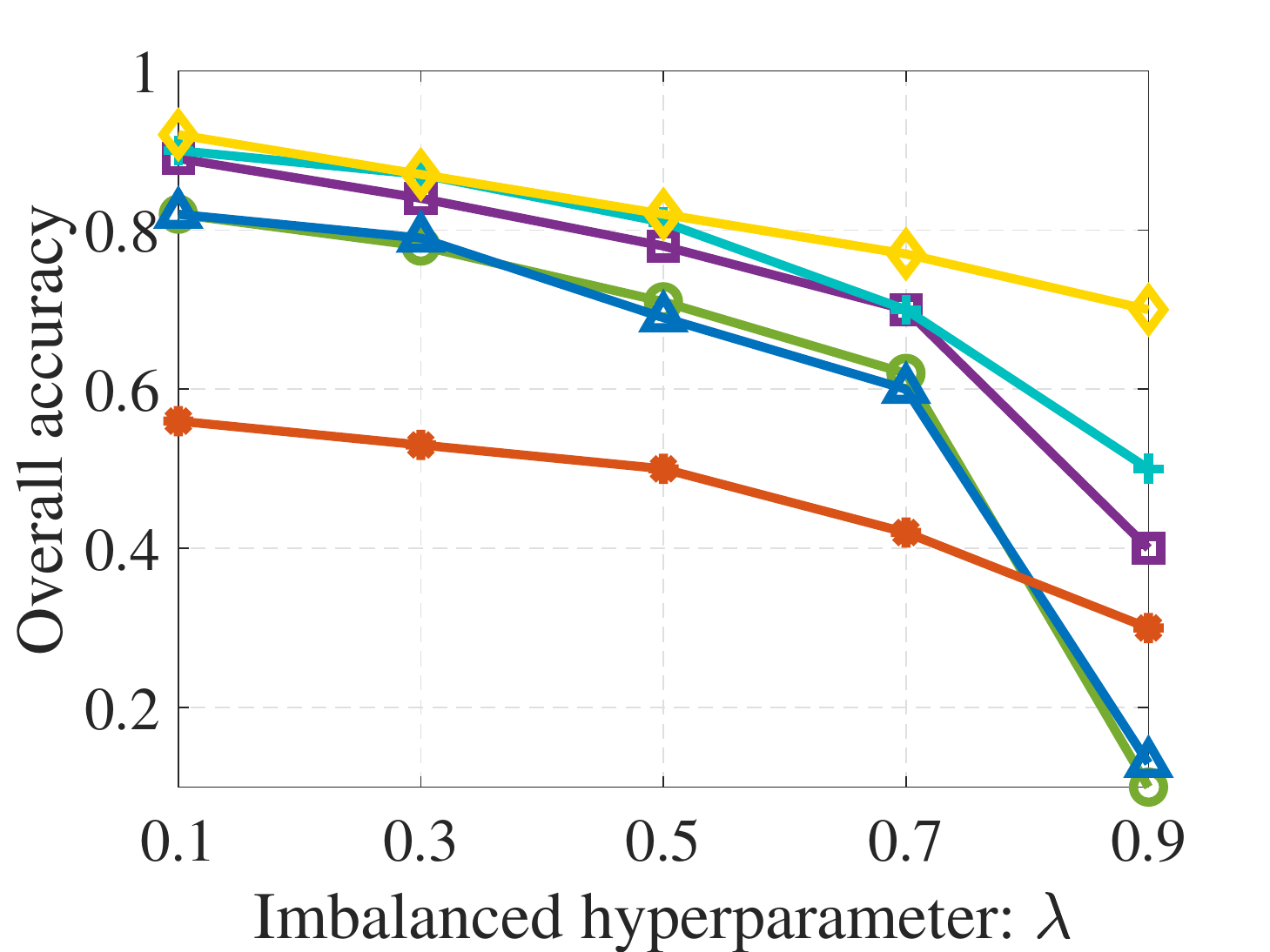}
	\end{subfigure}
	\vskip\baselineskip
	\begin{subfigure}{0.63\columnwidth}
		\includegraphics[width = 1\columnwidth]{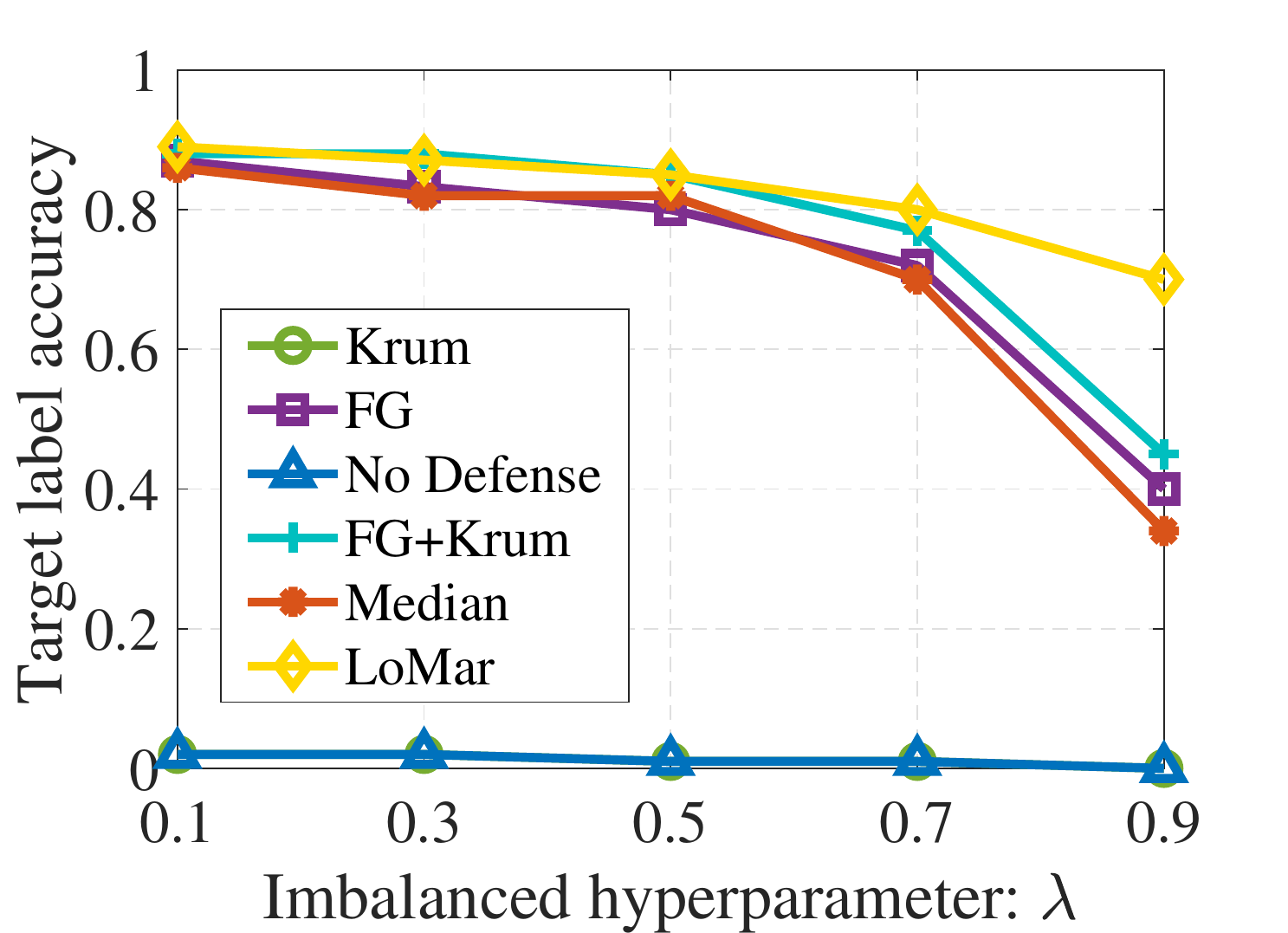}
		\caption{MNIST}
		\label{fig:noniidmnist_target}
	\end{subfigure}
	\hfill
	\begin{subfigure}{0.63\columnwidth}
		\includegraphics[width = 1\columnwidth]{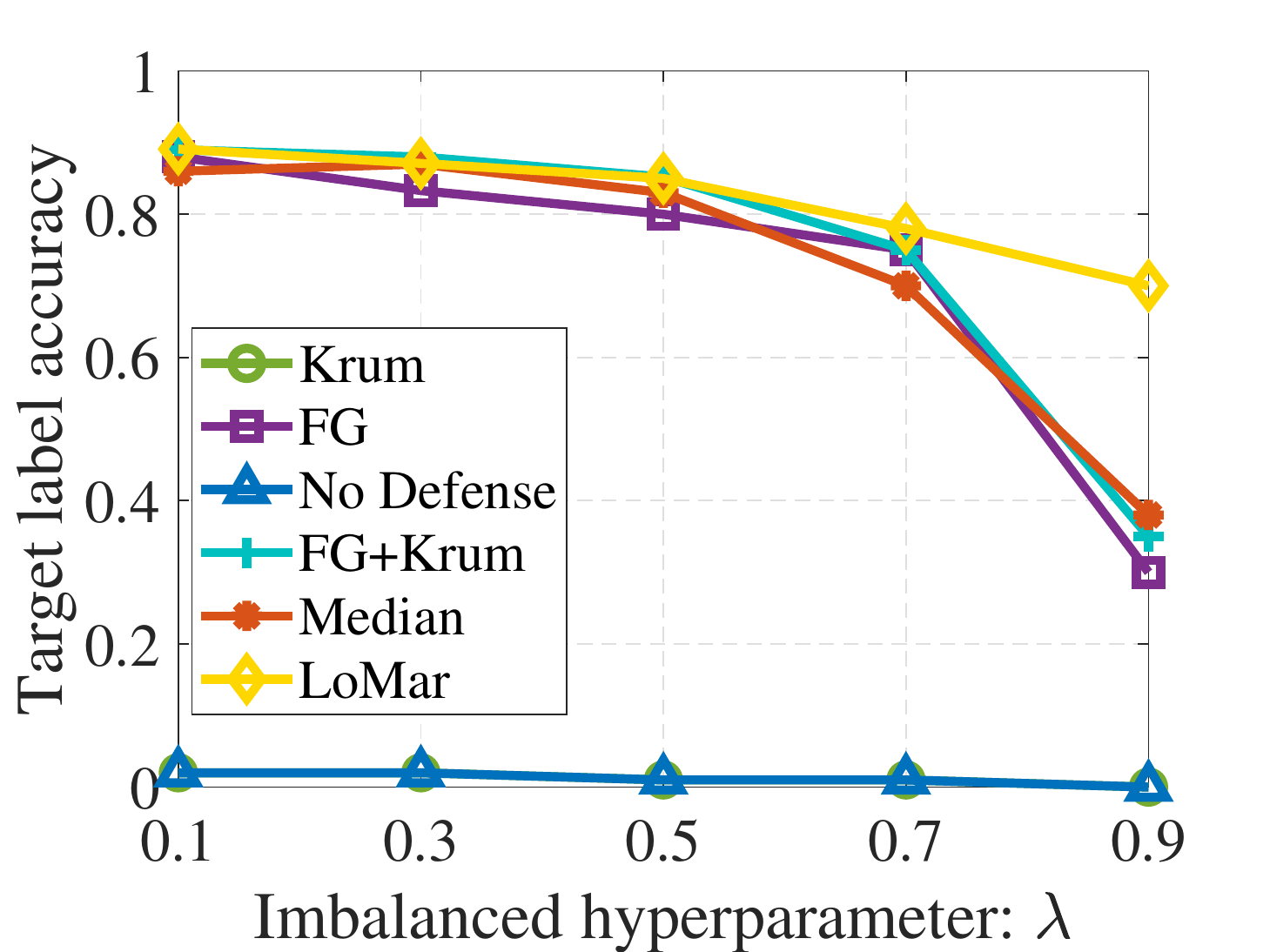}
		\caption{KDD}
		\label{fig:noniidkdd_target}
	\end{subfigure}
	\hfill
	\begin{subfigure}{0.63\columnwidth}
		\includegraphics[width = 1\columnwidth]{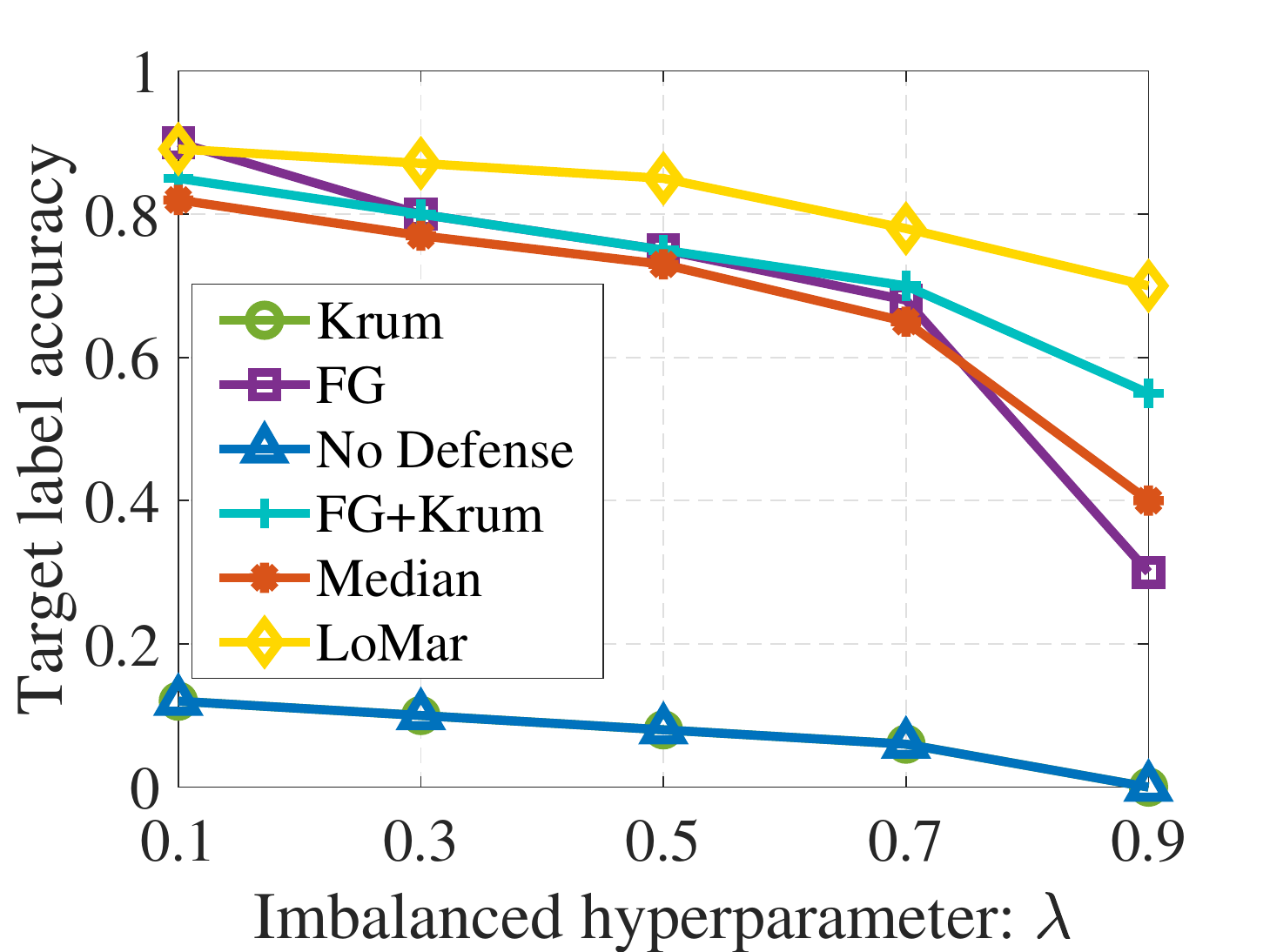}
		\caption{Amazon}
		\label{fig:noniidamazon_target}
	\end{subfigure}
	\caption{Overall and target label testing accuracy for the compared algorithms under different imbalanced hyperparameters $\lambda = [0.1, 0.9]$ and $\tau=0.1$ on MNIST, KDDCup99 and Amazon datasets.}
\end{figure*}

\subsubsection{Evaluation Metrics.} 
In order to evaluate the experiment results accurately, we introduce different evaluation metrics
\begin{itemize}
	\item \textbf{Target label accuracy.} We use the target label accuracy  to represent the testing performance of the target labels, e.g., a defense algorithm can be considered to be failed if the target label accuracy is low. 
	\item \textbf{Other label accuracy.} The other label accuracy indicates the learning result on other labels except the target label {and the source label. For example, in label flipping $1-7$ attack on MNIST, $1$ is the source label while $7$ is the target label}. If the testing results of other labels are low, we consider the FL system might be harmed by the defender because those labels are not related to the attacker. 
	\item \textbf{Overall learning accuracy.} We consider the overall learning accuracy denotes the average learning accuracy of all labels in the FL system and it can show the overall learning performance for compared defense methods under poisoning attacks. {Note that the results of overall accuracy could be different to the summation of target and overall arruracy, because the performance of the source label could be different due to the attack and defense strategies.}
	\item \textbf{Malicious alarm confidence.} We also use the receiver-operating characteristic (ROC) \cite{delong1988comparing} to evaluate the malicious alarm confidence of the compared defense algorithms. This metric is introduced in \ref{SubSec:ROC}.
\end{itemize}    
\subsection{Results and Analysis} \label{Section:analysis}
We show the results of our experiments with tables and figures in this section, and we provide the detailed analysis of the results based on our evaluation metrics. In order to present the results in the tables clearly, we mark the top-$3$ compared defense algorithms (except the no attack case) according to each evaluation metric: the $1$-st is marked as \textbf{bold} and $\underline{underline}$; the $2$-nd is marked as \textbf{bold}; and the $3$-rd is marked as $\underline{underline}$.

\subsubsection{Analysis of Compared Algorithms for Each Dataset}
We give the comparison between LoMar and other defense methods based on three datasets in Table.~\ref{tab:multi-data}. We can notice that our LoMar defense model has the best defense performance among all compared methods on overall accuracy, target label accuracy and source label accuracy in MNIST and Amazon datasets, for example, LoMar can achieve $97.0\%$, $98.8\%$ and $98.9\%$ in Amazon. The result of KDDCup$99$ dataset shows that LoMar has the second best learning accuracy on average of all three features, $97.1\%$, $99.1\%$ and $99.0\%$.  
By taking both attack rate and overall accuracy into consideration, LoMar has the best defense performance among these defense methods when the percentage of malicious updates increases.

The result of Krum on the three datasets shows that it is similar to no defense scenario so that we can consider Krum cannot defend label-flipping attack sufficiently. FG has the better performance of the target label, but worst for overall and other clean labels. It indicates that FG can influence the clean labels with false malicious detection on clean updates,  e.g., $78.3\%$, $91.1\%$ and $51.2\%$ for MNIST. Median has the worst performance of the overall and other label accuracy. Although in all three datasets, it performs sufficiently of the target label, $92.5\%$, $97.1\%$  and $95.9\%$. FG+Krum has the best defense performance among all three existing defense methods. However, it can still remove the clean updates from FL by mistake, especially in MNIST dataset as the overall learning accuracy is only $81.2\%$. 

\subsubsection{Analysis of The Different Malicious Ratio $\tau$}
Under different malicious ratio $\tau$, we show the performance with the compared defense methods on MNIST dataset in Fig.~\ref{fig:target_mnist}. We can see that LoMar has the best performance on overall and target label under the different values of $\tau$. Even with the number of malicious increasing, LoMar keeps the accuracy as $90\%$ overall and $98\%$ target label. FG and FG+Krum perform sufficiently on the targeted label, but their overall accuracy are worse than no defense scenario. Median defense takes the worst performance as overall accuracy $65\%$. For Krum, we could find out that the target label accuracy decreases with the increase of $\tau$, even to $2\%$ (similar to no defense). 

For KDD in Fig.\ref{fig:target_kdd}, LoMar has highest overall and target testing accuracy $98\%$ and $99\%$. Median also performs worst for the whole testing $55\%$. The overall testing accuracy of other defense methods decreases obviously, when $\tau$ increase, FG defense decreases from $90\%$ to $63\%$ with malicious ratio increases from $0.1$ to $0.4$. Krum defense cannot defend the target label, which shows as similar as no defense scenario.

Although the performance of LoMar reduces when more malicious clients are injected for Amazon dataset in Fig.~\ref{fig:target_amazon}, it has a better result than other defense methods. Median performs worst for the overall testing $58\%$, and the target label accuracy reduces from $98\%$ to $62\%$. Krum performs similar to no defense both on overall and target label. FG+Krum has overall accuracy $91\%$-$83\%$ and target label accuracy $99\%$-$77\%$. The overall accuracy of FG is $85\%$-$71\%$ and target label accuracy is $95\%$-$52\%$.

{\subsubsection{Analysis of Imbalanced Samples with Different $\lambda$}\label{SubSec:lambda}
	We then evaluate the performance of LoMar with different values of $\lambda$, which denotes the ratio of training samples in the major label at each private remote dataset. We setup the range of $\lambda = [0,1, 0.9]$ and compare the performance of LoMar with the existing defense approaches.} 

{Fig.~5 shows the overall and target label testing accuracy for different $\lambda$ values. We could notice from the results that when the value of $\lambda$ increases, the learning performance could decrease rapidly. We consider this phenomenon occurs because the introduced defense approaches falsely predict the imbalanced clean updates as malicious, especially under a higher imbalance degree. For example, the accuracy of Krum on MNIST dataset reduces from $61.2\%$ to $15.5\%$. We also find that though LoMar has a decreased performance at the same time, the decrease degree is lighter when compared to other defense methods. These results show that LoMar still outperforms existing approaches when the remote training samples are imbalanced. }

\begin{table*}
	\centering
	\caption{Testing accuracy on compared defenses under multi-labels flipping attack: $\lambda=0, \tau = 0.1$.}
	\begin{tabular}{|c||c|c|c|c|c|c|c|c|c|c|}
		\hline
		~ & Overall Acc. & Acc. 1 & Acc. 7 & Acc. 6 & Acc. 9 & Acc. 2 & Acc. 5 & Acc. 0 & Acc. 8 &  Other Acc.\\
		\hline
		LoMar & $\underline{\textbf{0.885}}$ & $\underline{\textbf{0.962}}$ & $\underline{\textbf{0.895}}$ & \textbf{0.936} & $\underline{0.864}$ & $\underline{\textbf{0.844}}$ & $\underline{\textbf{0.786}}$ & $\underline{\textbf{0.955}}$ & \textbf{0.836} & $\underline{\textbf{0.875}}$  \\
		\hline
		Krum & $\underline{0.734}$ & 0.834 & $\underline{0.800}$ & 0.765 & 0.750 & $\underline{0.624}$ & \textbf{0.586} & $\underline{0.750}$ & 0.687 & \textbf{0.801}  \\
		\hline
		FG & \textbf{0.802} & \textbf{0.905} & 0.766 & $\underline{0.918}$ & \textbf{0.925} & 0.622 & $\underline{0.457}$ & \textbf{0.893} & $\underline{\textbf{0.938}}$ & $\underline{0.766}$  \\
		\hline
		FG + Krum & 0.415 & 0.859 & \textbf{0.802} & $\underline{\textbf{0.996}}$ & $\underline{\textbf{0.991}}$ & 0.398 & 0.204 & 0.040 & 0.000 & 0.359  \\
		\hline
		Median & 0.655 & $\underline{0.885}$ & 0.579 & 0.812 & 0.472 & \textbf{0.706} & 0.282 & 0.484 & $\underline{0.778}$ & 0.619  \\
		\hline
		No Defense & 0.705 & 0.822 & 0.809 & 0.763 & 0.748 & 0.632 & 0.585 & 0.749 & 0.688 & 0.771  \\
		\hline
	\end{tabular}
	\label{tab:result_multilabel}
\end{table*}

\subsubsection{Analysis Compared Algorithms under Multiple-Labels Flipping Attack}
{In this part, we investigate the performance of LoMar under a label-flipping attack with multiple targets. Difference from the previously introduced single target label-flipping attack, the malicious updates in this scenario consists of four different source-target label pairs in MNIST dataset: $1$ to $7$, $6$ to $9$, $2$ to $5$ and $0$ to $8$. Note that we choose the source and target labels based on common sense in the development of an attacker: the malicious is more difficult to be found if two labels share a close distribution. Specifically, for better presentation and comparison, we still set the percentage of malicious clients to $10\%$ that each pair controls $25\%$ of malicious updates. In this setting, $4$ out of $10$ classes are under attack and the detailed experimental results are shown in Table.~\ref{tab:result_multilabel}. }

LoMar has the best overall testing accuracy and the accuracy for other labels which are clean $88.5\%$ and $87.5\%$, and FG+Krum has the worst performance $41.5\%$ and $35.9\%$, which is lower than no defense scenario $70.1\%$ and $77.1\%$. For each label, LoMar increases the testing accuracy $10\%$-$20\%$ than no defense, for instance, LoMar is $93.6\%$ for the label $6$, and there is only $76.3\%$ without defense. Krum has a similar performance as no defense scenario, the degree of increase is $-0.2\%$-$1.2\%$. The result indicates that it has limited abilities to defend the attack as the attack rate approaches the upper bound of an attacker in a system without defense. For FG, we can find the defense performance against attacks from label $1$ to label $7$ and label $0$ to label $8$ is sufficient, and it fails to other attacks. We assume that the reason FG fails to defense against this attack is that it penalizes the learning rate of many clean updates. For FG combined with Krum defense model, this method shows extreme results, for example, it has the highest accuracy $99.6\%$ at label $6$, but it cannot determine which label is $8$, i.e., $0\%$ for label $8$. We infer the reason is that the combination of FG and Krum causes more malicious alarms on clean updates which may come from the features of these two defense mechanisms. Median defense method does not increase the performance sufficiently, but it performs very low for the label $5$, i.e., $28.2\%$. Furthermore, on the hand of learning accuracy, Median defense fails to obtain a feasible learning accuracy on both target labels and other non-target labels.

\begin{table}
	\centering
	\caption{Training and testing accuracy for each metrics on the VGGFace2 dataset: $\lambda=0, \tau = 0.1$.}
	\begin{tabular}{|c||c|c|c|}
		\hline
		Training &  Overall Acc. & Target Acc. & Other Acc.\\
		\hline
		LoMar & $\underline{0.830}$ & $\underline{\textbf{0.733}}$ & \textbf{0.854} \\
		\hline
		FG & 0.799 & 0.679 & 0.829 \\
		\hline
		Krum & \textbf{0.834} & $\underline{0.686}$ & $\underline{\textbf{0.870}}$ \\
		\hline
		FG+Krum & $\underline{\textbf{0.862}}$ & \textbf{0.705} & $\underline{0.846}$ \\
		\hline  
		No defense & 0.834 & 0.653 & 0.880 \\
		\hline
		\hline
		Testing &  Overall Acc. & Target Acc. & Other Acc.\\
		\hline
		LoMar & $\underline{\textbf{0.897}}$ & $\underline{\textbf{0.839}}$ & $\underline{\textbf{0.912}}$ \\
		\hline
		FG & 0.867 & 0.770 & $\underline{0.891}$ \\
		\hline
		Krum & $\underline{0.869}$ & $\underline{0.780}$ & $\underline{0.891}$ \\
		\hline
		FG+Krum & \textbf{0.884} & \textbf{0.831} & \textbf{0.898} \\
		\hline  
		No defense & 0.872 & 0.789 & 0.893 \\
		\hline
	\end{tabular}
	\label{tab:DNN}
\end{table}

\subsubsection{Evaluation on SqueezeNet for VGGFace2 Dataset} We evaluate the defense performance based on the training and testing classification matrix in Table.~\ref{tab:DNN}. Although LoMar defense is influenced by the attacker with $73.3\%$ target label training accuracy and $83.9\%$ of the testing classification. It also has the best accuracy among these defenses. Moreover, it still has the best overall accuracy both on training and testing as $83.0\%$ and $89.7\%$, and the best training accuracy for other labels $85.4\%$. {We could notice that comparing to FG+ Krum method, LoMar obtains a higher target accuracy and other label accuracy but lower overall accuracy. We consider this might be because the performance of the source label is enhanced from the byzantine tolerance strategy, which is also supported by the results in Krum. }

{Other defenses do not perform efficiently on the target label either training or testing side, i.e., the accuracy of all defenses is lower than $70\%$ and $80\%$. For Krum, we find that the performance is only $3\%$ better than with no defense and we consider this method has almost zero ability on dealing with the attackers in this experiment. For FG, it has a weak overall training accuracy $79.9\%$ and other clean labels $82.9\%$. This indicates that FG has limited defense performance on addressing attackers at a DNN FL model. In conclusion, the results in DNN models support that LoMar defense algorithm has the ability to detect attacks and outperforms other methods.}

\begin{figure*}
	\begin{subfigure}{0.5\columnwidth}
		\includegraphics[width = 1\columnwidth]{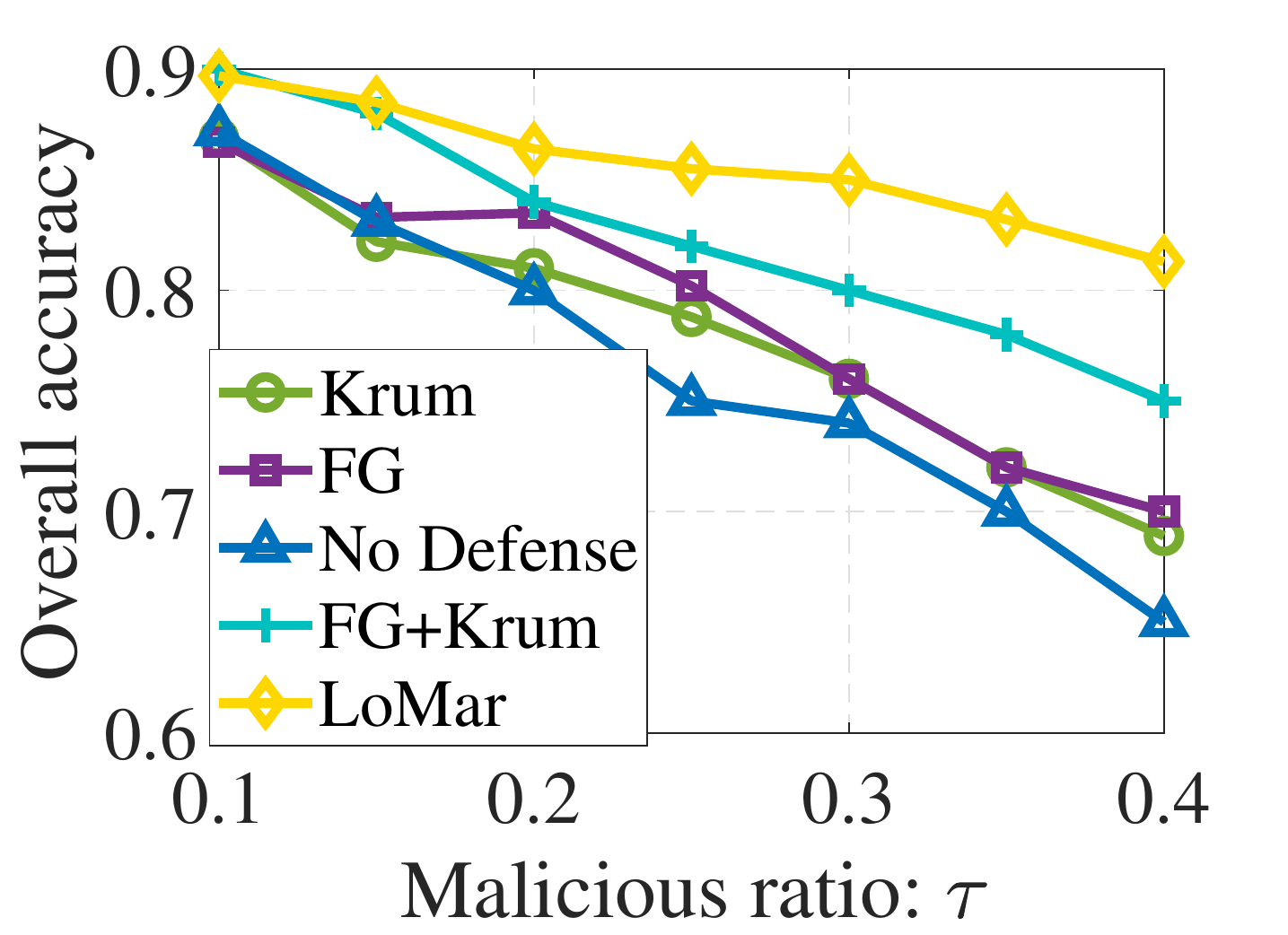}
		\caption{VGGFace2}
		\label{fig:overall_vgg}
	\end{subfigure}
	\begin{subfigure}{0.5\columnwidth}
		\includegraphics[width = 1\columnwidth]{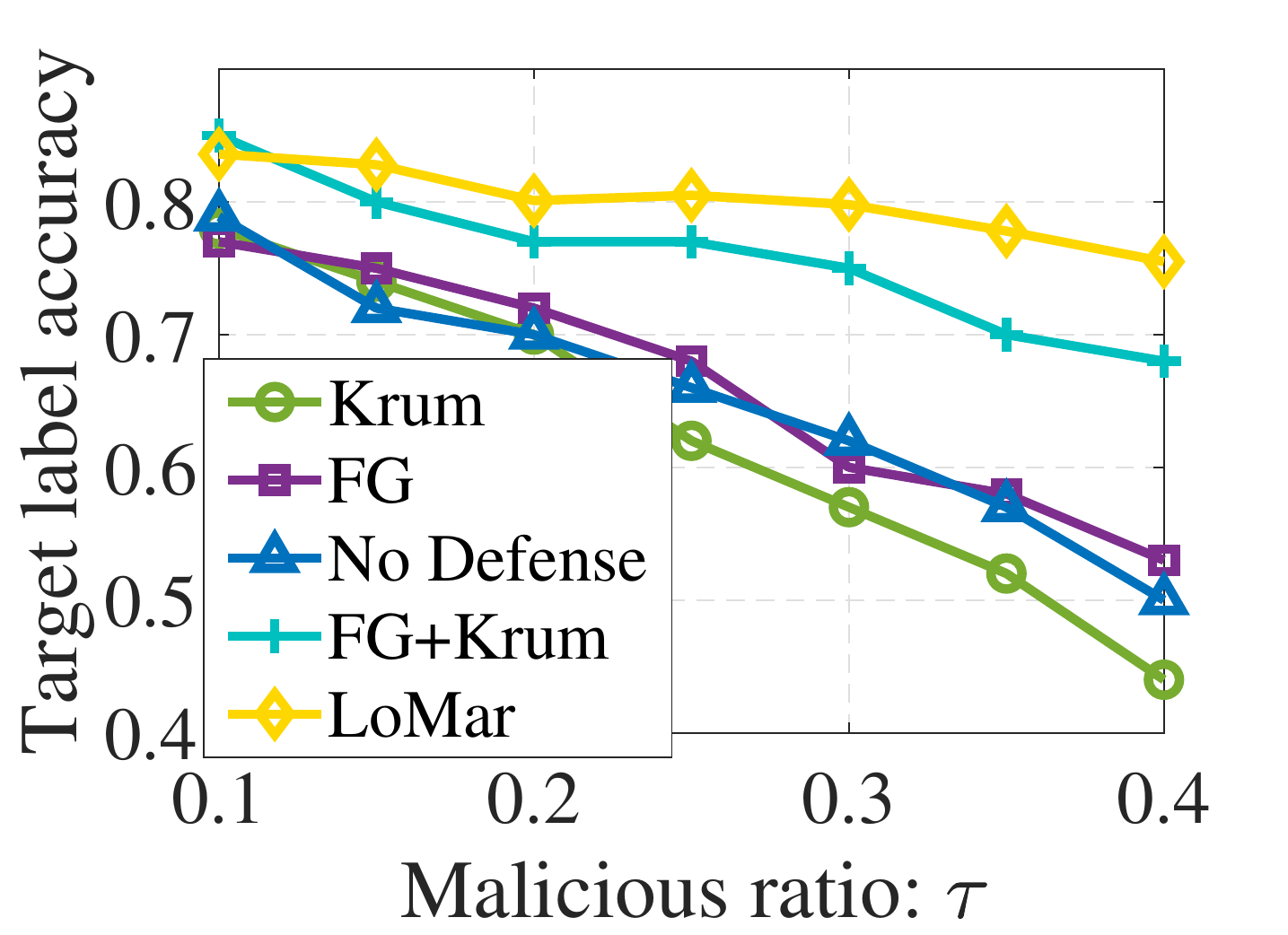}
		\caption{VGGFace2}
		\label{fig:target_vgg}
	\end{subfigure}
	\begin{subfigure}{0.5\columnwidth}
		\includegraphics[width = 1\columnwidth]{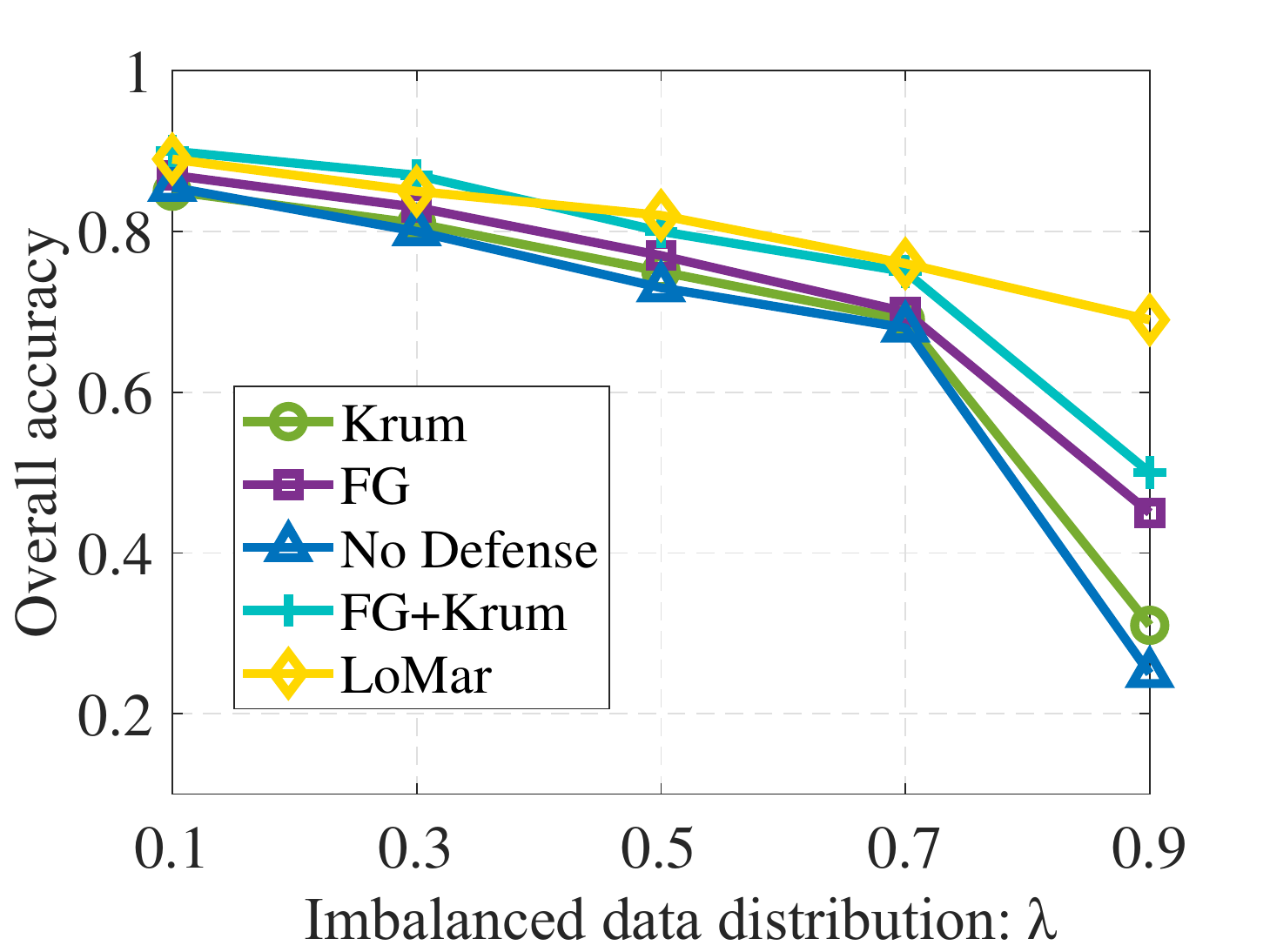}
		\caption{VGGFace2}
		\label{fig:overall_noniidvgg}
	\end{subfigure}
	\begin{subfigure}{0.5\columnwidth}
		\includegraphics[width = 1\columnwidth]{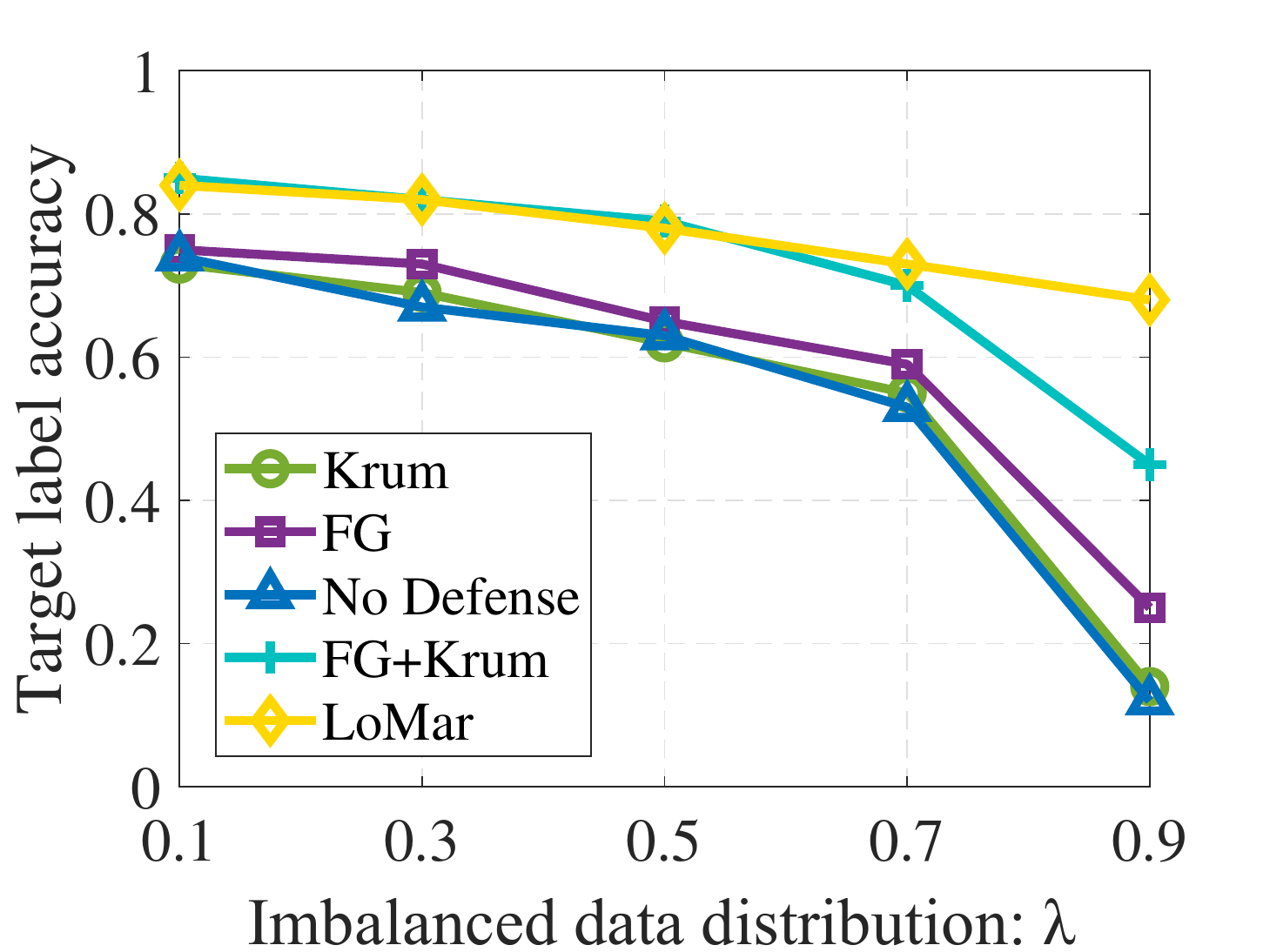}
		\caption{VGGFace2}
		\label{fig:target_noniidvgg}
	\end{subfigure}
	\caption{Testing accuracy for the compared algorithms on the VGGFace2 dataset with different malicious ratio $\tau$ and different imbalanced ratio $\lambda$. (a) the overall testing accuracy with $\tau = [0.1 - 0.4], \lambda = 0$; (b) the target label testing accuracy with $\tau = [0.1 - 0.4], \lambda = 0$; (c) the overall testing accuracy with $\lambda = [0.1 - 0.9], \tau = 0.1$; (d) the target label testing accuracy with $\lambda = [0.1 - 0.9], \tau = 0.1$.}
\end{figure*}

\begin{figure*}[ht]
	\centering
	\begin{subfigure}{0.5\columnwidth}
		\includegraphics[width = 1\columnwidth]{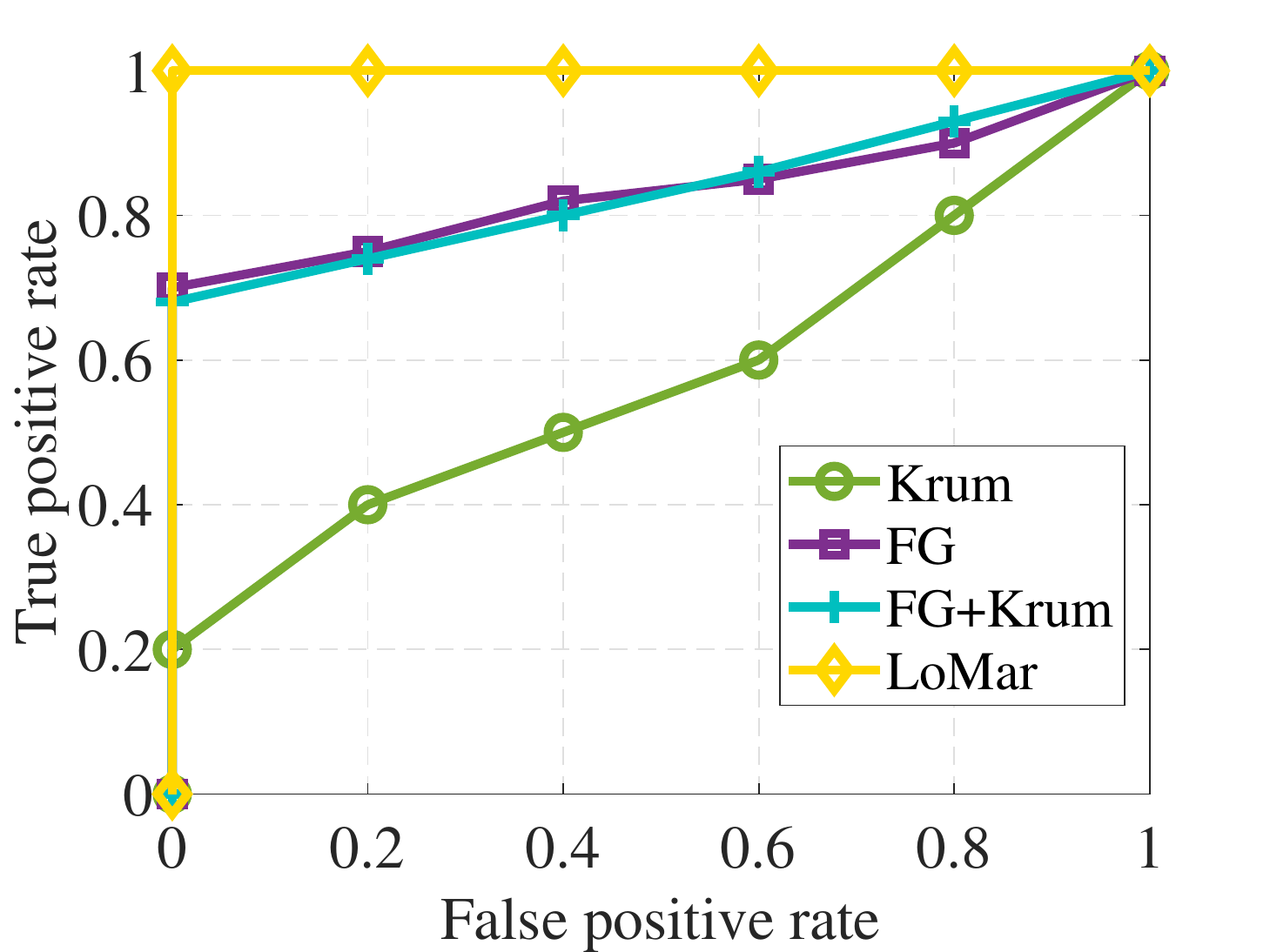}
		\caption{MNIST}
		\label{fig:roc_mnist}
	\end{subfigure}
	\begin{subfigure}{0.5\columnwidth}
		\includegraphics[width = 1\columnwidth]{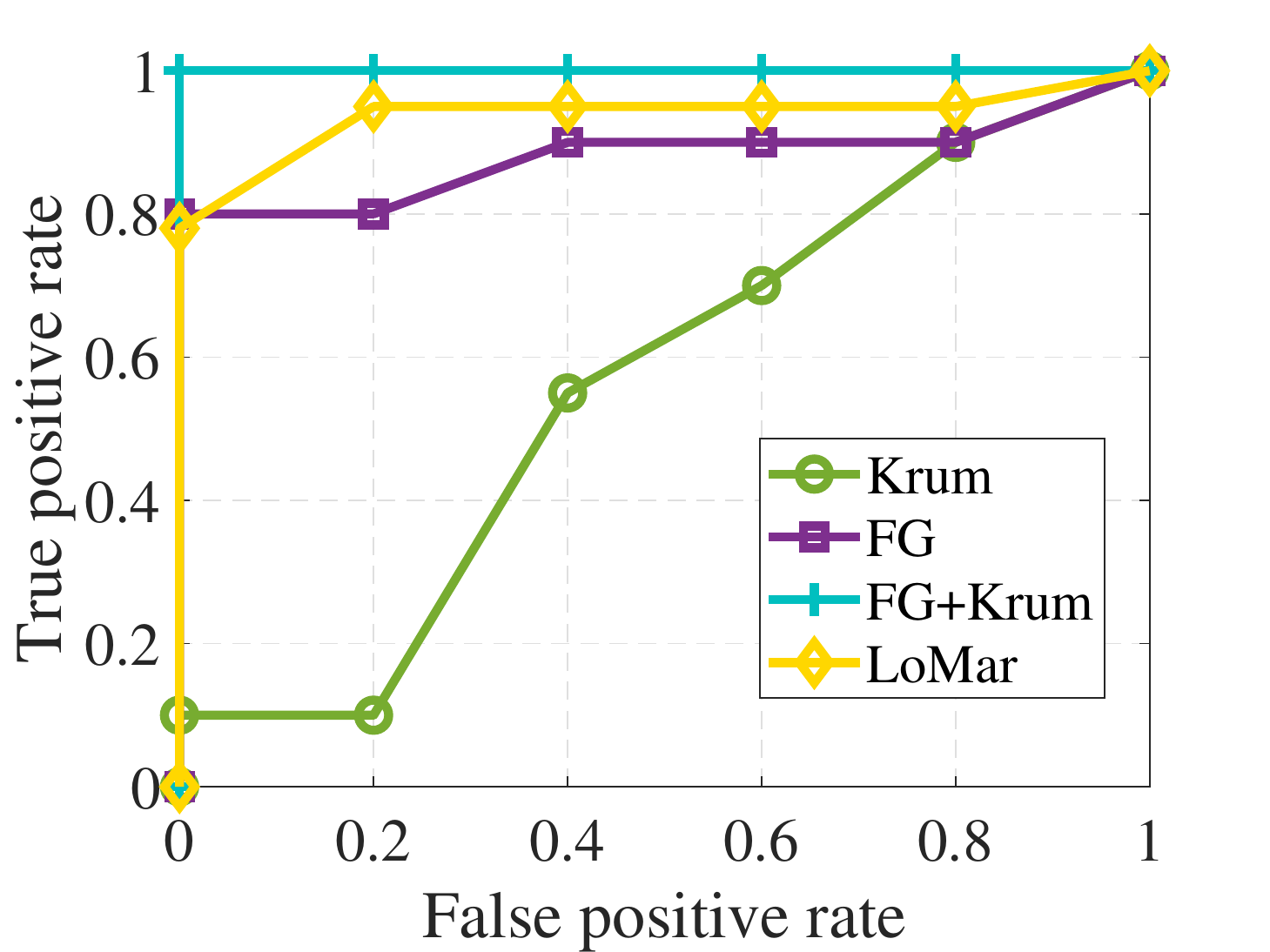}
		\caption{KDD}
		\label{fig:roc_kdd}
	\end{subfigure}
	\begin{subfigure}{0.5\columnwidth}
		\includegraphics[width = 1\columnwidth]{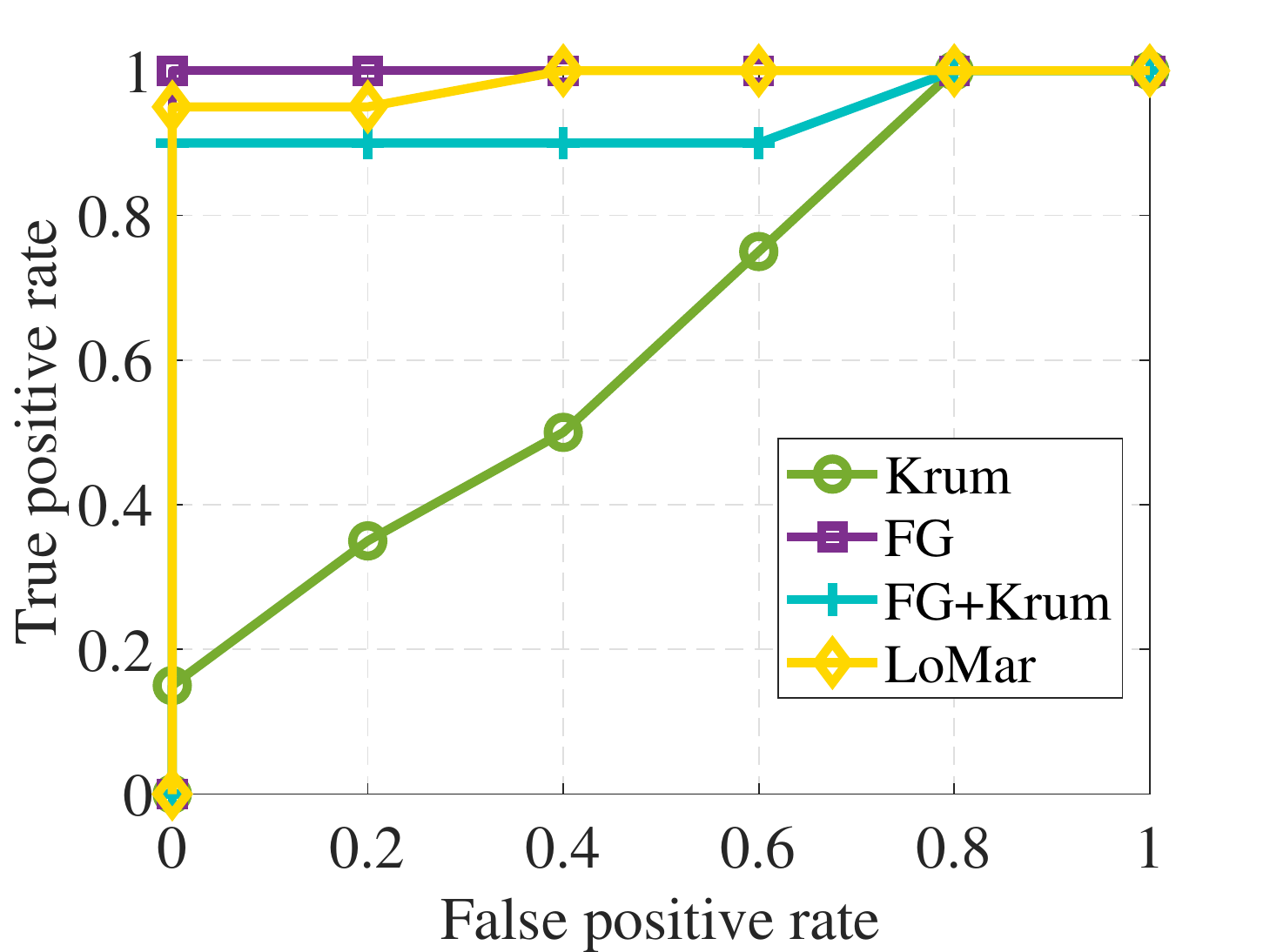}
		\caption{Amazon}
		\label{fig:roc_amazon}
	\end{subfigure}
	\begin{subfigure}{0.5\columnwidth}
		\includegraphics[width = 1\columnwidth]{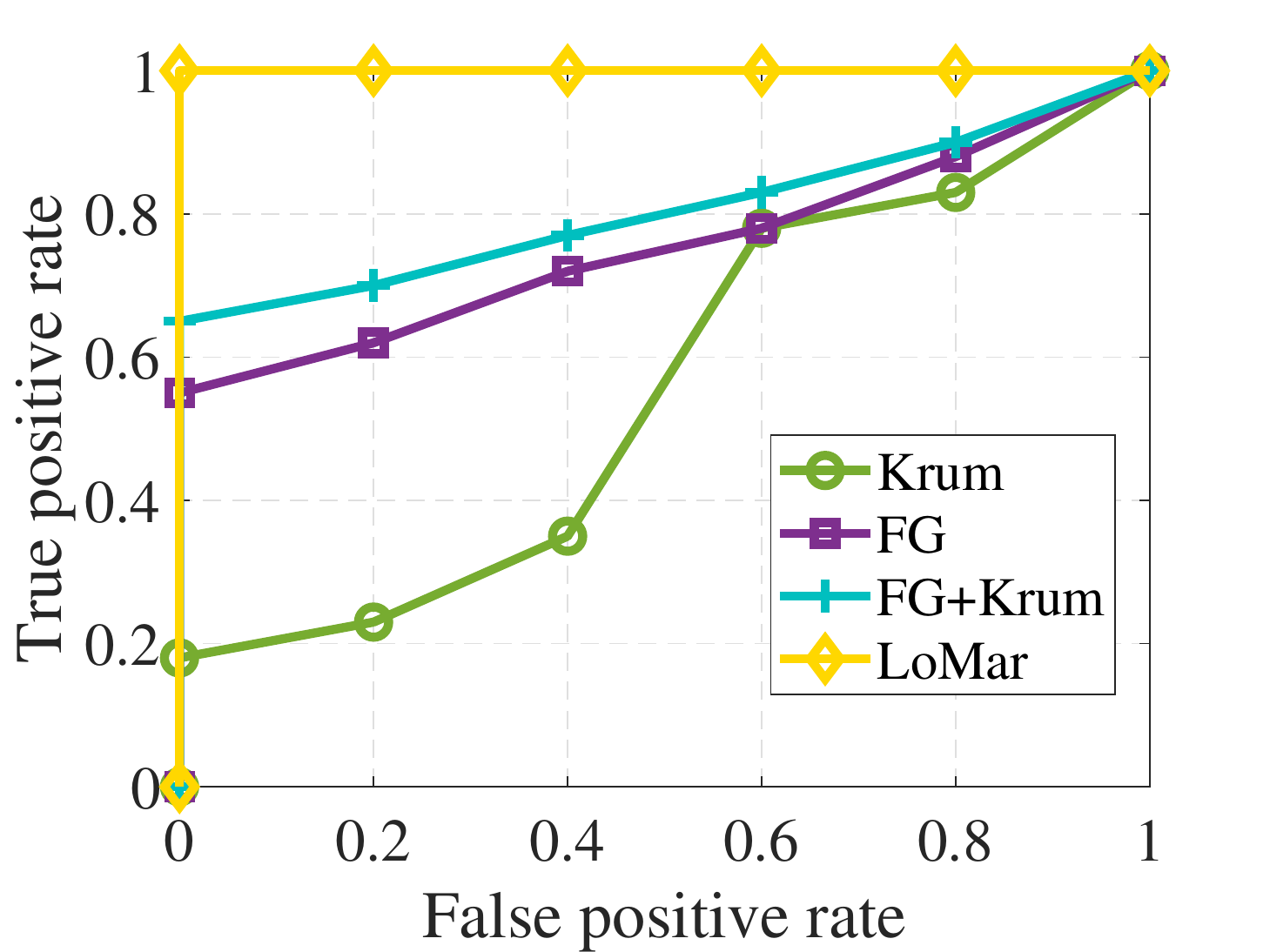}
		\caption{VGGFace2}
		\label{fig:roc_VGG}
	\end{subfigure}
	\caption{The ROC curves of true and false malicious alarms for the compared defense algorithms with $\tau = 0.1$ and $\lambda = 0$.}
	\label{fig:roc}
\end{figure*}

\subsubsection{Impact of Different $\lambda$ and $\tau$ on VGGFace2} 
The results of impact with different $\tau$ on VGGFace2 dataset in Fig.~\ref{fig:overall_vgg} and \ref{fig:target_vgg} show that the performance of compared defense methods does not have a significant reduce if more malicious clients are injected into the dataset. It might come from the reason that defense performance of compared defense methods cannot remove the impact of attacker completely even from the lowest ratio $\tau$, and attack objectives reach the upper bound as a low malicious update ratio. Fig.~\ref{fig:overall_noniidvgg} and \ref{fig:target_noniidvgg} show that when imbalanced ratio $\lambda$ increases from $0.1$ to $0.7$, the performance does not have obvious decreasing. When $\lambda$ increases to $0.7$, we can see existing defense methods reduce sharply, but LoMar also keeps the reduction degree and sufficiently protects the system.

\subsection{Malicious Alarm Evaluation}\label{SubSec:ROC}
{
	In this part, we further investigate the reasons that lead to the different experimental results between our proposed LoMar and compared approaches under FL attacks. Specifically, we study the false alarms in the defense approaches, which could be both the false malicious and the false clean determinations. In this paper, we use ROC analysis to study this phenomenon, which is a state-of-art statistical model evaluation tool.}

\subsubsection{ROC Curve Metrics} 
{For presentation, we first introduce several important parameters in the ROC analysis. Generally, there are two benchmarks to illustrate the performance of a learned classifier in ML, sensitivity and specificity. In this work, we denote the number of true clean updates as $N$ and the number of malicious updates as $M$. Based on $N$ and $M$, we add four corresponding variables to introduce the definition of remote updates under the compared defense approaches. Specifically, $N_f$ is the number of updates that are falsely determined as clean while  $N_t$ is the number of correctly determined clean updates. Meanwhile, $M_t$ represents the number of successfully detected malicious and $M_f$ denotes the number of false alarms. Obviously, we could notice that $N = N_t + M_f$ and $M = N_f + M_t$. As such, the ROC analysis considers the sensitivity as the rate of correctly determined clean updates $\frac{N_t}{N}$ and the specificity as the rate of correctly determined malicious $\frac{M_t}{M}$. In this condition, we could obtain the ROC curve, which is a plot of the sensitivity on the $y$ axis and the values of $1-$ specificity on the $x$ axis.} Note that the worst results for a defender classifier in the ROC curve reflect into a $45^\circ$ diagonal line from $(0,0)$ to $(1,1)$, which indicates that malicious updates are detected by random chance. And the better the classifier is, the area of ROC curve is bigger.


\subsubsection{ROC Analysis} Fig.~\ref{fig:roc} shows the ROC curves of compared defense methods on datasets mentioned in this paper. Note that due to the mechanism of Median which obtains the joint model from sorting each parameter of the updates and selecting median value as the contribution to the joint model, this method leads to a large number of false detecting and we only discuss LoMar, FG, Krum and FG+krum defense methods.

Firstly, we can notice that Krum has the worst results on each ROC curve. In Fig.~\ref{fig:roc_mnist}, the Krum defense classifier is close to a $45^\circ$ diagonal line from $(0,0)$ to $(1,1)$ which means that the efficiency of Krum almost approaches a random chance. Secondly, the FG method has the best ROC curve on Amazon dataset in Fig.~\ref{fig:roc_amazon} and the FG combined with Krum method has the best performance on KDD dataset in Fig.~\ref{fig:roc_kdd} as the curves are nearly a square area. However, the results of these two defense methods on MNIST and VGGFace2 datasets are not ideal as the areas are far smaller than the results from our LoMar defense method. Especially, our LoMar defense has the best ROC curve on MNIST and VGGface2 dataset and the second performance on KDD and Amazon dataset. {Furthermore, the obtained results show that our proposed LoMar has the best overall average ROC curve area on all four datasets, and we believe that the results on ROC are related to defense performance results in Sec.~\ref{Section:analysis}. The correlation between ROC curves and the FL model performance supports our intuition that compared to existing defense methods, LoMar can develop a successful malicious removing strategy, which provides less number of both the false malicious and the false cleans in the learning process of the FL system. }

\section{Related Works}\label{Sec:Relatedworks}

\subsection{Poisoning Attacks} \label{Sec:Poisongingattacks}

\textbf{Poisoning attacks in ML.} 
Poisoning is the most widespread type of attacks in the history of the learning field \cite{angluin1988learning, aslam1996sample, stempfel:hal-00186391, natarajan2013learning}. In general, poisoning attacks reduce the learning model accuracy by manipulating the learning training process to change the decision boundary of the machine learning system. Depending on the goal of poisoning attacks, we classify those attacks into two categories: targeted poisoning attacks \cite{shafahi2018poison, Gu2017, Yang2017} and non-targeted poisoning attacks \cite{biggio2012poisoning, jagielski2018manipulating, yang2017fake, papernot2016cleverhans}. Non-targeted poisoning attacks are designed to reduce the prediction confidence and mislead the output of the ML system into a class different from the original one \cite{papernot2017practical}. In targeted poisoning attacks, the ML system is forced to output a particular target class designed by the attacker \cite{papernot2018sok}. Compared to non-targeted poisoning attacks, targeted poisoning attacks are more difficult to be found by learning defense systems because targeted poisoning attacks can affect the ML system on the targeted class without changing the output of other classes. 

\noindent\textbf{Poisoning attacks against FL.} Recent studies \cite{bagdasaryan2018backdoor, bhagoji2019analyzing, melis2019exploiting, fung2018mitigating, fang2020local, bhagoji2018model, nasr2019comprehensive, tolpegin2020data} on data poisoning attacks explore the privacy and system risks of a decentralized machine learning system. Basically, targeted poisoning data attacks could manipulate the training dataset in FL system in two ways: label vector manipulation \cite{xiao2012adversarial} and input matrix manipulation \cite{bagdasaryan2018backdoor}. {In label vector manipulation, the attackers can directly modify the labels of the training data into a targeted class, e.g., Label flipping attack \cite{xiao2012adversarial}, where some labels of training data (known as the ``target" classes of the attacker) are flipped into another class to reduce the recognition performance of the target classes. Meanwhile, the attackers can also train a generative model for producing poisoning data \cite{zhang2020poisongan}.} On the other hand, the features of training data could be manipulated to achieve the goal of targeted data poisoning attack by input matrix manipulation \cite{bhagoji2019analyzing, fang2020local}. 

\subsection{Defense Against Poisoning Attack} \label{Sec:Defenses}
\textbf{Defense against for centralized machine learning.} Existing defenses against data poisoning attack are mainly designed on centralized machine learning system. One kind of defenses \cite{barreno2010security, cretu2008casting,suciu2018does, tran2018spectral, rieck2008self} detects the poisoning data based on a negative impact of the learning model. \cite{barreno2010security} proposed to Reject on Negative Impact, which can perform the impact of each training data and discards the training data which has a large negative impact. The other one is to change the worst-case loss from a given attack strategy \cite{koh2017understanding, burkard2017analysis, mei2015using}. 

\noindent\textbf{Defense against poisoning attack at FL.} {One category of existing defense approaches on FL aims to separate the malicious and clean remote clients, e.g., Auror \cite{shen2016auror}. However, Auror uses the trusted training data to determine a threshold between malicious and non-malicious features, which is not realistic in FL settings. Another type of method is developed based on Byzantine-tolerant learning theory \cite{damaskinos2018asynchronous, el2018hidden, xie2018generalized, blanchard2017machine, fung2018mitigating}. For instance, Zeno defense algorithm in \cite{xie2019zeno} removes several largest descents in each local training iteration and combine the rest of the updates to be the joint model. However, the measurement is based on Euclidean distance, which cannot determine the true largest descents due to the dimension reduction. Trim Mean \cite{yin2018byzantine} method achieves the goal by finding a subset of training dataset to minimize the loss function. }

\subsection{Density based Anomaly Detection}
The density based anomaly detection is widely used in data outlier detection \cite{tang2017local, silverman2018density, latecki2007outlier}. Several improvements of the basic density based model have been proposed, for example, a connectivity anomaly detection \cite{tang2002enhancing}, or based on a reverse nearest neighborhoods adding to the nearest neighborhoods and thinking about the relationship between different values as a measurement of anomaly \cite{schubert2012evaluation}. In \cite{guo2014detecting}, they proposed FIND, a density based detection to detect nodes with data faults that do not need to assume the sensing model nor the event injections cost.  \cite{qin2019scalable} uses density estimation to detect the outlier data in large data streams in online applications.

\section{Conclusion}\label{Sec:Conclusion}
In this paper, we propose a new \textit{two-phase} defense algorithm called \textit{LoMar} to address the poisoning attacks against FL systems. In Phase I, LoMar define a kernel density based estimator to indicate the degree of the maliciousness for each update compared to the reference set, which is collected by its $k$-nearest neighborhoods. In Phase II, LoMar design an asymptotic threshold to provide a binary determination of the poisoned updates. Specifically, the provided threshold also protects the clean updates of the FL system from being regraded as malicious by the defender. Our empirical results on four real world datasets with the comparison against four existing defense methods demonstrate that LoMar can address both data and model poisoning attacks against FL.

\begin{appendices}
\begin{figure*}[ht]
\centering
\begin{minipage}{0.5\columnwidth}
	\includegraphics[width = 1\columnwidth]{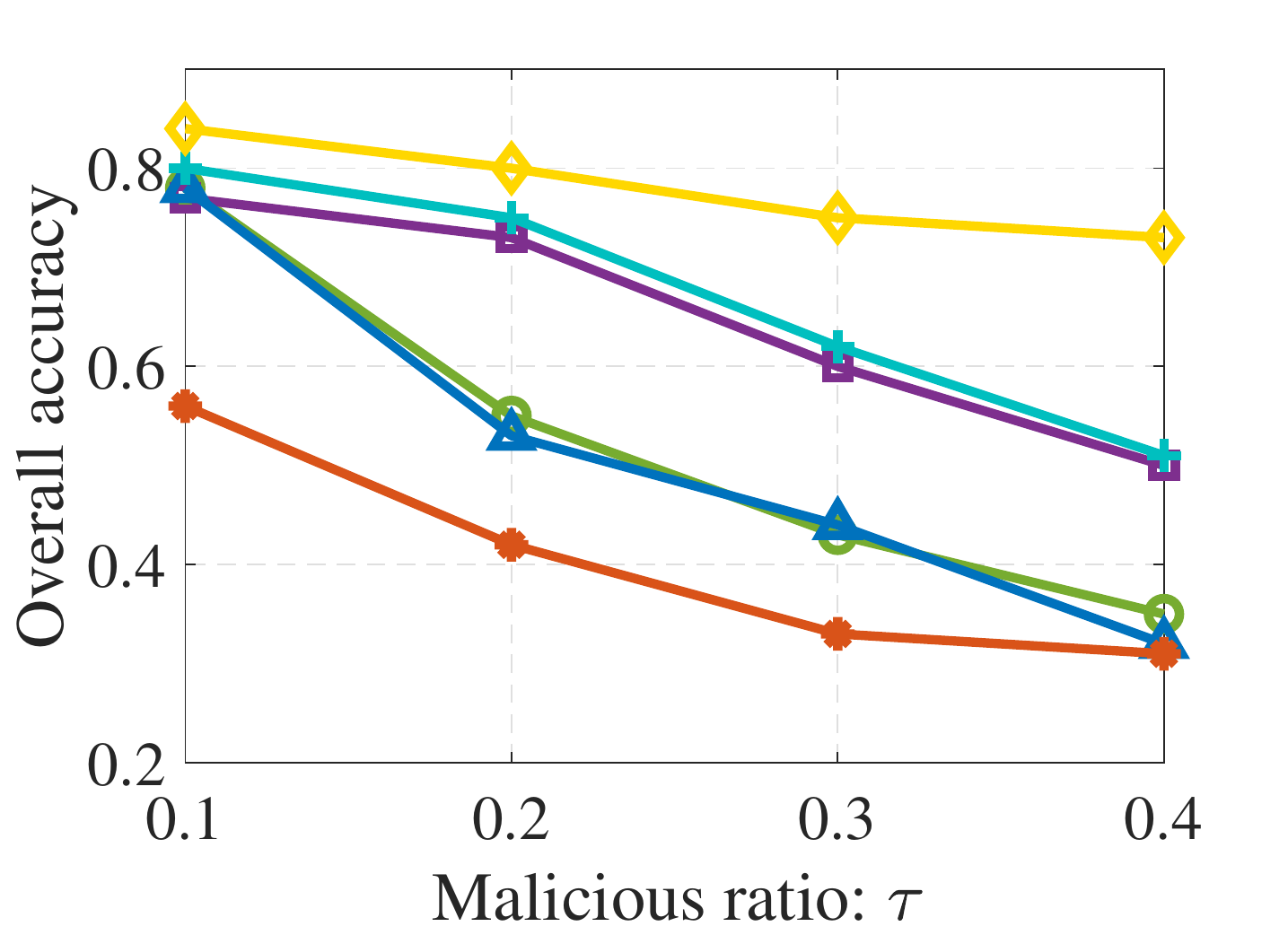}
	\subcaption{MNIST}
	\label{fig:modelmnist_overall}
\end{minipage}
\begin{minipage}{0.5\columnwidth}
	\includegraphics[width = 1\columnwidth]{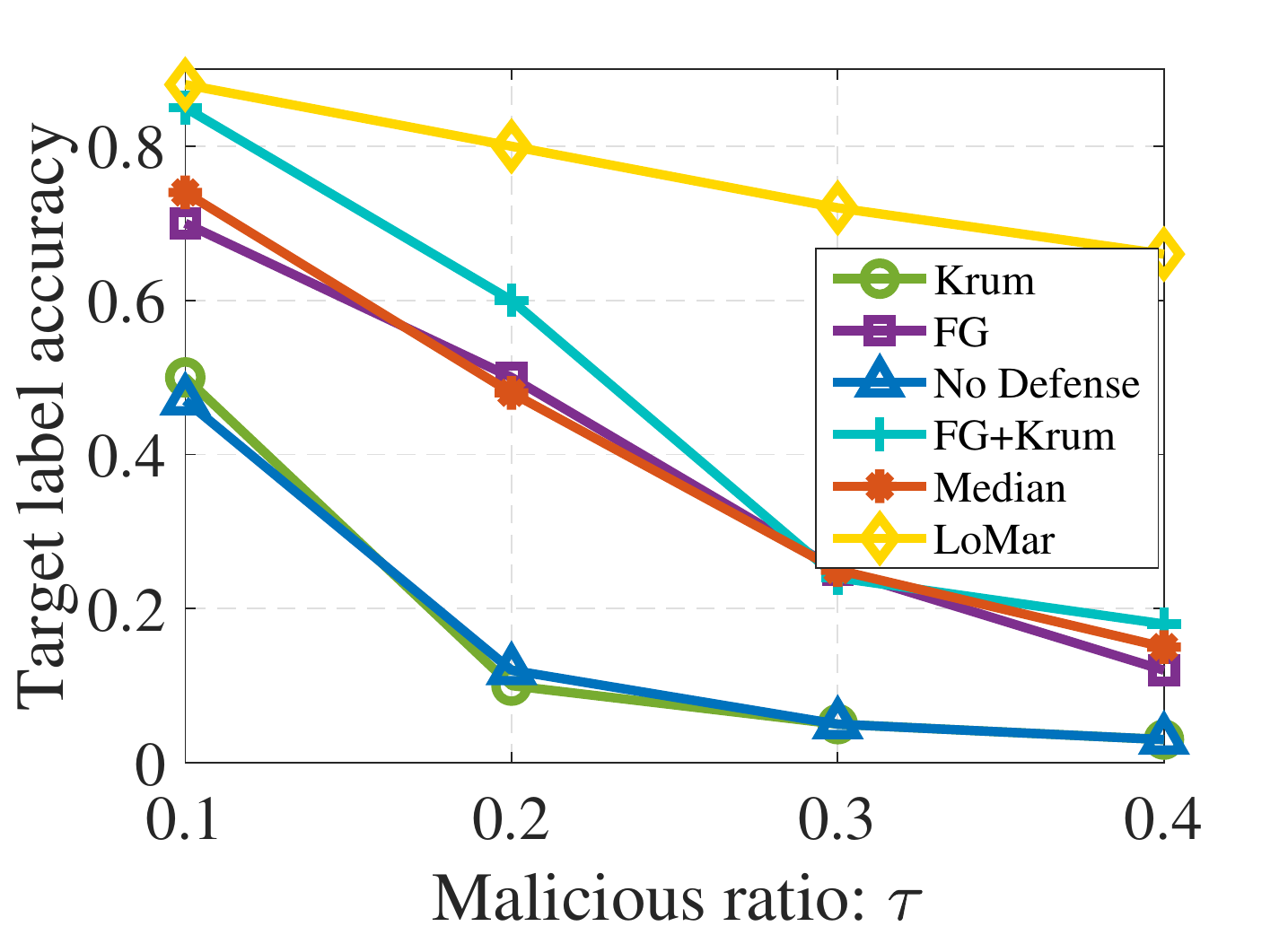}
	\subcaption{MNIST}
	\label{fig:modelmnist_target}
\end{minipage}
\begin{minipage}{0.5\columnwidth}
	\includegraphics[width = 1\columnwidth]{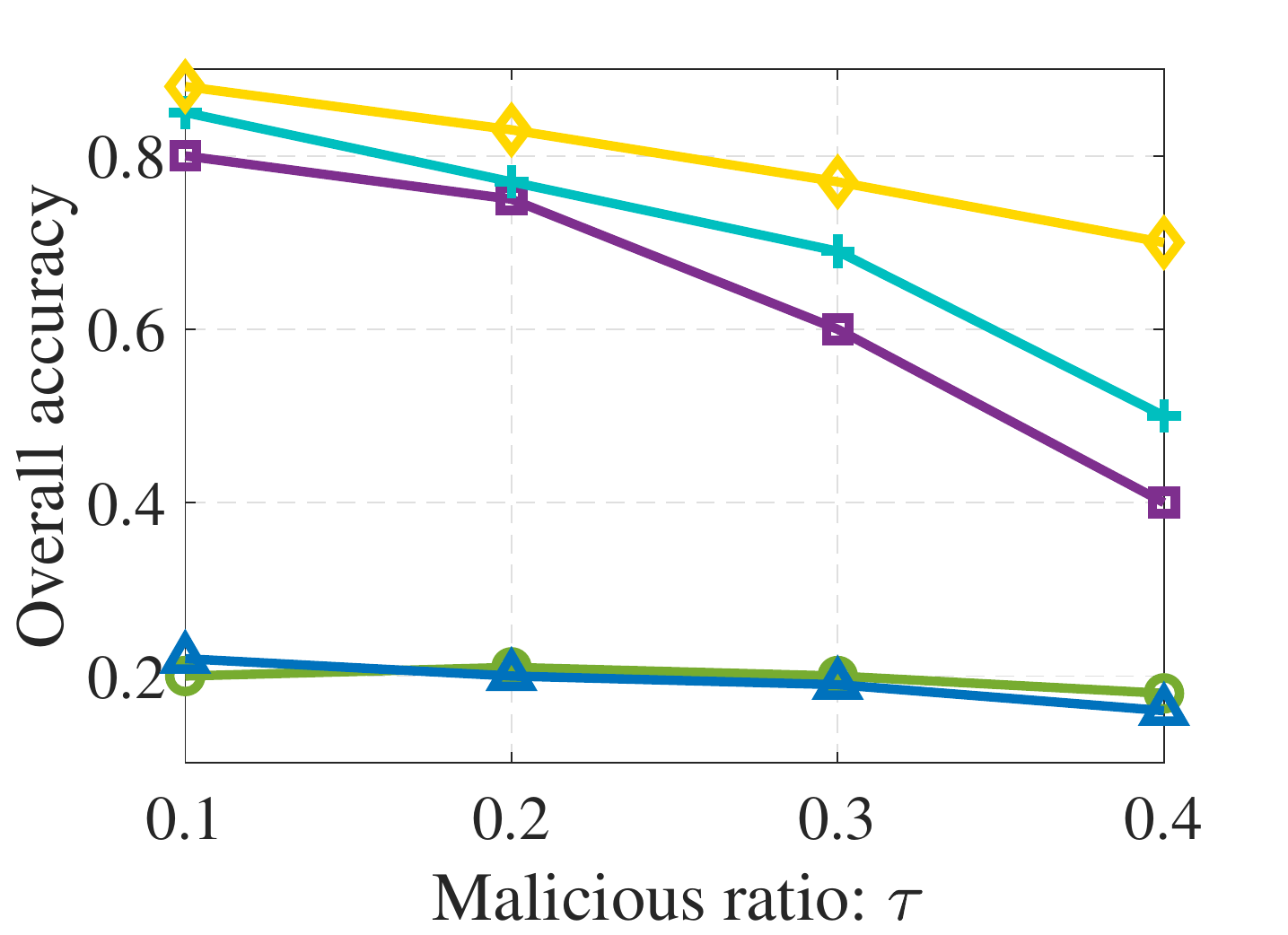}
	\subcaption{VGGFace2}
	\label{fig:modelvgg_overall}
\end{minipage}
\begin{minipage}{0.5\columnwidth}
	\includegraphics[width = 1\columnwidth]{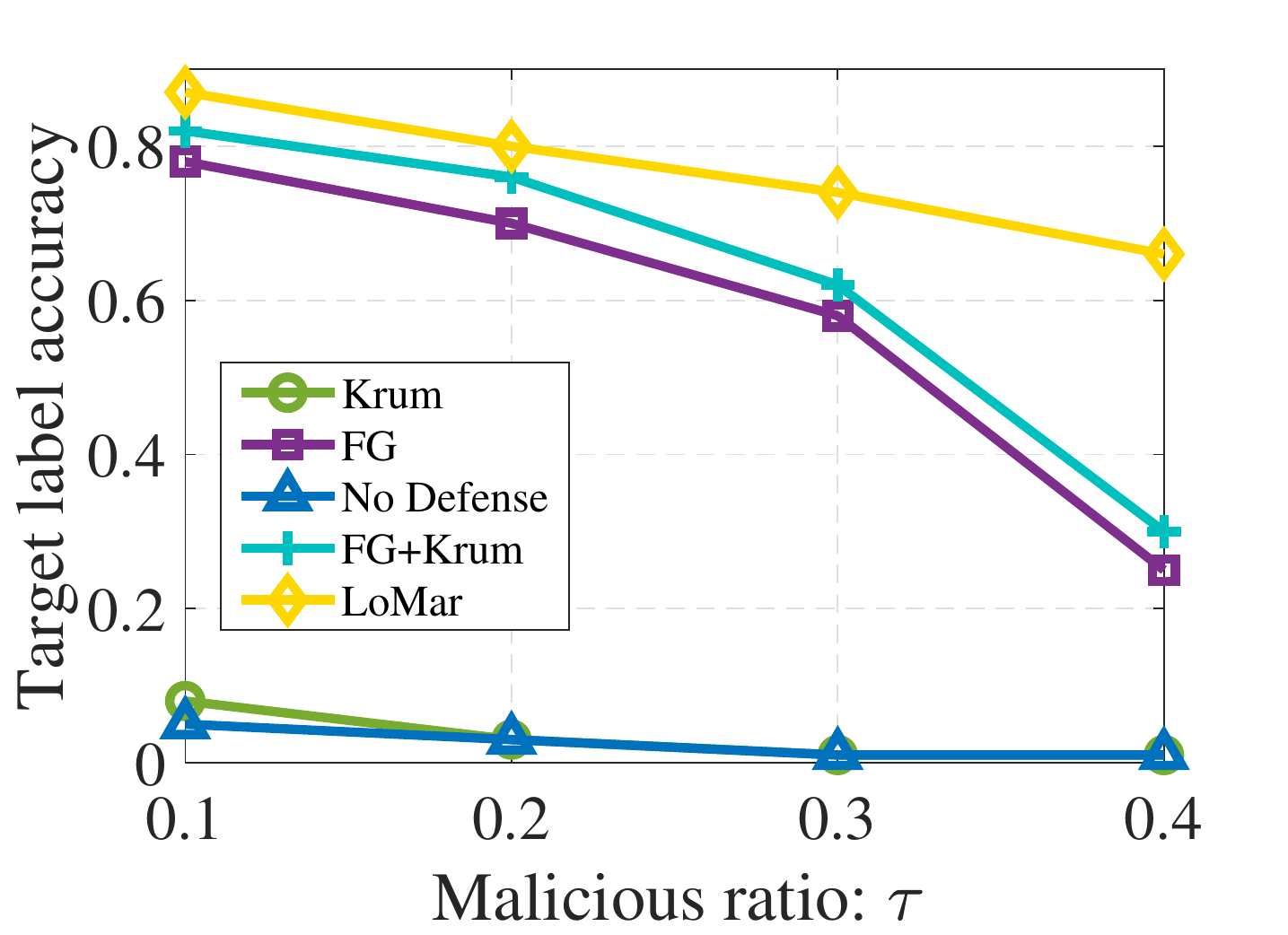}
	\subcaption{VGGFace2}
	\label{fig:modelvgg_target}
\end{minipage}
\caption{Testing accuracy for the compared algorithms under the model poisoning attack with different malicious ratio $\tau = [0.1 - 0.4]$when $\lambda=0$. a: the overall testing accuracy on MNIST; b: the target label testing accuracy on MNIST; c:  the overall testing accuracy on VGGFace2; d: the target label testing accuracy on VGGFace2.}
\label{fig:modelattack}
\end{figure*}

\begin{figure*}[ht]
\centering
\begin{minipage}{0.5\columnwidth}
	\includegraphics[width = 1\columnwidth]{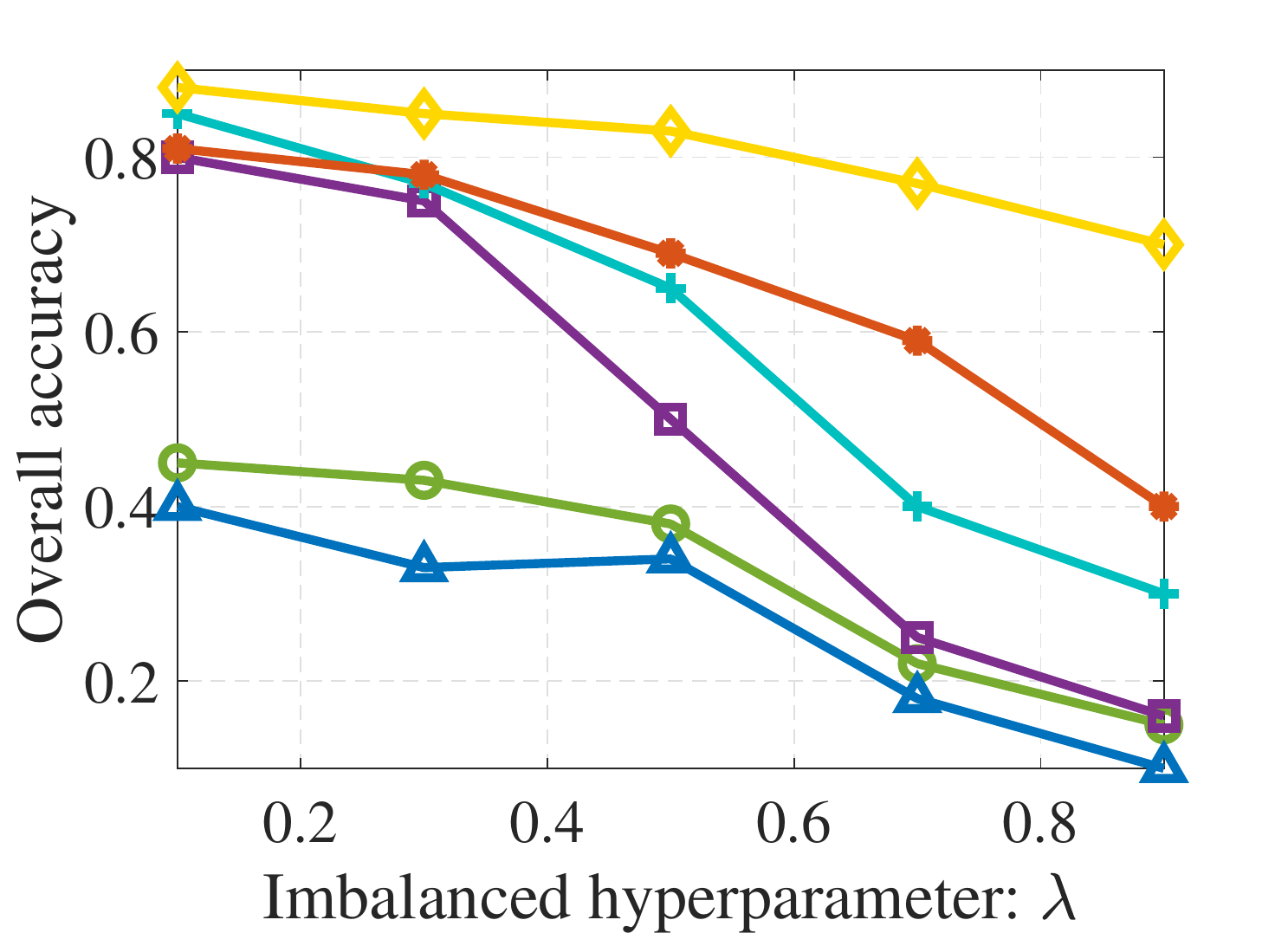}
	\subcaption{MNIST}
	\label{fig:modelmnistiid_overall}
\end{minipage}
\begin{minipage}{0.5\columnwidth}
	\includegraphics[width = 1\columnwidth]{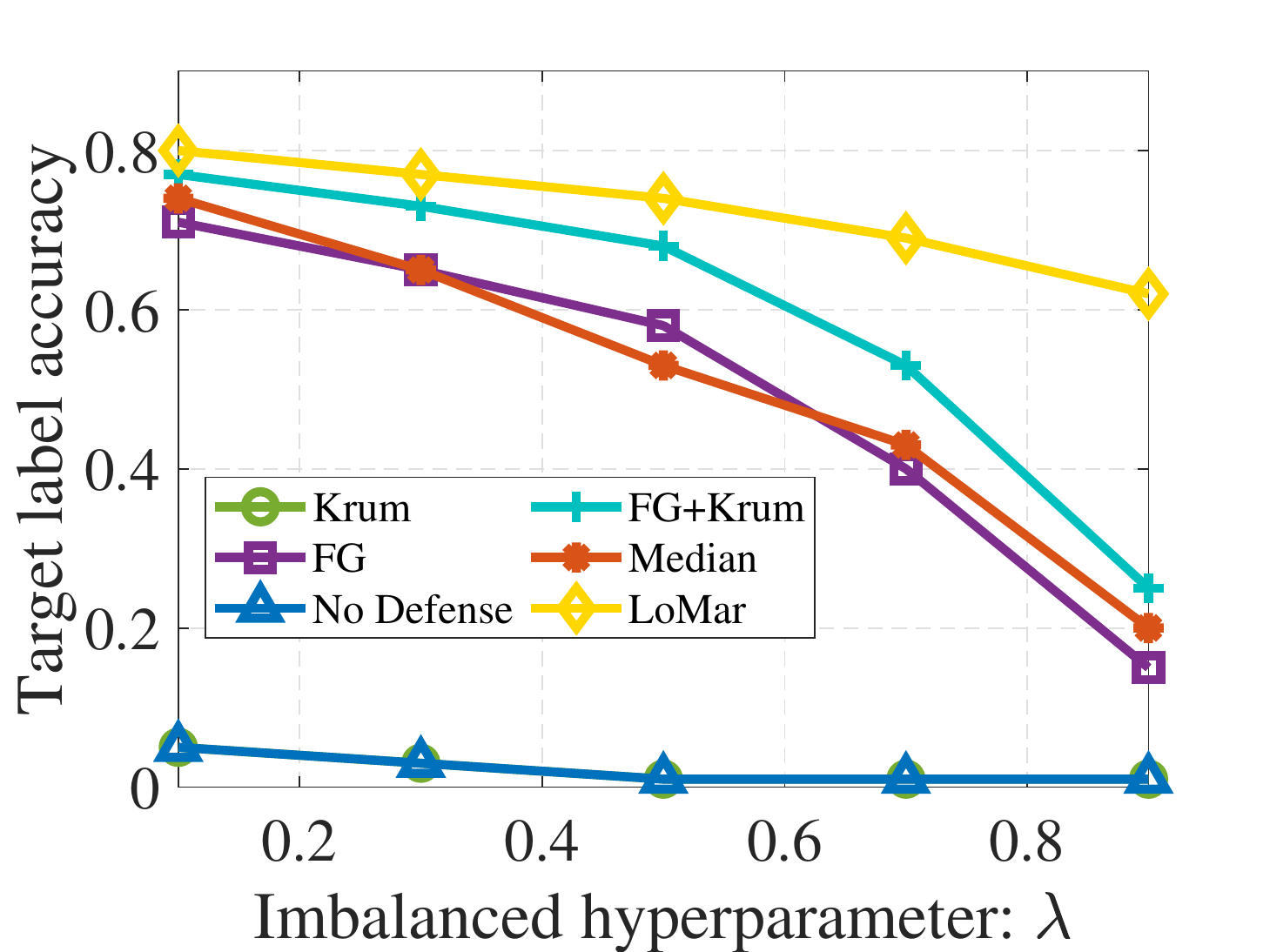}
	\subcaption{MNIST}
	\label{fig:modelmnistiid_target}
\end{minipage}
\begin{minipage}{0.5\columnwidth}
	\includegraphics[width = 1\columnwidth]{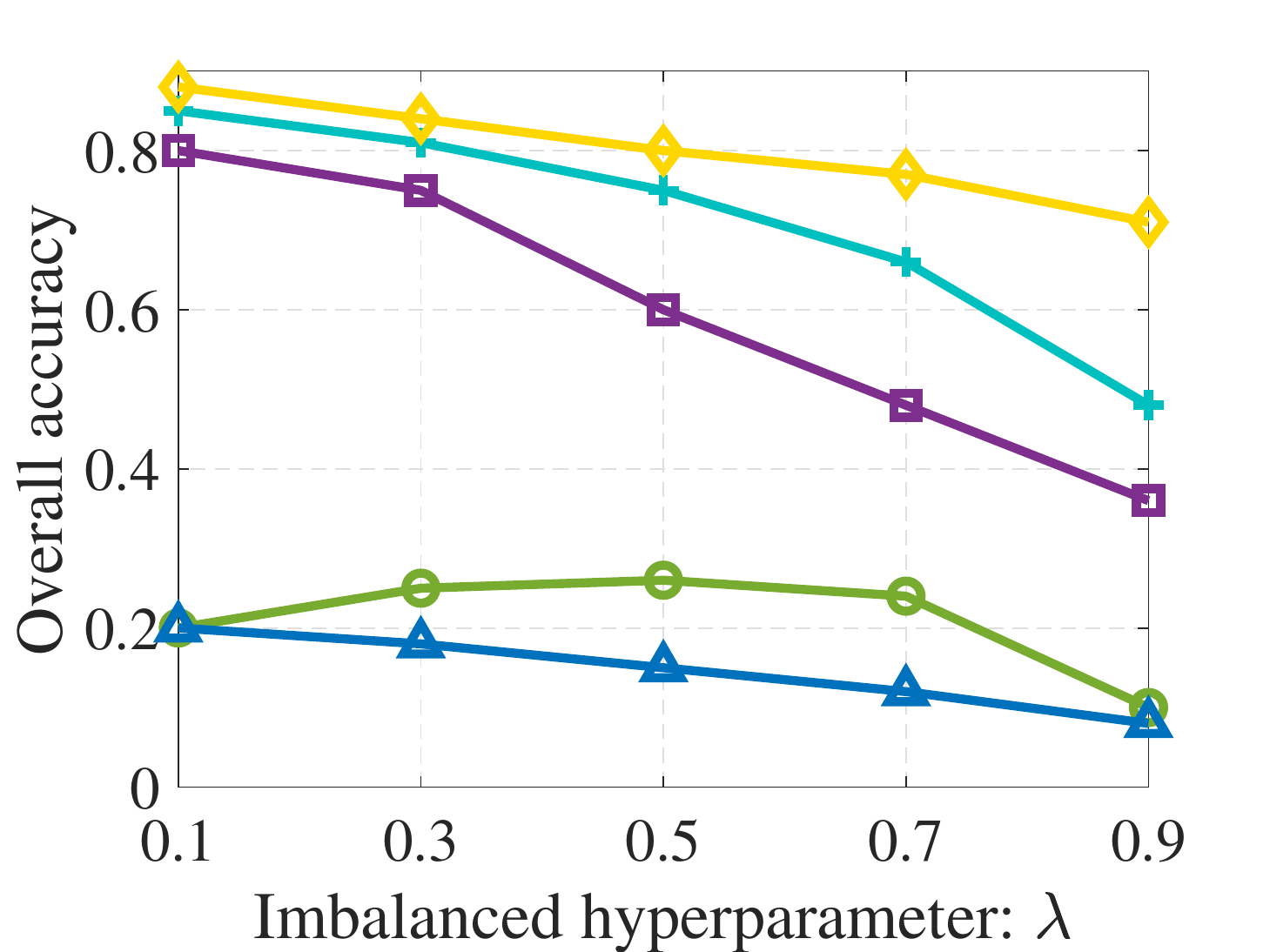}
	\subcaption{VGGFace2}
	\label{fig:modelvggiid_overall}
\end{minipage}
\begin{minipage}{0.5\columnwidth}
	\includegraphics[width = 1\columnwidth]{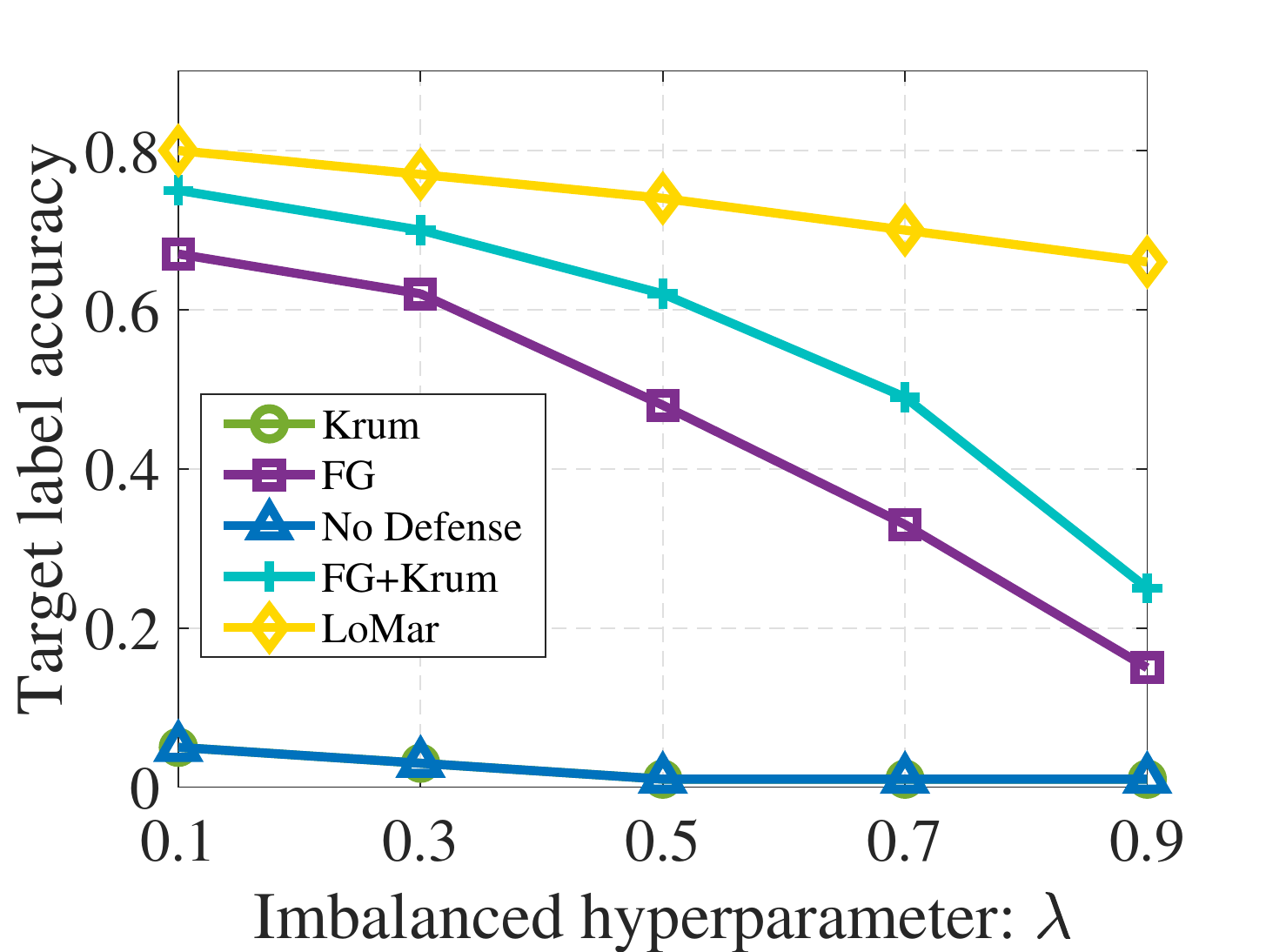}
	\subcaption{VGGFace2}
	\label{fig:modelvggiid_target}
\end{minipage}
\caption{Testing accuracy for the compared algorithms under the model poisoning attack with different imbalanced ratio $\lambda = [0.1 - 0.9]$when $\tau = 0.1$. a: the overall testing accuracy on MNIST; b: the target label testing accuracy on MNIST; c:  the overall testing accuracy on VGGFace2; d: the target label testing accuracy on VGGFace2.}
\label{fig:modelattackiid}
\end{figure*}

{
\section{Evaluation of LoMar under Model Poisoning Attack}\label{Sec:modelattack}
In this part, we evaluate the defense performance of LoMar against model poisoning attacks. Different from traditional data poisoning attacks whose poisons are some manipulated data samples, model poisoning in FL directly sends back a malicious model to the aggregator. Usually, model poisoning attacks establish a stealth metric to bypass a plausible aberrant model detection strategy implemented by the server, which makes it more difficult to be detected, especially for Byzantine-resilient learning strategies. 

\subsection{Model Poisoning Setup}
Followed by the baseline in \cite{bhagoji2019analyzing}, we setup the model poisoning attack on both MNIST and VGGFace2 datasets. Specifically, \cite{bhagoji2019analyzing} generates the malicious update $\hat{\bm{u}}_m$ under a $l_p$-norm distance metric and a explicit boosting factor $\phi$ as $\hat{\bm{u}}'_m = \phi \hat{\bm{u}}_m$. In this work, we set $p=2$ and $\phi = 10$ for the experimental setup. Similarly as the settings in label-flipping attacks, we choose $7$ as the target digit and $1$ as the source digit for MNIST and class ``$2$" to be target from the source class ``$9$'' in VGGFace2. 

\subsection{Results Analysis} 
Like Sec.~4.2.2 and 4.2.3, we analysis the performance of LoMar with the compared existing defenses under two scenarios: different malicious ratio $\tau$ with a fixed $\lambda$ and different imbalanced ratio $\lambda$ with a fixed $\tau$. Specifically, the overall and testing accuracy are considered to be the evaluation metric. Results in Fig.~\ref{fig:modelattack} show that our proposed LoMar outperforms the compared defense approaches against model poisoning attacks. We could notice that as the value of $\tau$ increases, the performance of all defense approaches decrease rapidly. For instance, the target label testing accuracy of FG+Krum method reduces from $83.5\%$ to $21.4\%$. Though LoMar also decreases, the speed of decline is obviously slower than others, e.g., for MNIST with $\tau = 0.4$, it achieves $77.2\%$ for overall accuracy and $63.3\%$ for the target label.

Additionally, we analysis the performance of compared defense approaches with different major class training sample ratio under the imbalanced data distribution scenario, where the results are shown in Fig.~\ref{fig:modelattackiid}. We could notice that under different $\lambda$ values, each defense method has has different performance. For example, Krum cannot determine the true label for the target label both on MNIST and VGGFace2. Generally, whenever the value of $\lambda$ is, LoMar outperforms other methods, especially when $\lambda = 0.9,$ LoMar reaches $30\%$ percent better than FG+Krum defense.}

\section{Explanation of Definition 1}\label{Sec:definition1}
\textit{Proof.}  {The $i$-th remote update $\bm{u}_i$ comes from the local training process on the private dataset $\mathcal{D}_{i} $, which could be denoted as a set of training sample pairs $\mathcal{D}_{i} = \{\mathbf{x}_{i}, {y}_{i} \}$ where $\mathbf{x}_{i}$ is the training dataset input vector and ${y}_{i}$ is the output result. Additionally, we can describe the relationship between predicting output vector from the learning classifier (donated as $\hat{{y}}_{i}$) and the input matrix $\mathbf{x}_{i}$ as} 
\begin{equation}
\centering
\hat{{y}}_{i} = softmax(\bm{u}_i^T \times \mathbf{x}_{i} + b),
\label{Eq:relation_yx}
\end{equation}
where $b$ is the bias of learning model. As the learning update $\bm{u}_i$ is high correlated to the training data distribution, (which is also assumed in \cite{shen2016auror}), we assume that $\bm{u}_i$ follows a similar distribution with $\mathbf{x}$. Then the result probability of ${y}$ at a certain output label $r \in [1, 2, \dots R]$ can be written as $\mathbb{P}(y_r) =  \mathbb{P}(y_r| \mathcal{U})$, where $\mathcal{U} = \{\bm{u}_1, \bm{u}_2, \dots, \bm{u}_{N+M} \}$. However, in our defender, the distribution of $\bm{u}_i$ is not known to the defender and we could estimate the probability distribution of $\mathbf{y}$ according to each label. Specially, assuming that the contribution from the parameters in $\bm{u}_i$ are independent for each output label as $\mathbb{P}(\bm{u}_{i}^{r}, \bm{u}_{i}^{r'} | y_r) = \mathbb{P}(\bm{u}_{i}^{r} | y_r)\mathbb{P}(\bm{u}_{i}^{r'}|y_r )$ when $r \neq r'$, then we can estimate the density distribution of $\bm{u}_i$ from the distribution of $\mathbf{y}$ as
\begin{eqnarray}
\mathbb{P}(\bm{u}_i) && = \mathbb{P}((\bm{u}_i^1, \bm{u}_i^2, \dots, \bm{u}_i^R)|{y}_r)\mathbb{P}({y}_r)  \nonumber\\
&& = \prod\nolimits_{r=1}^R \mathbb{P}(\bm{u}_i^r|{y}_r)\mathbb{P}({y}_r) = \prod\nolimits_{r=1}^R \mathbb{P}(\bm{u}_i^r|{y})\mathbb{P}({y}) 
\end{eqnarray}
Then the value of $F(i)^r$ for $r$-th label can be written as
\begin{eqnarray}
&& F(i)^r = \frac{\sum\nolimits_{j = 1}^{k}\tilde{q}(\bm{u}^{r}_{j})}{k \tilde{q}(\bm{u}^{r}_i ) } = \frac{\sum_{j = 1}^{k}\tilde{q}(\bm{u}^{r}_{j}|  {y})| \mathbb{P}( {y})}{k \tilde{q}(\bm{u}^{r}_i |  {y})\mathbb{P}( {y}) }
\end{eqnarray} 
Taking this equation into the definition of $F(i)$ as
\begin{eqnarray}
&&F(i)  = \frac{\sum_{j = 1}^{k}\tilde{q}(\bm{u}_{j})}{k \tilde{q}(\bm{u}_i ) } \nonumber\\
&& =  \frac{\sum_{j=1}^{k} \prod_{r=1}^R \tilde{q}(\bm{u}_{j}^{r} | {y}) \mathbb{P}({y})}{k \tilde{q} \prod\nolimits_{r=1}^R(\bm{u}^{r}|{y})\mathbb{P}({y}) } \nonumber = \prod\nolimits_{r=1}^R F(i)^r.     
\end{eqnarray}
Proof done. \hfill~$\Box$

\section{Proof of Theorem 1}\label{Sec:theorem1}
\textit{Proof.} The density estimation at $i$th update $\bm{u}_i$ is
\begin{equation}
q(\bm{u}_i ) = \frac{1}{k+1}\sum_{j=1}^{k}\frac{1}{(2\pi)^{r/2}h^r }\exp\left(-\frac{|\bm{u}_{i,j}|^2}{2\bar{h}}\right)
\end{equation}
and  {similar in \cite{tang2017local} the average local density estimation in the neighborhood of $\bm{u}_i$ could be defined as}
\begin{equation}
\begin{split}
& \bar{q}(\bm{u}_i ) = \frac{1}{k}\sum\nolimits_{j=1}^{k}q(\bm{u}_{j} )\\
& = \frac{1}{k(k+1)}\sum\nolimits_{j=1}^{k}\frac{1}{(2\pi)^{\frac{r}{2}}h^r }\exp\left(-\frac{|\bm{u}_{i}-\bm{u}_{i,j}|^2}{2\bar{h}}\right)
\end{split}
\end{equation}

Assuming that $\bm{u}_{i,j}$ is uniformly distributed in a $o$ dimension ball $B_\beta$, where $\beta$ is the radius of the ball (i.e., the distance between $k$th nearest neighborhood and $\bm{u}_i$) we can calculate the expectation of $q(\bm{u}_i )$ and $\bar{q}(\bm{u}_i )$ from Theorem~$\bf{2}$
\begin{equation}
\mathbb{E}(\bar{q}(\bm{u}_i )) = \mathbb{E}(q(\bm{u}_i )) = \frac{1}{V} = \frac{2\pi^{(r-1)/2}\beta^{r-1}}{\Gamma((r-1)/2+1)}
\end{equation}
where $V$ is the volume of $B_\beta$. The rest of proof is followed by the McDiarmid's Inequality \cite{mcdiarmid1998concentration} which gives the upper bound of the probability that a function of $g(\bm{u}_{i,1}, \dots, \bm{u}_{i,j}, \dots, \bm{u}_{i,k})$ deviates with the expectation. Let $g:\mathbb{R}^r \to \mathbb{R}, \forall j, \forall \bm{u}_{i,1} , \bm{u}_{i,k}, \bm{u}'_{i,j}\in \mathcal{U}_i$, 
\begin{equation}
|g(\bm{u}_{i,1} , \dots, \bm{u}_{i,j},\dots,\bm_{i,k}) - g(\bm{u}_{i,1} , \dots, \bm{u}'_{i,j}, \dots, \bm{u}_{i,k})|  \leq \xi_{i,j}
\end{equation}
For all $\gamma > 0$,
\begin{equation}
\mathbb{P}(g - \mathbb{E}(g) \geq \gamma) \leq \exp\left(\frac{-2\gamma^2}{\sum_{j=1}^{k}c_{i,j}^{2}}\right)
\end{equation}
For $g_{1} = q(\bm{u}_{i})$,
\begin{equation}
\begin{split}
& |g_{1}(\bm{u}_{i,1}, \dots, \bm{u}_{i,j},\dots, \bm{u}_{i,k}) - g_{1}(\bm{u}_{i,1}, \dots, \bm{u}'_{i,j}, \dots, \bm{u}_{i,k})|\\
& = \frac{K(\bm{u}_{i,j}/\bar{h}) - K(\bm{u}'_{i,j}/\bar{h})}{\bar{h}^{r}(k+1)}\leq \frac{1 - \exp(-\beta^2 /2\bar{h})}{(2\pi)^{r/2}\bar{h}^r (k+1)} = \xi_1
\end{split}
\end{equation}
For $g_{2} = \bar{q}(\bm{u}_{i})$,
\begin{eqnarray}
&& |g_{2}(\bm{u}_{i,1}, \dots, \bm{u}_{i,j},\dots, \bm{u}_{i,k}) - \bm{u}_{2}(\bm{u}_{i,1}, \dots, \bm{u}'_{i,j}, \dots, \bm{u}_{i,k})|\nonumber\\
&& = \frac{K\left(\frac{\bm{u}_{i,j}}{\bar{h}}\right) - \left(\frac{\bm{u}'_{i,j}}{\bar{h}}\right) + 2\sum_{j=1}^{k}\left[K\left(\frac{\bm{u}_{i} - \bm{u}_{i,j}}{\bar{h}}\right) - \left(\frac{\bm{u}'_i - \bm{u}_{i,j}}{\bar{h}}\right)\right]}{\bar{h}^r (k+1)}\nonumber\\
&& \leq \frac{1 - \exp(-\beta^2 /2\bar{k}) + 2k(1 - \exp(-2\beta^2 /\bar{h}))}{(2\pi)^{r/2}\bar{h}^r (k+1)} = \xi_2
\end{eqnarray}
The probability of LMF is 
\begin{equation}
\begin{split}
\mathbb{P}(LMF(i) > \epsilon_m) & = \mathbb{P}[\bar{q}(\bm{u}_{i} - \epsilon_{m}q(\bm{u}_i ))]\\
& = \mathbb{P}(g - \mathbb{E}(g) > \theta)
\end{split}
\end{equation}
where $\theta = (\epsilon_m -1)/V$. From Theorem~$\bf{2}$, we are only interested in the case of $LMF(i) > 1$, i.e., $\epsilon_m > 1$, and $\theta > 0$. Following by the McDiarmid's Inequality, we have
\begin{eqnarray}
&& \mathbb{P}(LMF(i) > \epsilon_m )\leq \exp\left(-\frac{2\theta^2}{\sum_{j=1}^{k}\xi^2}\right) = \exp\left(-\frac{2\theta^2}{k\xi^2}\right)\nonumber\\
&& \leq \exp\left(-\frac{2(\epsilon_m -1)^2 (k+1)^2 (2\pi)^r \bar{h}^{2r}}{k(2k + \epsilon +1)^2 V^2}\right)
\end{eqnarray}
Proof done.  \hfill~$\Box$

{
\section{Evaluation of $\epsilon$ on LoMar}
\begin{figure}[ht]
	\centering
	\includegraphics[width = 0.82\columnwidth]{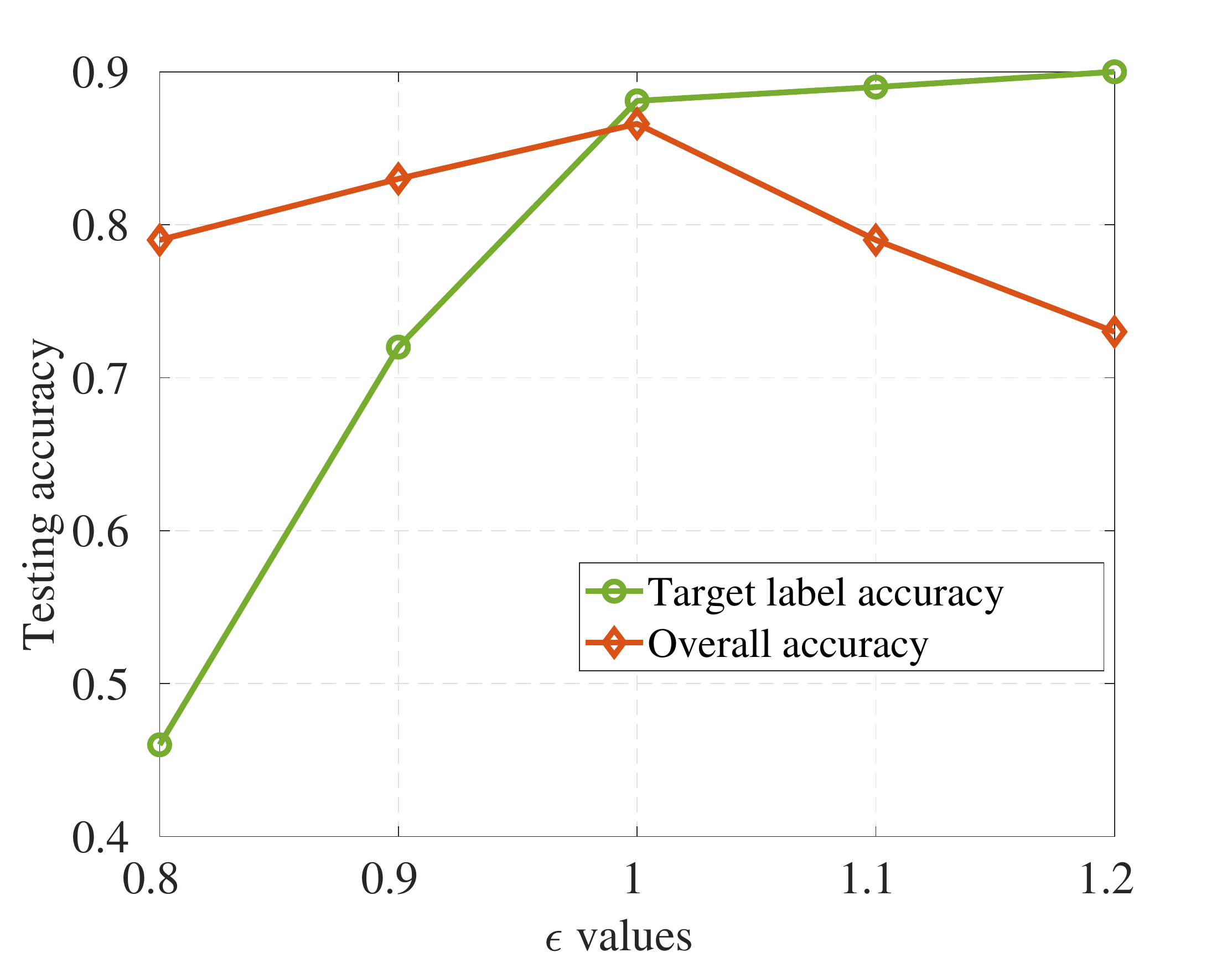}
	\caption{Testing accuracy of target and overall labels with different $\epsilon$ thresholds.  }
	\label{fig:epsilon}
\end{figure}
In this part, we study the defense performance of the proposed LoMar algorithm with different $\epsilon$ values. Specifically, we setup the $1-7$ label-flipping attack on MNIST with $100$ malicious clients and $1000$ clean clients, the other parameters are defined as the following: $\lambda = 0, \tau = 0.1$. From the results in Fig.~\ref{fig:epsilon}, we could notice that when $\epsilon = 1.0$, the averaged performance of considered metrics is the best. Specifically, though the target label accuracy increases slightly when $\epsilon > 1$, the overall performance decreases. We consider this might because of the removing of clean remote updates. And when $\epsilon <1$, the defender can not correctly remove the malicious from the FL system that leads a performance drop on target label accuracy.

}
\end{appendices}


\bibliographystyle{ieeetr}
\bibliography{ref}

\begin{thebibliography}{10}

\bibitem{konecny2017federated}
J.~Konečný, H.~B. McMahan, F.~X. Yu, P.~Richtárik, A.~T. Suresh, and
  D.~Bacon, ``Federated learning: Strategies for improving communication
  efficiency,'' 2017.

\bibitem{mcmahan2017communication}
B.~McMahan, E.~Moore, D.~Ramage, S.~Hampson, and B.~A. y~Arcas,
  ``Communication-efficient learning of deep networks from decentralized
  data,'' in {\em Artificial Intelligence and Statistics}, pp.~1273--1282,
  2017.

\bibitem{dabek2004vivaldi}
F.~Dabek, R.~Cox, F.~Kaashoek, and R.~Morris, ``Vivaldi: A decentralized
  network coordinate system,'' in {\em ACM SIGCOMM Computer Communication
  Review}, vol.~34, pp.~15--26, ACM, 2004.

\bibitem{bagdasaryan2018backdoor}
E.~Bagdasaryan, A.~Veit, Y.~Hua, D.~Estrin, and V.~Shmatikov, ``How to backdoor
  federated learning,'' {\em arXiv preprint arXiv:1807.00459}, 2018.

\bibitem{bhagoji2019analyzing}
A.~N. Bhagoji, S.~Chakraborty, P.~Mittal, and S.~Calo, ``Analyzing federated
  learning through an adversarial lens,'' in {\em International Conference on
  Machine Learning}, pp.~634--643, 2019.

\bibitem{melis2019exploiting}
L.~Melis, C.~Song, E.~De~Cristofaro, and V.~Shmatikov, ``Exploiting unintended
  feature leakage in collaborative learning,'' in {\em 2019 IEEE Symposium on
  Security and Privacy (SP)}, pp.~691--706, IEEE, 2019.

\bibitem{xiao2012adversarial}
H.~Xiao, H.~Xiao, and C.~Eckert, ``Adversarial label flips attack on support
  vector machines.,'' in {\em ECAI}, pp.~870--875, 2012.

\bibitem{fung2018mitigating}
C.~Fung, C.~J. Yoon, and I.~Beschastnikh, ``Mitigating sybils in federated
  learning poisoning,'' {\em arXiv e-prints}, pp.~arXiv--1808, 2018.

\bibitem{fang2020local}
M.~Fang, X.~Cao, J.~Jia, and N.~Gong, ``Local model poisoning attacks to
  byzantine-robust federated learning,'' in {\em 29th $\{$USENIX$\}$ Security
  Symposium ($\{$USENIX$\}$ Security 20)}, pp.~1605--1622, 2020.

\bibitem{bhagoji2018model}
A.~N. Bhagoji, S.~Chakraborty, S.~Calo, and P.~Mittal, ``Model poisoning
  attacks in federated learning,'' in {\em In Workshop on Security in Machine
  Learning (SecML), collocated with the 32nd Conference on Neural Information
  Processing Systems (NeurIPS’18)}, 2018.

\bibitem{blanchard2017machine}
P.~Blanchard, R.~Guerraoui, J.~Stainer, {\em et~al.}, ``Machine learning with
  adversaries: Byzantine tolerant gradient descent,'' in {\em Advances in
  Neural Information Processing Systems}, pp.~119--129, 2017.

\bibitem{yin2018byzantine}
D.~Yin, Y.~Chen, R.~Kannan, and P.~Bartlett, ``Byzantine-robust distributed
  learning: Towards optimal statistical rates,'' in {\em International
  Conference on Machine Learning}, pp.~5650--5659, 2018.

\bibitem{el2018hidden}
E.~M. El~Mhamdi, R.~Guerraoui, and S.~L.~A. Rouault, ``The hidden vulnerability
  of distributed learning in byzantium,'' in {\em International Conference on
  Machine Learning}, no.~CONF, 2018.

\bibitem{baracaldo2017mitigating}
N.~Baracaldo, B.~Chen, H.~Ludwig, and J.~A. Safavi, ``Mitigating poisoning
  attacks on machine learning models: A data provenance based approach,'' in
  {\em Proceedings of the 10th ACM Workshop on Artificial Intelligence and
  Security}, pp.~103--110, ACM, 2017.

\bibitem{barreno2010security}
M.~Barreno, B.~Nelson, A.~D. Joseph, and J.~D. Tygar, ``The security of machine
  learning,'' {\em Machine Learning}, vol.~81, no.~2, pp.~121--148, 2010.

\bibitem{xie2018generalized}
C.~Xie, O.~Koyejo, and I.~Gupta, ``Generalized byzantine-tolerant sgd,'' {\em
  arXiv preprint arXiv:1802.10116}, 2018.

\bibitem{tolpegin2020data}
V.~Tolpegin, S.~Truex, M.~E. Gursoy, and L.~Liu, ``Data poisoning attacks
  against federated learning systems,'' in {\em European Symposium on Research
  in Computer Security}, pp.~480--501, Springer, 2020.

\bibitem{shen2016auror}
S.~Shen, S.~Tople, and P.~Saxena, ``Auror: Defending against poisoning attacks
  in collaborative deep learning systems,'' in {\em Proceedings of the 32nd
  Annual Conference on Computer Security Applications}, pp.~508--519, 2016.

\bibitem{tang2015kerneladasyn}
B.~Tang and H.~He, ``Kerneladasyn: Kernel based adaptive synthetic data
  generation for imbalanced learning,'' in {\em 2015 IEEE Congress on
  Evolutionary Computation (CEC)}, pp.~664--671, IEEE, 2015.

\bibitem{huang2020metapoison}
W.~R. Huang, J.~Geiping, L.~Fowl, G.~Taylor, and T.~Goldstein, ``Metapoison:
  Practical general-purpose clean-label data poisoning,'' {\em Advances in
  Neural Information Processing Systems}, vol.~33, 2020.

\bibitem{fraboni2021free}
Y.~Fraboni, R.~Vidal, and M.~Lorenzi, ``Free-rider attacks on model aggregation
  in federated learning,'' in {\em International Conference on Artificial
  Intelligence and Statistics}, pp.~1846--1854, PMLR, 2021.

\bibitem{loftsgaarden1965nonparametric}
D.~O. Loftsgaarden, C.~P. Quesenberry, {\em et~al.}, ``A nonparametric estimate
  of a multivariate density function,'' {\em The Annals of Mathematical
  Statistics}, vol.~36, no.~3, pp.~1049--1051, 1965.

\bibitem{breiman1977variable}
L.~Breiman, W.~Meisel, and E.~Purcell, ``Variable kernel estimates of
  multivariate densities,'' {\em Technometrics}, vol.~19, no.~2, pp.~135--144,
  1977.

\bibitem{tang2017local}
B.~Tang and H.~He, ``A local density-based approach for outlier detection,''
  {\em Neurocomputing}, vol.~241, pp.~171--180, 2017.

\bibitem{li2020learning}
S.~Li, Y.~Cheng, W.~Wang, Y.~Liu, and T.~Chen, ``Learning to detect malicious
  clients for robust federated learning,'' {\em arXiv preprint
  arXiv:2002.00211}, 2020.

\bibitem{paszke2017automatic}
A.~Paszke, S.~Gross, S.~Chintala, G.~Chanan, E.~Yang, Z.~DeVito, Z.~Lin,
  A.~Desmaison, L.~Antiga, and A.~Lerer, ``Automatic differentiation in
  pytorch,'' in {\em NIPS-W}, 2017.

\bibitem{lecun1998gradient}
Y.~LeCun, L.~Bottou, Y.~Bengio, P.~Haffner, {\em et~al.}, ``Gradient-based
  learning applied to document recognition,'' {\em Proceedings of the IEEE},
  vol.~86, no.~11, pp.~2278--2324, 1998.

\bibitem{bay2000uci}
S.~D. Bay, D.~Kibler, M.~J. Pazzani, and P.~Smyth, ``The uci kdd archive of
  large data sets for data mining research and experimentation,'' {\em ACM
  SIGKDD explorations newsletter}, vol.~2, no.~2, pp.~81--85, 2000.

\bibitem{asuncion2007uci}
A.~Asuncion and D.~Newman, ``Uci machine learning repository,'' 2007.

\bibitem{Cao18}
Q.~Cao, L.~Shen, W.~Xie, O.~M. Parkhi, and A.~Zisserman, ``Vggface2: A dataset
  for recognising faces across pose and age,'' in {\em International Conference
  on Automatic Face and Gesture Recognition}, 2018.

\bibitem{marcel2010torchvision}
S.~Marcel and Y.~Rodriguez, ``Torchvision the machine-vision package of
  torch,'' in {\em Proceedings of the 18th ACM international conference on
  Multimedia}, pp.~1485--1488, 2010.

\bibitem{douceur2002sybil}
J.~R. Douceur, ``The sybil attack,'' in {\em International workshop on
  peer-to-peer systems}, pp.~251--260, Springer, 2002.

\bibitem{biggio2012poisoning}
B.~Biggio, B.~Nelson, and P.~Laskov, ``Poisoning attacks against support vector
  machines,'' in {\em Proceedings of the 29th International Coference on
  International Conference on Machine Learning}, pp.~1467--1474, 2012.

\bibitem{delong1988comparing}
E.~R. DeLong, D.~M. DeLong, and D.~L. Clarke-Pearson, ``Comparing the areas
  under two or more correlated receiver operating characteristic curves: a
  nonparametric approach,'' {\em Biometrics}, pp.~837--845, 1988.

\bibitem{angluin1988learning}
D.~Angluin and P.~Laird, ``Learning from noisy examples,'' {\em Machine
  Learning}, vol.~2, no.~4, pp.~343--370, 1988.

\bibitem{aslam1996sample}
J.~A. Aslam and S.~E. Decatur, ``On the sample complexity of noise-tolerant
  learning,'' {\em Information Processing Letters}, vol.~57, no.~4,
  pp.~189--195, 1996.

\bibitem{stempfel:hal-00186391}
G.~Stempfel, L.~Ralaivola, and F.~Denis, ``{Learning from Noisy Data using
  Hyperplane Sampling and Sample Averages},'' May 2007.
\newblock working paper or preprint.

\bibitem{natarajan2013learning}
N.~Natarajan, I.~S. Dhillon, P.~K. Ravikumar, and A.~Tewari, ``Learning with
  noisy labels,'' in {\em Advances in neural information processing systems},
  pp.~1196--1204, 2013.

\bibitem{shafahi2018poison}
A.~Shafahi, W.~R. Huang, M.~Najibi, O.~Suciu, C.~Studer, T.~Dumitras, and
  T.~Goldstein, ``Poison frogs! targeted clean-label poisoning attacks on
  neural networks,'' in {\em Advances in Neural Information Processing
  Systems}, pp.~6103--6113, 2018.

\bibitem{Gu2017}
T.~Gu, B.~Dolan{-}Gavitt, and S.~Garg, ``Badnets: Identifying vulnerabilities
  in the machine learning model supply chain,'' {\em CoRR},
  vol.~abs/1708.06733, 2017.

\bibitem{Yang2017}
G.~Yang, N.~Z. Gong, and Y.~Cai, ``Fake co-visitation injection attacks to
  recommender systems,'' in {\em Proceedings 2017 Network and Distributed
  System Security Symposium}, Internet Society, 2017.

\bibitem{jagielski2018manipulating}
M.~Jagielski, A.~Oprea, B.~Biggio, C.~Liu, C.~Nita-Rotaru, and B.~Li,
  ``Manipulating machine learning: Poisoning attacks and countermeasures for
  regression learning,'' in {\em 2018 IEEE Symposium on Security and Privacy
  (SP)}, pp.~19--35, IEEE, 2018.

\bibitem{yang2017fake}
G.~Yang, N.~Z. Gong, and Y.~Cai, ``Fake co-visitation injection attacks to
  recommender systems.,'' in {\em NDSS}, 2017.

\bibitem{papernot2016cleverhans}
N.~Papernot, I.~Goodfellow, R.~Sheatsley, R.~Feinman, and P.~McDaniel,
  ``cleverhans v2. 0.0: an adversarial machine learning library,'' {\em arXiv
  preprint arXiv:1610.00768}, vol.~10, 2016.

\bibitem{papernot2017practical}
N.~Papernot, P.~McDaniel, I.~Goodfellow, S.~Jha, Z.~B. Celik, and A.~Swami,
  ``Practical black-box attacks against machine learning,'' in {\em Proceedings
  of the 2017 ACM on Asia conference on computer and communications security},
  pp.~506--519, ACM, 2017.

\bibitem{papernot2018sok}
N.~Papernot, P.~McDaniel, A.~Sinha, and M.~P. Wellman, ``Sok: Security and
  privacy in machine learning,'' in {\em 2018 IEEE European Symposium on
  Security and Privacy (EuroS\&P)}, pp.~399--414, IEEE, 2018.

\bibitem{nasr2019comprehensive}
M.~Nasr, R.~Shokri, and A.~Houmansadr, ``Comprehensive privacy analysis of deep
  learning: Passive and active white-box inference attacks against centralized
  and federated learning,'' in {\em 2019 IEEE Symposium on Security and Privacy
  (SP)}, pp.~739--753, IEEE, 2019.

\bibitem{zhang2020poisongan}
J.~Zhang, B.~Chen, X.~Cheng, H.~T.~T. Binh, and S.~Yu, ``Poisongan: Generative
  poisoning attacks against federated learning in edge computing systems,''
  {\em IEEE Internet of Things Journal}, 2020.

\bibitem{cretu2008casting}
G.~F. Cretu, A.~Stavrou, M.~E. Locasto, S.~J. Stolfo, and A.~D. Keromytis,
  ``Casting out demons: Sanitizing training data for anomaly sensors,'' in {\em
  2008 IEEE Symposium on Security and Privacy (sp 2008)}, pp.~81--95, IEEE,
  2008.

\bibitem{suciu2018does}
O.~Suciu, R.~Marginean, Y.~Kaya, H.~Daume~III, and T.~Dumitras, ``When does
  machine learning $\{$FAIL$\}$? generalized transferability for evasion and
  poisoning attacks,'' in {\em 27th $\{$USENIX$\}$ Security Symposium
  ($\{$USENIX$\}$ Security 18)}, pp.~1299--1316, 2018.

\bibitem{tran2018spectral}
B.~Tran, J.~Li, and A.~Madry, ``Spectral signatures in backdoor attacks,'' in
  {\em Advances in Neural Information Processing Systems}, pp.~8000--8010,
  2018.

\bibitem{rieck2008self}
K.~Rieck, S.~Wahl, P.~Laskov, P.~Domschitz, and K.-R. M{\"u}ller, ``A
  self-learning system for detection of anomalous sip messages,'' in {\em
  International Conference on Principles, Systems and Applications of IP
  Telecommunications}, pp.~90--106, Springer, 2008.

\bibitem{koh2017understanding}
P.~W. Koh and P.~Liang, ``Understanding black-box predictions via influence
  functions,'' in {\em Proceedings of the 34th International Conference on
  Machine Learning-Volume 70}, pp.~1885--1894, JMLR. org, 2017.

\bibitem{burkard2017analysis}
C.~Burkard and B.~Lagesse, ``Analysis of causative attacks against svms
  learning from data streams,'' in {\em Proceedings of the 3rd ACM on
  International Workshop on Security And Privacy Analytics}, pp.~31--36, 2017.

\bibitem{mei2015using}
S.~Mei and X.~Zhu, ``Using machine teaching to identify optimal training-set
  attacks on machine learners,'' in {\em Twenty-Ninth AAAI Conference on
  Artificial Intelligence}, 2015.

\bibitem{damaskinos2018asynchronous}
G.~Damaskinos, R.~Guerraoui, R.~Patra, M.~Taziki, {\em et~al.}, ``Asynchronous
  byzantine machine learning (the case of sgd),'' in {\em International
  Conference on Machine Learning}, pp.~1145--1154, 2018.

\bibitem{xie2019zeno}
C.~Xie, S.~Koyejo, and I.~Gupta, ``Zeno: Distributed stochastic gradient
  descent with suspicion-based fault-tolerance,'' in {\em International
  Conference on Machine Learning}, pp.~6893--6901, 2019.

\bibitem{silverman2018density}
B.~W. Silverman, {\em Density estimation for statistics and data analysis}.
\newblock Routledge, 2018.

\bibitem{latecki2007outlier}
L.~J. Latecki, A.~Lazarevic, and D.~Pokrajac, ``Outlier detection with kernel
  density functions,'' in {\em International Workshop on Machine Learning and
  Data Mining in Pattern Recognition}, pp.~61--75, Springer, 2007.

\bibitem{tang2002enhancing}
J.~Tang, Z.~Chen, A.~W.-C. Fu, and D.~W. Cheung, ``Enhancing effectiveness of
  outlier detections for low density patterns,'' in {\em Pacific-Asia
  Conference on Knowledge Discovery and Data Mining}, pp.~535--548, Springer,
  2002.

\bibitem{schubert2012evaluation}
E.~Schubert, R.~Wojdanowski, A.~Zimek, and H.-P. Kriegel, ``On evaluation of
  outlier rankings and outlier scores,'' in {\em Proceedings of the 2012 SIAM
  International Conference on Data Mining}, pp.~1047--1058, SIAM, 2012.

\bibitem{guo2014detecting}
S.~Guo, H.~Zhang, Z.~Zhong, J.~Chen, Q.~Cao, and T.~He, ``Detecting faulty
  nodes with data errors for wireless sensor networks,'' {\em ACM Transactions
  on Sensor Networks (TOSN)}, vol.~10, no.~3, pp.~1--27, 2014.

\bibitem{qin2019scalable}
X.~Qin, L.~Cao, E.~A. Rundensteiner, and S.~Madden, ``Scalable kernel density
  estimation-based local outlier detection over large data streams.,'' in {\em
  EDBT}, pp.~421--432, 2019.

\bibitem{mcdiarmid1998concentration}
C.~McDiarmid, ``Concentration,'' in {\em Probabilistic methods for algorithmic
  discrete mathematics}, pp.~195--248, Springer, 1998.

\end{thebibliography}

\end{document}